\documentclass[twoside,11pt]{article}
\pdfoutput=1

\usepackage{jmlr2e}

\usepackage[utf8]{inputenc} %
\usepackage[T1]{fontenc}    %
\usepackage{hyperref}       %
\usepackage{url}            %
\usepackage{booktabs}       %
\usepackage{amsfonts}       %
\usepackage{nicefrac}       %
\usepackage{microtype}      %

\usepackage{graphicx}
\usepackage{amsmath}
\usepackage{xcolor}
\usepackage{wrapfig}
\usepackage{cite}
\usepackage{enumitem}
\usepackage{placeins}
\usepackage[bottom]{footmisc}
\usepackage{subcaption}

\newlength{\twosubht}
\newsavebox{\twosubbox}

\ShortHeadings{Benchmarking Unsupervised Object Representations}{Weis, Chitta, Sharma, Brendel, Bethge, Geiger and Ecker}
\firstpageno{1}
\newcommand{\ba}{\mathbf{a}}\newcommand{\bA}{\mathbf{A}}

\newcommand{\bc}{\mathbf{c}}

\newcommand{\bh}{\mathbf{h}}

\newcommand{\bk}{\mathbf{k}}

\newcommand{\bm}{\mathbf{m}}

\newcommand{\bo}{\mathbf{o}}
\newcommand{\bp}{\mathbf{p}}
\newcommand{\bq}{\mathbf{q}}
\newcommand{\br}{\mathbf{r}}
\newcommand{\bs}{\mathbf{s}}
\newcommand{\bT}{\mathbf{T}}

\newcommand{\bW}{\mathbf{W}}
\newcommand{\bx}{\mathbf{x}}
\newcommand{\by}{\mathbf{y}}\newcommand{\bY}{\mathbf{Y}}
\newcommand{\bz}{\mathbf{z}}

\newcommand{\bgamma}{\boldsymbol{\gamma}}

\newcommand{\bmu}{\boldsymbol{\mu}}

\newcommand{\bsigma}{\boldsymbol{\sigma}}

\newcommand{\nR}{\mathbb{R}}

\newcommand{\figref}[1]{Fig.~\ref{#1}}
\newcommand{\secref}[1]{Section~\ref{#1}}

\newcommand{\eqnref}[1]{Eq.~\eqref{#1}}
\newcommand{\tabref}[1]{Table~\ref{#1}}

\makeatletter
\DeclareRobustCommand\onedot{\futurelet\@let@token\@onedot}
\def\@onedot{\ifx\@let@token.\else.\null\fi\xspace}

\makeatother

\newcommand{\boldparagraph}[1]{\noindent{\bf #1:}}

\definecolor{darkgreen}{rgb}{0,0.7,0}
\definecolor{brown}{rgb}{0.59,0.29,0.00}
\definecolor{gray}{gray}{0.9}

\definecolor{mediumgray}{rgb}{0.5,0.5,0.5}
\newcommand{\pmsd}[1]{{\color{mediumgray}{\scriptsize $\pm$ #1}}}

\newcommand{\datatable}
{
\begin{table}[t!]
    \centering
    \resizebox{\textwidth}{!}{%
    \begin{tabular}{lllllllll}
        \toprule
        & \multicolumn{6}{c}{Objects} & \multicolumn{2}{c}{Background} \\
        \cmidrule(r){2-7}
        \cmidrule(r){8-9}
        Data Set & Shape & Motion & Count Over Sequence & Size Variation & Orientation & Color & Motion & Color \\
        \midrule
        SpMOT & 4 Templates (2D) & Linear & Varies (0-3) & Minimal & Fixed & 6 Colors & None & Black \\
        VOR & 2 Templates (3D) & Static & Varies (0-4) & Moderate & Random & 6 Colors & Moving Camera & Random \\
        VMDS & 3 Templates (2D) & Non-Linear & Fixed (1-4) & Moderate & Random & 256\textsuperscript{3} Colors & None & 256\textsuperscript{3} Colors \\
        texVMDS & 3 Templates (2D) & Non-Linear & Fixed (1-4) & Moderate & Random & ImageNet textures & Linear & ImageNet crop \\
        \bottomrule
    \end{tabular}}
    \includegraphics[width=\textwidth]{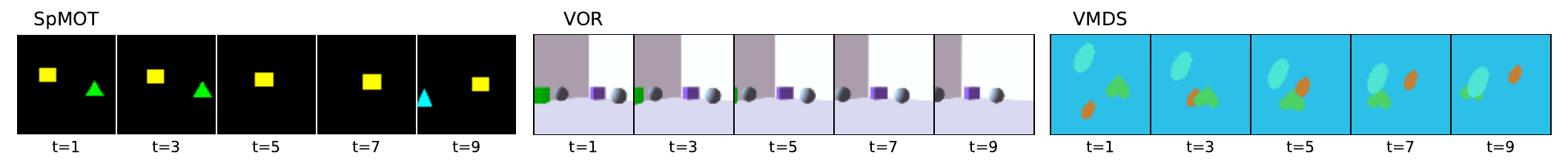}
    \caption{Summary of data sets and example video sequences. See Appendix~\ref{datasets_suppl} for details.}
    \label{tab:data}
\end{table}
}

\newcommand{\resultstable}{
\begin{table}[t!]
    \centering
    \resizebox{\textwidth}{!}{%
    \begin{tabular}{llllllllll}
        \toprule
        Model  & MOTA $\uparrow$ & MOTP $\uparrow$ & MD $\uparrow$ & MT $\uparrow$ & Match $\uparrow$ & Miss $\downarrow$ & ID S. $\downarrow$ & FPs $\downarrow$ & MSE $\downarrow$ \\
        \midrule
        &\multicolumn{9}{c}{\textbf{SpMOT}} \\
        \cmidrule(r){2-10}
        \textsc{k-Means} & 36.2 \pmsd{0.0} & 77.6 \pmsd{0.0} & 76.8 \pmsd{0.0} & 76.3 \pmsd{0.0} & 80.5 \pmsd{0.0} & 19.3 \pmsd{0.0} & \textbf{0.2 \pmsd{0.0}} & 44.3 \pmsd{0.0} & - \\
        \textsc{MONet} & 70.2 \pmsd{0.8} & 89.6 \pmsd{1.0} & 92.4 \pmsd{0.6} & 50.4 \pmsd{2.4} & 75.3 \pmsd{1.3} & 4.4 \pmsd{0.4} & 20.3 \pmsd{1.6} & 5.1 \pmsd{0.5} & 13.0 \pmsd{2.0} \\ 
        \textsc{ViMON} & 92.9 \pmsd{0.2} & \textbf{91.8 \pmsd{0.2}} & 87.7 \pmsd{0.8} & 87.2 \pmsd{0.8} & 95.0 \pmsd{0.2} & 4.8 \pmsd{0.2} & \textbf{0.2 \pmsd{0.0}} & 2.1 \pmsd{0.1} & 11.1 \pmsd{0.6} \\
        \textsc{TBA}   & 79.7 \pmsd{15.0} & 71.2 \pmsd{0.3} & 83.4 \pmsd{9.7} & 80.0 \pmsd{13.6} & 87.8 \pmsd{9.0} & 9.6 \pmsd{6.0} & 2.6 \pmsd{3.0} & 8.1 \pmsd{6.0} & 11.9 \pmsd{1.9} \\
         \textsc{OP3}  & 89.1 \pmsd{5.1} & 78.4 \pmsd{2.4} & 92.4 \pmsd{4.0} & 91.8 \pmsd{3.8} & \textbf{95.9 \pmsd{2.2}} & 3.7 \pmsd{2.2} & 0.4 \pmsd{0.0} & 6.8 \pmsd{2.9} & 13.3 \pmsd{11.9} \\
        \textsc{SCALOR} & \textbf{94.9 \pmsd{0.5}} & 80.2 \pmsd{0.1} & \textbf{96.4 \pmsd{0.1}} & \textbf{93.2 \pmsd{0.7}} & \textbf{95.9 \pmsd{0.4}} & \textbf{2.4 \pmsd{0.0}} & 1.7 \pmsd{0.4} & \textbf{1.0 \pmsd{0.1}} & \textbf{3.4 \pmsd{0.1}} \\
        \midrule
        &\multicolumn{9}{c}{\textbf{VOR}} \\
        \cmidrule(r){2-10}
        \textsc{k-Means} & -38.0 \pmsd{0.1} & 76.5 \pmsd{0.0} & 69.9 \pmsd{0.1} & 62.9 \pmsd{0.1} & 72.7 \pmsd{0.1} & 22.1 \pmsd{0.0} & 5.2 \pmsd{0.0} & 110.7 \pmsd{0.1} & - \\
        \textsc{MONet} & 37.0 \pmsd{6.8} & 81.7 \pmsd{0.5} & 76.9 \pmsd{2.2} & 37.3 \pmsd{7.8} & 64.4 \pmsd{5.0} & 15.8 \pmsd{1.6} & 19.8 \pmsd{3.5} & 27.4 \pmsd{2.3} & 12.2 \pmsd{1.4} \\
        \textsc{ViMON} & \textbf{89.0 \pmsd{0.0}} & \textbf{89.5 \pmsd{0.5}} & \textbf{90.4 \pmsd{0.5}} & \textbf{90.0 \pmsd{0.4}} & \textbf{93.2 \pmsd{0.4}} & \textbf{6.5 \pmsd{0.4}} & 0.3 \pmsd{0.0} & 4.2 \pmsd{0.4} & 6.4 \pmsd{0.6} \\
        \textsc{OP3}   & 65.4 \pmsd{0.6} & 89.0 \pmsd{0.6} & 88.0 \pmsd{0.6} & 85.4 \pmsd{0.5} & 90.7 \pmsd{0.3} & 8.2 \pmsd{0.4} & 1.1 \pmsd{0.2} & 25.3 \pmsd{0.6} & \textbf{3.0 \pmsd{0.1}} \\
        \textsc{SCALOR} & 74.6 \pmsd{0.4} & 86.0 \pmsd{0.2} & 76.0 \pmsd{0.4} & 75.9 \pmsd{0.4} & 77.9 \pmsd{0.4} & 22.1 \pmsd{0.4} & \textbf{0.0 \pmsd{0.0}} & \textbf{3.3 \pmsd{0.2}} & 6.4 \pmsd{0.1} \\
        \midrule
        &\multicolumn{9}{c}{\textbf{VMDS}} \\
        \cmidrule(r){2-10}
        \textsc{k-Means} & -3.3 \pmsd{0.0} & 89.8 \pmsd{0.0} & \textbf{98.3 \pmsd{0.0}} & 93.3 \pmsd{0.1} & 96.4 \pmsd{0.0} & \textbf{1.0 \pmsd{0.0}} & 2.6 \pmsd{0.0} & 99.7 \pmsd{0.0} & - \\
        \textsc{MONet} & 49.4 \pmsd{3.6} & 78.6 \pmsd{1.8} & 74.2 \pmsd{1.7} & 35.7 \pmsd{0.8} & 66.7 \pmsd{0.7} & 13.6 \pmsd{1.0} & 19.7 \pmsd{0.6} & 17.2 \pmsd{3.1} & 22.2 \pmsd{2.2}\\
        \textsc{ViMON} & 86.8 \pmsd{0.3} & 86.8 \pmsd{0.0} & 86.2 \pmsd{0.3} & 85.0 \pmsd{0.3} & 92.3 \pmsd{0.2} & 7.0 \pmsd{0.2} & 0.7 \pmsd{0.0} & 5.5 \pmsd{0.1} & 10.7 \pmsd{0.1}\\
        \textsc{TBA}  & 54.5 \pmsd{12.1} & 75.0 \pmsd{0.9} & 62.9 \pmsd{5.9} & 58.3 \pmsd{6.1} & 75.9 \pmsd{4.3} & 21.0 \pmsd{4.2} & 3.2 \pmsd{0.3} & 21.4 \pmsd{7.8} & 28.1 \pmsd{2.0}\\
        \textsc{OP3}   & \textbf{91.7 \pmsd{1.7}} & \textbf{93.6 \pmsd{0.4}} & 96.8 \pmsd{0.5} & \textbf{96.3 \pmsd{0.4}} & \textbf{97.8 \pmsd{0.1}} & 2.0 \pmsd{0.1} & \textbf{0.2 \pmsd{0.0}} & 6.1 \pmsd{1.5} & \textbf{4.3 \pmsd{0.2}}\\
        \textsc{SCALOR} & 74.1 \pmsd{1.2} & 87.6 \pmsd{0.4} & 67.9 \pmsd{1.1} & 66.7 \pmsd{1.1} & 78.4 \pmsd{1.0} & 20.7 \pmsd{1.0} & 0.8 \pmsd{0.0} & \textbf{4.4 \pmsd{0.4}} & 14.0  \pmsd{0.1} \\
        \bottomrule
    \end{tabular}}
    \caption{Analysis of SOTA object-centric representation learning models for MOT. Results shown as mean $\pm$ standard deviation of three runs with different random training seeds.}
    \label{tab:result}
\end{table}
}

\newcommand{\resultstablefull}{
\begin{table}[b!]
    \centering
    \resizebox{\textwidth}{!}{%
    \begin{tabular}{lllllllllll}
        \toprule
        Model  & Run & MOTA $\uparrow$ & MOTP $\uparrow$ & MD $\uparrow$ & MT $\uparrow$ & Match $\uparrow$ & Miss $\downarrow$ & ID S. $\downarrow$ & FPs $\downarrow$ & MSE $\downarrow$ \\
        \midrule
        &\multicolumn{9}{c}{\textbf{SpMOT}} \\
        \cmidrule(r){2-11}
        & 1 & 36.2 & 77.6 & 76.8 & 76.3 & 80.5 & 19.2 & 0.2 & 44.3 & -  \\
        \textsc{k-Means} & 2 & 36.2 & 77.6 & 76.8 & 76.3 & 80.5 & 19.2 & 0.2 & 44.3 & -  \\
        & 3 & 36.3 & 77.6 & 76.8 & 76.2 & 80.5 & 19.3 & 0.2 & 44.2 & -  \\
         \cmidrule(r){2-11}
         & 1 & 70.0 & 90.6 & 92.8 & 49.4 & 74.7 & 4.1 & 21.2 & 4.7 & 10.4\\
        \textsc{MONet} & 2 & 69.4 & 90.0 & 92.7 & 48.1 & 74.2 & 4.1 & 21.6 & 4.8 & 13.4\\
         & 3 & 71.3 & 88.1 & 91.6 & 53.8 & 77.1 & 4.9 & 18.0 & 5.8 & 15.2\\
         \cmidrule(r){2-11}
         & 1 & 92.7 & 92.0 & 87.5 & 87.0 & 94.9 & 4.9 & 0.2 & 2.2 & 10.5\\
        \textsc{ViMON} & 2 & 92.8 & 92.0 & 86.9 & 86.3 & 94.8 & 5.0 & 0.2 & 2.0 & 11.8\\
         & 3 & 93.2 & 91.6 & 88.8 & 88.3 & 95.2 & 4.6 & 0.2 & 2.0 & 10.9\\
         \cmidrule(r){2-11}
         & 1 & 90.5 & 71.4 & 90.2 & 89.8 & 94.4 & 5.3 & 0.3 & 3.9 & 10.3 \\
        \textsc{TBA} & 2 & 58.4 & 70.7 & 69.6 & 60.8 & 75.0 & 18.1 & 6.9 & 16.6 & 14.6\\
         & 3 & 90.1 & 71.5 & 90.3 & 89.4 & 94.0 & 5.5 & 0.5 & 3.9 & 10.9 \\
         \cmidrule(r){2-11}
        & 1 & 92.4 & 80.0 & 94.5 & 93.7 & 97.3 & 2.4 & 0.4 & 4.8  & 4.3\\
        \textsc{OP3} & 2 & 81.9 & 74.9 & 86.9 & 86.5 & 92.8 & 6.8 & 0.3 & 10.9 & 30.1\\
        & 3 & 92.9 & 80.1 & 95.9 & 95.2 & 97.6 & 2.0 & 0.4 & 4.7 & 5.6\\
         \cmidrule(r){2-11}
        & 1 & 94.4 & 80.1 & 96.5 & 92.3 & 95.4 & 2.4 & 2.2 & 1.0 & 3.3 \\
        \textsc{SCALOR} & 2 & 94.7 & 80.2 & 96.4 & 93.1 & 95.8 & 2.4 & 1.8 & 1.1 & 3.4 \\
        & 3 & 95.5 & 80.2 & 96.3 & 94.0 & 96.4 & 2.4 & 1.2 & 0.9 & 3.6 \\
        \midrule
        &\multicolumn{9}{c}{\textbf{VOR}} \\
        \cmidrule(r){2-11}
        & 1 & -38.0 & 76.5 & 69.8 & 62.8 & 72.6 & 22.1 & 5.3 & 110.7 & -  \\
        \textsc{k-Means} & 2 & -38.1 & 76.5 & 69.8 & 63.0 & 72.7 & 22.1 & 5.2 & 110.8 & -  \\
        & 3 & -37.9 & 76.5 & 70.0 & 62.9 & 72.8 & 22.0 & 5.2 & 110.7 & -  \\
        \cmidrule(r){2-11}
        & 1 & 28.0 & 81.3 & 73.8 & 26.7 & 57.4 & 18.0 & 24.6 & 29.4 & 14.1 \\
        \textsc{MONet} & 2 & 44.5 & 82.4 & 78.2 & 45.4 & 68.7 & 15.0 & 16.3 & 24.2 & 11.8 \\
        & 3 & 38.5 & 81.6 & 78.7 & 39.8 & 67.0 & 14.4 & 18.5 & 28.5 & 10.8 \\
        \cmidrule(r){2-11}
        & 1 & 89.0 & 88.9 & 90.2 & 89.8 & 92.9 & 6.8 & 0.3 & 3.9 & 7.1\\
        \textsc{ViMON} & 2 & 89.0 & 89.8 & 89.9 & 89.6 & 93.0 & 6.8 & 0.2 & 4.0 & 6.2\\
        & 3 & 89.0 & 89.9 & 91.0 & 90.6 & 93.8 & 6.0 & 0.2 & 4.8 & 5.9\\
        \cmidrule(r){2-11}
        & 1 & 64.8 & 89.5 & 87.2 & 85.1 & 90.3 & 8.8 & 0.9 & 25.5 & 3.1 \\
        \textsc{OP3} & 2 & 66.2 & 88.1 & 88.6 & 85.1 & 90.7 & 7.9 & 1.4 & 24.5 & 2.9 \\
        & 3 & 65.3 & 89.3 & 88.2 & 86.1 & 91.1 & 8.0 & 0.9 & 25.8 & 3.0 \\
        \cmidrule(r){2-11}
        & 1 & 74.1 & 85.8 & 75.6 & 75.5 & 77.4 & 22.6 & 0.0 & 3.3 & 6.4 \\
        \textsc{SCALOR} & 2 & 74.6 & 86.0 & 75.9 & 75.9 & 78.1 & 21.9 & 0.1 & 3.5 & 6.4 \\
        & 3 & 75.1 & 86.1 & 76.5 & 76.4 & 78.2 & 21.7 & 0.0 & 3.1 & 6.3 \\
        \bottomrule
    \end{tabular}}
    \caption{Analysis of SOTA object-centric representation learning models for MOT. Results for three runs with different random training seeds.}
    \label{tab:resultfull}
\end{table}
}

\newcommand{\resultstablefulltwo}{
\begin{table}[b!]
    \centering
    \resizebox{\textwidth}{!}{%
    \begin{tabular}{lllllllllll}
        \toprule
        Model  & Run & MOTA $\uparrow$ & MOTP $\uparrow$ & MD $\uparrow$ & MT $\uparrow$ & Match $\uparrow$ & Miss $\downarrow$ & ID S. $\downarrow$ & FPs $\downarrow$ & MSE $\downarrow$ \\
        \midrule
        &\multicolumn{9}{c}{\textbf{VMDS}} \\
        \cmidrule(r){2-11}
        & 1 & -3.3 & 89.8 & 98.3 & 93.4 & 96.4 & 1.0 & 2.6 & 99.7 & - \\
        \textsc{k-Means}  & 2 & -3.3 & 89.8 & 98.2 & 93.2 & 96.4 & 1.1 & 2.6 & 99.7 & -  \\
        & 3 & -3.2 & 89.8 & 98.2 & 93.4 & 96.4 & 1.0 & 2.6 & 99.6 & -  \\
        \cmidrule(r){2-11}
        & 1 & 51.7 & 79.6 & 75.1 & 36.7 & 67.6 & 12.9 & 19.5 & 15.9 & 20.8 \\
        \textsc{MONet} & 2 & 44.3 & 76.1 & 71.8 & 34.8 & 65.9 & 15.0 & 19.1 & 21.5 & 25.3 \\
        & 3 & 52.2 & 80.2 & 75.6 & 35.5 & 66.5 & 13.0 & 20.5 & 14.2 & 20.4 \\
        \cmidrule(r){2-11}
        & 1 & 87.0 & 86.8 & 86.7 & 85.4 & 92.4 & 6.8 & 0.7 & 5.5 & 10.6 \\
        \textsc{ViMON} & 2 & 87.1 & 86.8 & 86.1 & 85.1 & 92.3 & 7.1 & 0.6 & 5.3 & 10.8\\
        & 3 & 86.5 & 86.7 & 86.0 & 84.6 & 92.1 & 7.2 & 0.7 & 5.6 & 10.6\\
        \cmidrule(r){2-11}
        & 1 & 68.5 & 76.1 & 69.3 & 65.3 & 80.7 & 16.5 & 2.8 & 12.2 & 26.0 \\
        \textsc{TBA}   & 2 & 38.9 & 73.8 & 55.1 & 50.5 & 70.2 & 26.6 & 3.2 & 31.3 & 30.8\\
        & 3 & 56.0 & 75.0 & 64.3 & 59.2 & 76.7 & 19.8 & 3.5 & 20.8 & 27.5 \\
        \cmidrule(r){2-11}
        & 1 & 93.1 & 94.2 & 97.2 & 96.7 & 98.0 & 1.9 & 0.2 & 4.9 & 4.0 \\
        \textsc{OP3}   & 2 & 92.7 & 93.4 & 96.9 & 96.3 & 97.8 & 2.0 & 0.2 & 5.1 & 4.3 \\
        & 3 & 89.4 & 93.3 & 96.2 & 95.8 & 97.6 & 2.2 & 0.2 & 8.3 & 4.6 \\
        \cmidrule(r){2-11}
        & 1 & 75.7 & 88.1 & 69.4 & 68.3 & 79.8 & 19.4 & 0.8 & 4.0 & 13.9 \\
        \textsc{SCALOR} & 2 & 72.7 & 87.2 & 66.7 & 65.6 & 77.6 & 21.6 & 0.8 & 4.9 & 14.2 \\
        & 3 & 73.7 & 87.6 & 67.5 & 66.2 & 77.9 & 21.2 & 0.9 & 4.2 & 14.0 \\
        \midrule
        &\multicolumn{9}{c}{\textbf{texVMDS}} \\
        \cmidrule(r){2-11}
        & 1 & -99.5 & 76.4 & 25.4 & 24.3 & 30.3 & 69.2 & 0.5 & 129.8 & - \\
        \textsc{k-Means} & 2 & -99.5 & 76.4 & 25.3 & 24.4 & 30.4 & 69.2 & 0.5 & 129.9 & - \\
        & 3 & -99.5 & 76.4 & 25.3 & 24.3 & 30.3 & 69.2 & 0.5 & 129.8 & - \\
        \cmidrule(r){2-11}
        & 1 & -70.8 & 67.1 & 15.5 & 13.4 & 24.9 & 73.9 & 1.2 & 95.7 & 203.0 \\
        \textsc{MONet} & 2 & -68.1 & 69.2 & 20.4 & 15.4 & 30.4 & 66.5 & 3.1 & 98.6 & 192.5 \\
        & 3 & -80.9 & 66.7 & 12.1 & 8.1 & 18.9 & 78.9 & 2.2 & 99.8 & 205.9 \\
        \cmidrule(r){2-11}
        & 1 & -88.2 & 68.3 & 24.2 & 23.8 & 35.5 & 64.2 & 0.3 & 123.7 & 174.6 \\
        \textsc{ViMON} & 2 & -86.6 & 69.0 & 22.6 & 22.0 & 32.4 & 67.4 & 0.3 & 119.0 & 172.7 \\
        & 3 & -81.6 & 69.8 & 25.8 & 25.4 & 36.3 & 63.4 & 0.3 & 117.9 & 166.9 \\
        \cmidrule(r){2-11}
        & 1 & -105.6 & 71.5 & 20.6 & 20.4 & 28.3 & 71.6 & 0.2 & 133.9 & 113.6 \\
        \textsc{OP3} & 2 & -116.0 & 70.1 & 9.4 & 9.2 & 13.6 & 86.3 & 0.1 & 129.6 & 153.2 \\
        & 3 & -109.7 & 70.3 & 19.5 & 19.0 & 26.9 & 72.8 & 0.3 & 136.6 & 131.8 \\
        \cmidrule(r){2-11}
        & 1 & -84.8 & 73.3 & 6.4 & 6.3 & 12.5 & 87.3 & 0.2 & 97.3 & 149.5 \\
        \textsc{SCALOR} & 2 & -99.5 & 74.4 & 7.3 & 7.0 & 12.7 & 87.1 & 0.3 & 112.1 & 125.0 \\
        & 3 & -113.4 & 74.4 & 5.8 & 5.6 & 11.8 & 88.0 & 0.3 & 125.2 & 126.7 \\
        \bottomrule
    \end{tabular}}
    \caption{Analysis of SOTA object-centric representation learning models for MOT. Results for three runs with different random training seeds.}
    \label{tab:resultfulltwo}
\end{table}
}

\newcommand{\challengetable}{
    \begin{table}[t]
        \centering
        \resizebox{\textwidth}{!}{%
        \begin{tabular}{lllllllllllll}
            \toprule
            &
            \multicolumn{3}{c}{Occlusion}
            &
            \multicolumn{3}{c}{Same Color}
            &
            \multicolumn{3}{c}{Small Objects}
            &
            \multicolumn{3}{c}{Large Objects} \\
            \cmidrule(r){2-4}
            \cmidrule(r){5-7}
            \cmidrule(r){8-10}
            \cmidrule(r){11-13}
            Model  & MOTA & MOTP & MT & MOTA & MOTP & MT & MOTA & MOTP & MT & MOTA & MOTP & MT \\
            \midrule
            \textsc{ViMON} & 67.1 \pmsd{0.4} & 82.5 \pmsd{0.0} & 63.0 \pmsd{0.1} & 72.2 \pmsd{0.1} & 83.6 \pmsd{0.1} & 70.4 \pmsd{0.3} & 86.3 \pmsd{0.2} & 83.3 \pmsd{0.2} & 83.4 \pmsd{0.4} & 70.7 \pmsd{0.5} & 85.1 \pmsd{0.1} & 76.1 \pmsd{0.7} \\
            \textsc{TBA}   & 37.5 \pmsd{10.4} & 72.8 \pmsd{0.8} & 38.3 \pmsd{4.6} & 47.2 \pmsd{9.4} & 73.0 \pmsd{0.7} & 45.2 \pmsd{3.9} & 74.3 \pmsd{0.7} & 71.9 \pmsd{0.4} & 65.3 \pmsd{1.6} & 25.6 \pmsd{15.0} & 73.4 \pmsd{0.9} & 44.7 \pmsd{6.7} \\
            \textsc{OP3}   & 85.3 \pmsd{1.0} & 91.6 \pmsd{0.4} & 89.6 \pmsd{0.9} & 51.5 \pmsd{1.3} & 86.5 \pmsd{0.3} & 66.3 \pmsd{1.3} & 93.3 \pmsd{1.6} & 93.0 \pmsd{0.4} & 97.0 \pmsd{0.2} & 83.8 \pmsd{2.0} & 92.2 \pmsd{0.4} & 93.5 \pmsd{0.4} \\
            \textsc{SCALOR} & 58.8 \pmsd{1.0} & 86.6 \pmsd{0.4} & 46.8 \pmsd{1.2} & 53.7 \pmsd{1.1} & 83.4 \pmsd{0.3} & 46.2 \pmsd{1.1} & 74.4 \pmsd{0.7} & 86.1 \pmsd{0.4} & 67.6 \pmsd{1.3} & 66.1 \pmsd{1.9} & 86.6 \pmsd{0.5} & 62.4 \pmsd{1.4} \\   
            \bottomrule
        \end{tabular}}
        \includegraphics[width=\textwidth]{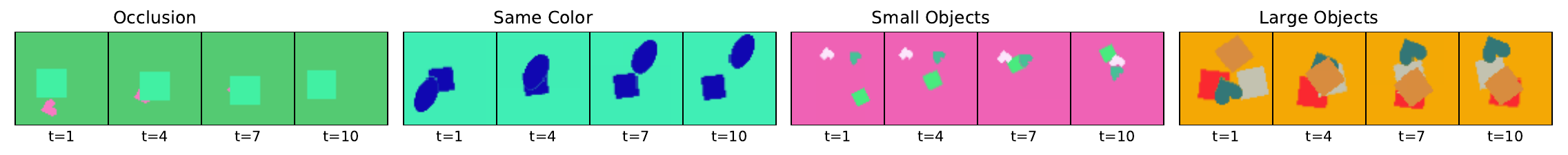}
        \caption{Performance on \textbf{VMDS challenge sets}. Results shown as mean $\pm$ standard deviation for three runs with different random training seeds. Examples sequences for each challenge set shown below.}
        \label{tab:challenge}
    \end{table}
}

\newcommand{\testtableood}{
    \begin{table}[t]
        \centering
        \resizebox{\textwidth}{!}{%
        \begin{tabular}{lrrrrrrrrrrrr}
            \toprule
            &
            \multicolumn{4}{c}{Size}
            & 
            \multicolumn{4}{c}{Color}
            &
            \multicolumn{4}{c}{Rotation} \\
            \cmidrule(r){2-5}
            \cmidrule(r){6-9}
            \cmidrule(r){10-13}
            Model  & MOTA & MOTP & MD & MT & MOTA & MOTP &  MD & MT & MOTA & MOTP & MD & MT \\
            \midrule
            \textsc{ViMON}  & 61.4 \pmsd{2.5} & 78.0 \pmsd{0.3} & 71.3 \pmsd{2.1} & 66.8 \pmsd{1.9} & 87.4 \pmsd{0.4} & 86.2 \pmsd{0.2} & 86.4 \pmsd{0.1} & 85.0 \pmsd{0.2} & -10.4 \pmsd{4.0} & 70.5 \pmsd{0.4} & 39.5 \pmsd{2.6} & 29.8 \pmsd{1.0} \\
            \textsc{ViMON}*  & 80.3 \pmsd{0.9} & 82.1 \pmsd{0.5} & 82.5 \pmsd{0.4} & 79.8 \pmsd{0.5} & 84.5 \pmsd{0.6} & 84.6 \pmsd{0.5} & 83.4 \pmsd{0.5} & 81.8 \pmsd{0.3} & 78.7 \pmsd{1.6} & 82.0 \pmsd{0.6} & 79.2 \pmsd{0.4} & 76.4 \pmsd{0.6} \\
            \textsc{TBA}    & 52.3 \pmsd{8.7} & 73.3 \pmsd{0.7} & 59.8 \pmsd{4.9} & 51.8 \pmsd{4.9} & 56.1 \pmsd{11.4} & 75.1 \pmsd{0.9} & 63.7 \pmsd{5.4} & 59.0 \pmsd{5.2} & 52.4 \pmsd{9.9} & 73.6 \pmsd{0.8} & 59.3 \pmsd{6.2} & 49.8 \pmsd{5.5} \\
            \textsc{TBA}* & 1.3 \pmsd{7.8} & 68.4 \pmsd{1.9} & 30.6 \pmsd{4.5} & 24.8 \pmsd{3.4} & -16.5 \pmsd{8.1} & 69.6 \pmsd{1.5} & 29.1 \pmsd{3.8} & 25.4 \pmsd{3.3} & -7.5 \pmsd{7.9} & 69.4 \pmsd{1.4} & 26.6 \pmsd{4.0} & 20.6 \pmsd{3.4} \\
            \textsc{OP3}  & 87.0 \pmsd{1.9} & 90.8 \pmsd{0.4} & 96.4 \pmsd{0.1} & 95.3 \pmsd{0.1} & 90.8 \pmsd{1.2} & 93.5 \pmsd{0.5} & 97.3 \pmsd{0.1} & 95.8 \pmsd{0.1} & 54.7 \pmsd{5.7} & 84.2 \pmsd{0.7} & 87.1 \pmsd{1.7} & 80.5 \pmsd{2.5} \\
            \textsc{OP3}* & 84.0 \pmsd{2.8} & 91.2 \pmsd{1.0} & 95.9 \pmsd{0.8} & 94.5 \pmsd{1.2} & 83.6 \pmsd{3.7} & 91.6 \pmsd{1.3} & 95.5 \pmsd{0.5} & 92.9 \pmsd{1.6} & 74.5 \pmsd{2.2} & 89.8 \pmsd{0.7} & 94.8 \pmsd{0.6} & 93.3 \pmsd{0.8} \\
            \textsc{SCALOR} & 68.1 \pmsd{1.7} & 84.9 \pmsd{0.4} & 63.3 \pmsd{1.7} & 60.0 \pmsd{2.0} & 75.5 \pmsd{1.1} & 89.9 \pmsd{0.5} & 67.0 \pmsd{1.4} & 65.7 \pmsd{1.6} & 46.5 \pmsd{1.8} & 82.1 \pmsd{0.5} & 41.9 \pmsd{1.7} & 37.1 \pmsd{1.3} \\
            \textsc{SCALOR}* & 67.5 \pmsd{1.2} & 85.2 \pmsd{0.6} & 61.2 \pmsd{1.2} & 57.1 \pmsd{0.7} & 73.3 \pmsd{0.7} & 89.8 \pmsd{0.5} & 64.8 \pmsd{1.1} & 63.0 \pmsd{0.9} & 61.6 \pmsd{1.4} & 83.5 \pmsd{0.4} & 53.4 \pmsd{1.5} & 50.2 \pmsd{1.1} \\
            \bottomrule
        \end{tabular}}
        \scriptsize{* Models trained on a data set that featured color, size and orientation changes of objects during the sequence.}
        \includegraphics[width=\textwidth]{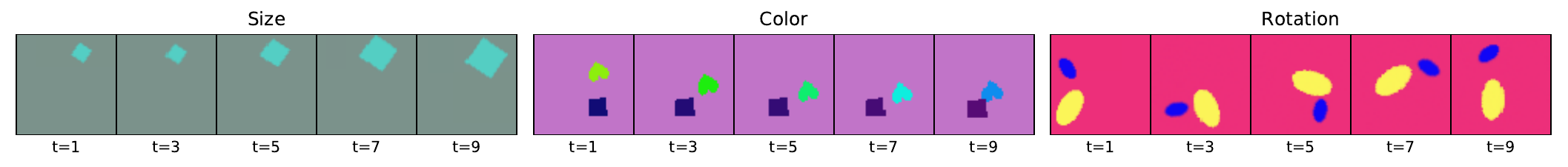}
        \caption{Performance on \textbf{VMDS OOD test sets}. Results shown as mean $\pm$ standard deviation for three runs with different random training seeds. Examples sequences for each o.o.d. set shown below.}
        \label{tab:ood}
    \end{table}
}

\newcommand{\resultstexvmds}{
\begin{table}[t!]
    \centering
    \resizebox{\textwidth}{!}{%
    \begin{tabular}{llllllllll}
        \toprule
        Model  & MOTA $\uparrow$ & MOTP $\uparrow$ & MD $\uparrow$ & MT $\uparrow$ & Match $\uparrow$ & Miss $\downarrow$ & ID S. $\downarrow$ & FPs $\downarrow$ & MSE $\downarrow$ \\
        \midrule
        &\multicolumn{9}{c}{\textbf{texVMDS}} \\
        \cmidrule(r){2-10}
        \textsc{k-Means} & -99.5 \pmsd{0.0} & \textbf{76.4 \pmsd{0.0}} & \textbf{25.3 \pmsd{0.0}} & \textbf{24.3 \pmsd{0.1}} & 30.3 \pmsd{0.0} & 69.2 \pmsd{0.0} & 0.5 \pmsd{0.0} & 129.8 \pmsd{0.0} &  -  \\
        \textsc{MONet} & \textbf{-73.3 \pmsd{5.5}} & 67.7 \pmsd{1.1} & 16.0 \pmsd{3.4} & 12.3 \pmsd{3.1} & 24.7 \pmsd{4.7} & 73.1 \pmsd{5.1} & 2.2 \pmsd{0.8} & \textbf{98.0 \pmsd{1.7}} & 200.5 \pmsd{5.7} \\ 
        \textsc{ViMON} & -85.5 \pmsd{2.8} & 69.0 \pmsd{0.6} & 24.2 \pmsd{1.3} & 23.8 \pmsd{1.4} & \textbf{34.7 \pmsd{1.7}} & \textbf{65.0 \pmsd{1.7}} & 0.3 \pmsd{0.0} & 120.2 \pmsd{2.5} & 171.4 \pmsd{3.3} \\
         \textsc{OP3}  & -110.4 \pmsd{4.3} & 70.6 \pmsd{0.6} & 16.5 \pmsd{5.1} & 16.2 \pmsd{5.0} & 22.9 \pmsd{6.6} & 76.9 \pmsd{6.7} & \textbf{0.2 \pmsd{0.1}} & 133.4 \pmsd{2.9} & \textbf{132.8 \pmsd{16.2}} \\
        \textsc{SCALOR} & -99.2 \pmsd{11.7} & 74.0 \pmsd{0.5} & 6.5 \pmsd{0.6} & 6.3 \pmsd{0.6} & 12.3 \pmsd{0.4} & 87.5 \pmsd{0.4} & \textbf{0.2 \pmsd{0.0}} & 111.5 \pmsd{11.4} & \textbf{133.7 \pmsd{11.1}} \\
        \bottomrule
    \end{tabular}}
    \includegraphics[width=\textwidth]{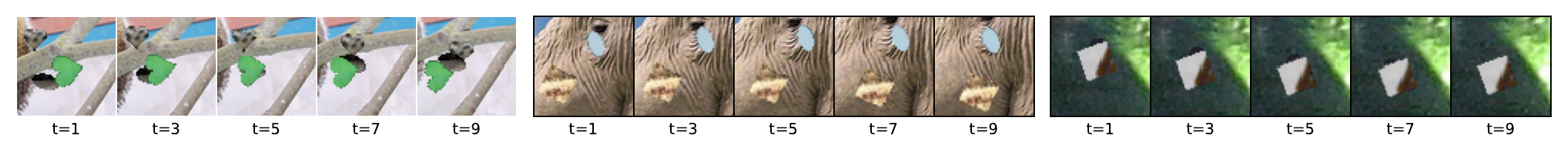}
    \caption{MOT performance on texVMDS and example video sequences. Results shown as mean $\pm$ standard deviation of three runs with different random training seeds. Note that MOTA can become negative, since the number of FPs is unbounded.}
    \label{tab:resultstexvmds}
\end{table}
}

\newcommand{\runtimetable}{
    \begin{table}[t]
        \centering
        \resizebox{\textwidth}{!}{%
        \begin{tabular}{llllllllll}
            \toprule
            & & & \multicolumn{4}{c}{Training}
            &
            \multicolumn{3}{c}{Inference} \\
            \cmidrule(r){4-7}
            \cmidrule(r){8-10}
            Model  & Resolution & No. Param. & Batch Size & Memory [MiB] & No. Iters & Epochs & Memory [MiB] & Avg. runtime / batch & Total runtime \\
            \midrule
            \textsc{ViMON} &  64$\times$64  & 714,900 & 18 & 10,860 & 3687 & 6.63 &  910 & 0.28 s/it & 4min 39s \\
            \textsc{TBA}   & 128$\times$128 & 3,884,644\textsuperscript{*} & 64 & 10,564 & 4421 & 28.29 & 972 & 0.24 s/it & 4min 05s  \\
            \textsc{OP3}   &  64$\times$64  & 876,305 & 10 & 10,874 & 2204 & 2.20 & 4092 & 0.54 s/it & 9min 04s \\
            \textsc{SCALOR}&  64$\times$64  & 2,763,526 & 48 & 10,942 & 2547 & 12.23 & 930 & 0.29 s/it & 4min 48s\\
            \bottomrule
        \end{tabular}}
        \scriptsize{* The TBA parameter count scales with the feature resolution, which is kept fixed using adaptive pooling. This makes the parameter count independent of input resolution.}
        \caption{Runtime analysis (using a single RTX 2080 Ti GPU). Training: models trained on VMDS for one hour. Inference: models evaluated on VMDS test set with batch size=1 (10 frames).}
        \label{tab:runtime}
    \end{table}
}

\newcommand{\vaeencodertable}{
    \begin{table}[t]
        \centering
        \begin{tabular}{llll}
            \toprule
            Type & Size/Ch. & Act. Func. & Comment \\
            \midrule
            Input & 4 & & RGB + Mask\\
            Conv 3x3 & 32 & ReLU & \\
            Conv 3x3 & 32 & ReLU & \\
            Conv 3x3 & 64 & ReLU & \\
            Conv 3x3 & 64 & ReLU & \\
            MLP & 256 & ReLU & \\
            MLP & 32 & Linear & \\
            \bottomrule
        \end{tabular}
        \caption{Architecture of \textsc{ViMON} \textbf{VAE Encoder}.}
        \label{tab:enc}
    \end{table}
}

\newcommand{\vaedecodertable}{
    \begin{table}[t]
        \centering
        \begin{tabular}{llll}
            \toprule
            Type & Size/Ch. & Act. Func. & Comment \\
            \midrule
            Input & 16 & & \\
            Broadcast & 18 &  & + coordinates \\
            Conv 3x3 & 32 & ReLU & \\
            Conv 3x3 & 32 & ReLU & \\
            Conv 3x3 & 32 & ReLU & \\
            Conv 3x3 & 32 & ReLU & \\
            Conv 1x1 & 4 & Linear & RGB + Mask \\
            \bottomrule
        \end{tabular}
        \caption{Architecture of \textsc{ViMON} \textbf{VAE Decoder}.}
        \label{tab:dec}
    \end{table}
}

\newcommand{\overviewfigure}{
\begin{figure}
\sbox\twosubbox{%
  \resizebox{\dimexpr\textwidth-1em}{!}{%
    \includegraphics[height=2cm]{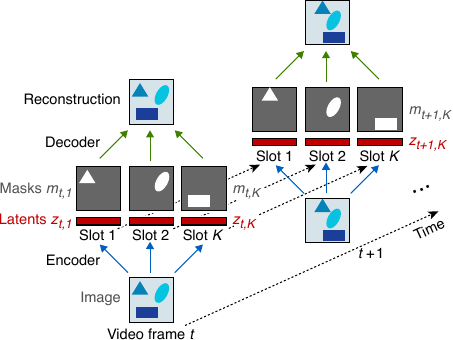}%
    \includegraphics[height=2cm]{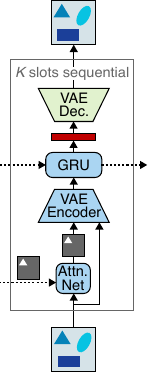}
    \includegraphics[height=2cm]{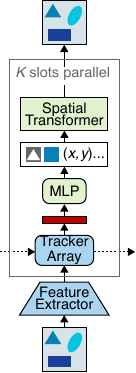}
    \includegraphics[height=2cm]{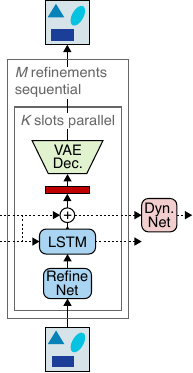}
    \includegraphics[height=2cm]{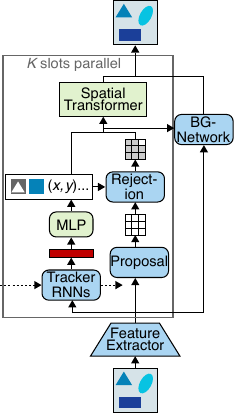}
  }%
}
\setlength{\twosubht}{\ht\twosubbox}
\centering
\subcaptionbox{Common Principle\label{fig:1a}}{%
  \includegraphics[height=\twosubht]{figures/models/architecture_a.pdf}%
}\hspace{0.25cm}
\subcaptionbox{ViMON\label{fig:1b}}{%
  \includegraphics[height=\twosubht]{figures/models/architecture_b.pdf}%
}\hfill
\subcaptionbox{TBA\label{fig:1c}}{%
  \includegraphics[height=\twosubht]{figures/models/architecture_c.pdf}
}\hfill
\subcaptionbox{OP3\label{fig:1d}}{%
  \includegraphics[height=\twosubht]{figures/models/architecture_d.pdf}
}\hfill
\subcaptionbox{SCALOR\label{fig:1e}}{%
  \includegraphics[height=\twosubht]{figures/models/architecture_e.pdf}
}
\caption{Common principles of all models: Decomposition of an image into a fixed number of slots, each of which contains an embedding $z_{t,k}$ and a mask $m_{t,k}$ of (ideally) a single object.  Dotted lines: temporal connections. Solid lines: information flow within one frame.}%
\label{fig:models}
\vspace{-10pt}
\end{figure}}

\newcommand{\ablationtable}{
\begin{table}[t!]
    \centering
    \resizebox{\textwidth}{!}{%
    \begin{tabular}{llllllllll}
        \toprule
        Model  & MOTA $\uparrow$ & MOTP $\uparrow$ & MD $\uparrow$ & MT $\uparrow$ & Match $\uparrow$ & Miss $\downarrow$ & ID S. $\downarrow$ & FPs $\downarrow$ & MSE $\downarrow$ \\
        \cmidrule(r){2-10}
        \textsc{ViMON w/o Mask conditioning} & 70.6 & 87.8 & 75.7 & 66.0 & 81.4 & 13.4 & 5.2 & 10.8 & 16.9 \\
        \textsc{ViMON w/o GRU} & 81.4 & 86.9 & 79.8 & 77.3 & 88.2 & 10.3 & 1.4 & 6.8 & 18.9 \\
        \bottomrule
    \end{tabular}}
    \caption{Ablation experiments for \textsc{ViMON} on VMDS.}
    \label{tab:abl}
\end{table}
}

\begin{document}

\title{Benchmarking Unsupervised Object Representations for Video Sequences}

\author{\name Marissa A.\ Weis\textsuperscript{1,4} \email marissa.weis@uni-goettingen.de
  \AND
  \name Kashyap Chitta\textsuperscript{3,6} \email kashyap.chitta@tue.mpg.de
  \AND
  \name Yash Sharma\textsuperscript{4} \email yash.sharma@uni-tuebingen.de
  \AND
  \name Wieland Brendel\textsuperscript{4,5} \email wieland.brendel@uni-tuebingen.de
  \AND
  \name Matthias Bethge\textsuperscript{4,5} \email matthias.bethge@uni-tuebingen.de
  \AND
  \name Andreas Geiger\textsuperscript{3,6} \email a.geiger@uni-tuebingen.de
  \AND
  \name Alexander S.\ Ecker\textsuperscript{1,2,7} \email ecker@cs.uni-goettingen.de \\
  \addr \textsuperscript{1}Institute of Computer Science, University of Göttingen, Germany \\
  \textsuperscript{2}Campus Institute Data Science, Göttingen, Germany \\
  \textsuperscript{3}Department of Computer Science, University of Tübingen, Germany \\
  \textsuperscript{4}Institute for Theoretical Physics, University of Tübingen, Germany \\
  \textsuperscript{5}Bernstein Center for Computational Neuroscience, Tübingen, Germany \\
  \textsuperscript{6}Max Planck Institute for Intelligent Systems, Tübingen, Germany \\
  \textsuperscript{7}Max Planck Institute for Dynamics and Self-Organization, Göttingen, Germany \\
  }
  
\maketitle

\begin{abstract}%

Perceiving the world in terms of objects and tracking them through time is a crucial prerequisite for reasoning and scene understanding. Recently, several methods have been proposed for unsupervised learning of object-centric representations. However, since these models were evaluated on different downstream tasks, it remains unclear how they compare in terms of basic perceptual abilities such as detection, figure-ground segmentation and tracking of objects. 
To close this gap, we design a benchmark with four data sets of varying complexity and seven additional test sets featuring challenging tracking scenarios relevant for natural videos.
Using this benchmark, we compare the perceptual abilities of four object-centric approaches: \textsc{ViMON}, a video-extension of \textsc{MONet}, based on recurrent spatial attention, \textsc{OP3}, which exploits clustering via spatial mixture models, as well as \textsc{TBA} and \textsc{SCALOR}, which use explicit factorization via spatial transformers. Our results suggest that the architectures with unconstrained latent representations learn more powerful representations in terms of object detection, segmentation and tracking than the spatial transformer based architectures. We also observe that none of the methods are able to gracefully handle the most challenging tracking scenarios despite their synthetic nature, suggesting that our benchmark may provide fruitful guidance towards learning more robust object-centric video representations.

\end{abstract}

\section{Introduction}

Humans understand the world in terms of objects \citep{Marr82,Spelke07,Johnson18}. Being able to decompose our environment into independent objects that can interact with each other is an important prerequisite for reasoning and scene understanding \citep{Spelke88}. Similarly, an artificial intelligence system would benefit from the ability to represent visual scenes in a structured way \citep{Ullman17,Lake17,Greff20} by both extracting objects and their interactions from video streams, and keeping track of them over time. 

Recently, there has been an increased interest in \emph{unsupervised} learning of \emph{object-centric representations}. The key insight of these methods is that the compositionality of visual scenes can be used to both discover~\citep{Eslami16, Greff19, Burgess19} and track objects in videos~\citep{Greff17, Steenkiste18,Veerapaneni19} without supervision. 
However, it is currently not well understood how the learned visual representations of different models compare to each other quantitatively, since the models have been developed with different downstream tasks in mind and have not been evaluated using a common protocol. Hence, in this work, we propose a benchmark based on procedurally generated video sequences to test basic perceptual abilities of object-centric video models under various challenging tracking scenarios.

An unsupervised object-based video representation should \textit{(1)} effectively identify objects as they enter a scene, \textit{(2)} accurately segment objects, as well as \textit{(3)} maintain a consistent representation for each individual object in a scene over time. These perceptual abilities can be evaluated quantitatively in the established \emph{multi-object tracking} framework \citep{Bernardin08, Milan16}. We propose to utilize this protocol for analyzing the strengths and weaknesses of different object-centric representation learning methods, independent of any specific downstream task, in order to uncover the different inductive biases hidden in their choice of architecture and loss formulation. We therefore compiled a benchmark consisting of four procedurally generated video data sets of varying levels of visual complexity and two generalization tests. Using this benchmark, we quantitatively compared three classes of object-centric models, leading to the following insights:

\begin{itemize}[nosep]
    \item All of the models have shortcomings handling occlusion, albeit to different extents.
    \item OP3~\citep{Veerapaneni19} performs strongest in terms of quantitative metrics, but exhibits a surprisingly strong dependency on color to separate objects and accumulates false positives when fewer objects than slots are present.
    \item Spatial transformer models, \textsc{TBA}~\citep{He19} and \textsc{SCALOR}~\citep{Jiang19}, train most efficiently and feature explicit depth reasoning in combination with amodal masks, but are nevertheless outperformed by the simpler model, \textsc{ViMON}, lacking a depth or interaction model, suggesting that the proposed mechanisms may not yet work as intended.
\end{itemize}

Our code, data, as well as a public leaderboard of results is available at \url{https://eckerlab.org/code/weis2020/}.
\section{Related work}

Several recent lines of work propose to learn object-centric representations from visual inputs for static and dynamic scenes without explicit supervision. Though their results are promising, methods are currently restricted to handling synthetic data sets and as of yet are unable to scale to complex natural scenes. Furthermore, a systematic quantitative comparison of methods is lacking. 

Selecting and processing parts of an image via \textit{spatial attention} has been one prominent approach for this task~\citep{Mnih14, Eslami16, Kosiorek18, Burgess19, Yuan19, Crawford19}. As an alternative, \textit{spatial mixture} models decompose scenes by performing image-space clustering of pixels that belong to individual objects~\citep{Greff16,Greff17,Greff19,Steenkiste18, Locatello20}. 
\citet{Xu19} use optic flow to learn to segment objects or object parts in dynamic scenes.
While some approaches aim at learning a suitable representation for downstream tasks \citep{Watters19a,Veerapaneni19}, others target scene generation \citep{Engelcke19,Kuegelgen20, Ehrhardt20, Jiang20}. We analyze three classes of models for processing videos, covering three models based on spatial attention and one based on spatial mixture modeling. 

\textbf{Spatial attention models with unconstrained latent representations} use per-object variational autoencoders, as introduced by \citet{Burgess19}. \citet{Kuegelgen20} adapt this approach for scene generation. So far, such methods have been designed for static images, but not for videos. We therefore extend \textsc{MONet} \citep{Burgess19} to be able to accumulate evidence over time for tracking, enabling us to include this class of approaches in our evaluation. 
Recent concurrent work on AlignNet \citep{Creswell20} applies \textsc{MONet} frame-by-frame and tracks objects by subsequently ordering the extracted objects consistently. 

\textbf{Spatial attention models with factored latents} use an explicit factorization of the latent representation into properties such as position, scale and appearance~\citep{Eslami16, Crawford19}. These methods use spatial transformer networks~\citep{Jaderberg15} to render per-object reconstructions from the factored latents~\citep{Kosiorek18,He19,Jiang19}. \textsc{SQAIR} \citep{Kosiorek18} does not perform segmentation, identifying objects only at the bounding-box level. \citet{Henderson20} explicitly reason about the underlying 3D scene structure.
We select Tracking-by-Animation (\textsc{TBA})~\citep{He19} and \textsc{SCALOR}~\citep{Jiang19} for analyzing spatial transformer methods in our experiments, which explicitly disentangle object shape and appearance, providing access to object masks.

\textbf{Spatial mixture models} cluster pixels using a deep neural network trained with expectation maximization~\citep{Greff17,Steenkiste18}. \textsc{IODINE}~\citep{Greff19} extends these methods with an iterative amortised variational inference procedure~\citep{Marino18}, improving segmentation quality. \textsc{SPACE}~\citep{Lin20} combines mixture models with spatial attention to improve scalability. %
To work with video sequences, \textsc{OP3}~\citep{Veerapaneni19} extends \textsc{IODINE} by modeling individual objects' dynamics as well as pairwise interactions. We therefore include \textsc{OP3} in our analysis as a representative spatial mixture model.

\datatable

\section{Object-Centric Representation Benchmark}
To compare the different object-centric representation learning models on their basic perceptual abilities, we use the well-established multi-object tracking (MOT) protocol \citep{Bernardin08}. In this section, we describe the data sets and metrics considered in our benchmark, followed by a brief description of the models evaluated. %

\subsection{Data Sets}
Current object-centric models are not capable of modeling complex natural scenes \citep{Burgess19, Greff19, Lin20}. Hence, we focus on synthetic data sets that resemble those which state-of-the-art models were designed for. Specifically, we evaluate on four synthetic data sets\footnote{\href{https://osf.io/ua6sk/?view_only=70080c40f5e6467d90b83b2eb1f41907}{Data sets are available at this https URL}.} (see \tabref{tab:data}), which cover multiple levels of visual and motion complexity. Synthetic stimuli enable us to precisely generate challenging scenarios in a controllable manner in order to disentangle sources of difficulty and glean insights on what models specifically struggle with. We design different scenarios that test complexities that would occur in natural videos such as partial or complete occlusion as well as similar object appearances and complex textures.

\textbf{Sprites-MOT} (\textbf{SpMOT}, \tabref{tab:data} left), as proposed by \citet{He19}, features simple 2D sprites moving linearly on a black background with objects moving in and out of frame during the sequence. \textbf{Video-Multi-dSprites} (\textbf{VMDS}, \tabref{tab:data} right) is a video data set we generated based on a colored, multi-object version of the dSprites data set~\citep{Matthey17}. Each video contains one to four sprites that move non-linearly and independently of each other with the possibility of partial or full occlusion. Besides the i.i.d. sampled training, validation and test sets of VMDS, we generate seven additional challenge sets that we use to study specific test situations we observed to be challenging, such as guaranteed occlusion, specific object properties, or out-of-distribution appearance variations.  
\textbf{Video Objects Room} (\textbf{VOR}, \tabref{tab:data} middle) is a video data set we generated based on the static Objects Room data set~\citep{Burgess19}, which features static objects in a 3D room with a moving camera. 
\textbf{Textured Video-Multi-dSprites} (\textbf{texVMDS}, \tabref{tab:resultstexvmds}) is based on the generative model of VMDS regarding shapes, number and movement dynamics of the objects, but instead of being uniformly colored the objects feature random texture crops taken from ImageNet \citep{Deng09}. Additionally, the background is a linearly moving ImageNet crop.
For full details on the data sets and their generation, see Appendix~\ref{datasets_suppl}.

\subsection{Metrics} 
Our evaluation protocol follows the multi-object tracking (MOT) challenge, a standard and widely-used benchmark for supervised object tracking \citep{Milan16}. The MOT challenge uses the CLEAR MOT metrics \citep{Bernardin08}, which quantitatively evaluate different performance aspects of object detection, tracking and segmentation. To compute these metrics, predictions have to be matched to ground truth. Unlike \citet{Bernardin08} and \citet{Milan16}, we use binary segmentation masks for this matching instead of bounding boxes, which helps us better understand the models' segmentation capabilities. We consider an intersection over union (IoU) greater than 0.5 as a match \citep{Voigtlaender2019}. We include an ablation experiment to test how this threshold affects the results. The error metrics used are the fraction of \textbf{Misses (Miss)}, \textbf{ID switches (ID S.)} and \textbf{False Positives (FPs)} relative to the number of ground truth masks. In addition, we report the \textbf{Mean Squared Error (MSE)} of the reconstructed image outputs summed over image channels and pixels.

To quantify the overall number of failures, we use the \textbf{MOT Accuracy (MOTA)}, which measures the fraction of all failure cases compared to the total number of objects present in all frames. A model with 100\% MOTA perfectly tracks all objects without any misses, ID switches or false positives. To quantify the segmentation quality, we define \textbf{MOT Precision (MOTP)} as the average IoU of segmentation masks of all matches. A model with 100\% MOTP perfectly segments all tracked objects, but does not necessarily track all objects. Further, to quantify detection and tracking performance independent of false positives, we measure the \textbf{Mostly Detected (MD)} and \textbf{Mostly Tracked (MT)} metrics, the fraction of ground truth objects that have been detected and tracked for at least 80\% of their lifespan, respectively. If an ID switch occurs, an object is considered detected but not tracked. For full details regarding the matching process and the evaluation metrics, refer to Appendix~\ref{metrics_suppl}.

\subsection{Models}\label{method}

We consider 
one color-segmentation baseline and 
three classes of unsupervised object-centric representation learning models:
\textit{(1)} a spatial attention model with unconstrained latents, \textsc{ViMON}, which is our video extension of \textsc{MONet}~\citep{Burgess19}; \textit{(2)} spatial transformer-based attention models, \textsc{TBA}~\citep{He19} and \textsc{SCALOR}~\citep{Jiang19}; \textit{(3)} a scene mixture model, \textsc{OP3}~\citep{Veerapaneni19}. At a high-level, these methods share a common structure which is illustrated in \figref{fig:1a}. They decompose an image into a fixed number of \emph{slots}~\citep{Burgess19}, each of which contains an embedding $\bz_{t,k}$ and a mask $\bm_{t,k}$ of (ideally) a single object. These slots are then combined in a decoding step to reconstruct the image. Below, we briefly describe each method. Appendix~\ref{method_suppl} provides a detailed explanation of the methods in a unified mathematical framework.

\overviewfigure

As a simple baseline, we evaluate the  performance of \textbf{k-Means} clustering on the RGB pixel values on a per frame basis and tracking performance by frame-to-frame matching based on the IoU of the segmentation masks using Hungarian matching \citep{Kuhn55}. For full details refer to Appendix~\ref{implem_kmeans}.

\textbf{Video MONet (\textsc{ViMON})}
is our video extension of \textsc{MONet}~\citep{Burgess19}. \textsc{MONet} recurrently decomposes a static scene into slots, using an attention network to sequentially extract attention masks~$\bm_k \in [0,1]^{H \times W}$ of individual objects $k$. A Variational Autoencoder (VAE) \citep{Kingma14} encodes each slot into a latent representation $\bz_k \in \nR^L$ of the corresponding object. We use \textsc{MONet} as a simple frame-by-frame baseline for detection and segmentation that does not employ temporal information.
\textsc{ViMON} accumulates evidence about the objects 
over time to maintain a consistent object-slot assignment throughout the video. This is achieved by \textit{(1)} seeding the attention network the predicted mask $\widehat{\bm}_{t,k} \in [0,1]^{H \times W}$ 
from the previous time step and \textit{(2)} introducing a gated recurrent unit (GRU) \citep{Cho14}, which aggregates information over time for each slot separately, enabling it to encode motion information.
For full details on \textsc{MONet} and \textsc{ViMON}, as well as information regarding hyperparameter tuning and ablations to provide context for the design decisions, refer to Appendix~\ref{method_monet}, \ref{method_vimon} and \ref{vimon_abl}.

\textbf{Tracking-by-Animation (\textsc{TBA})}~\citep{He19} is a spatial transformer-based attention model. Frames are encoded by a convolutional feature extractor $f$ before being passed to a recurrent block~$g$ called Reprioritized Attentive Tracking (RAT). %
RAT re-weights slot input features based on their cosine similarity with the slots from the previous time step and outputs latent representations for all $K$ slots in parallel.
Each slot latent is further decoded into a mid-level representation $\by_{t,k}$ consisting of pose and depth parameters, as well as object appearance and shape templates (see \figref{fig:1c}). 
For rendering, a Spatial Transformer Network (STN) \citep{Jaderberg15} is used with an additional occlusion check based on the depth estimate. \textsc{TBA} is trained on frame reconstruction with an additional penalty for large object sizes to encourage compact bounding boxes. \textsc{TBA} can only process scenes with static backgrounds, as it preprocesses sequences using background subtraction \citep{Bloisi12}.
For full details on \textsc{TBA} as well as information regarding hyperparameter tuning, refer to Appendix~\ref{method_tba}.

\textbf{Object-centric Perception, Prediction, and Planning (\textsc{OP3})}~\citep{Veerapaneni19}
extends \textsc{IODINE}~\citep{Greff19} to operate on videos.
\textsc{IODINE} decomposes an image into objects and represents them independently by starting from an initial guess of the segmentation of the entire frame, and subsequently iteratively refines it using the refinement network $f$~\citep{Marino18}. In each refinement step $m$, the image 
is represented by $K$ slots with latent representations 
$\bz_{m,k}$.
\textsc{OP3} applies \textsc{IODINE} to each frame $x_t$ to extract latent representations $\bz_{t,m,k}$, which are subsequently processed by a dynamics network $d$ (see \figref{fig:1d}), which models both the individual dynamics of each slot~$k$ as well as the pairwise interaction between all combinations of slots, aggregating them into a prediction of the posterior parameters for the next time step~$t+1$ for each slot~$k$. 
For full details on \textsc{IODINE} and \textsc{OP3}, refer to Appendix~\ref{method_iodine} and~\ref{method_op3}, respectively.

\textbf{SCALable Object-oriented Representation (\textsc{SCALOR})}~\citep{Jiang19} is a spatial transformer-based model that factors scenes into background and multiple foreground objects, which are tracked throughout the sequence. 
Frames are encoded using a convolutional LSTM~$f$. In the proposal-rejection phase, the current frame $t$ is divided into $H \times W$ grid cells. For each grid cell a object latent variable $\bz_{t,h,w}$ is proposed, that is 
factored into existence, pose, depth and appearance parameters. 
Subsequently, proposed objects that significantly overlap 
with a propagated object are rejected. In the propagation phase, per object GRUs
are updated for all objects present in the scene.
Additionally, \textsc{SCALOR} has a background module to encode the background and its dynamics. Frame reconstructions are rendered using a background decoder and foreground STNs for object masks and appearance. %
For full details on \textsc{SCALOR} as well as information regarding hyperparameter tuning, refer to Appendix~\ref{method_scalor}.

\section{Results}
We start with a summary of our overall results across the three data sets and four models (\tabref{tab:result}) before analyzing more specific challenging scenarios using variants of the VMDS data set.

\resultstable

A simple color segmentation algorithm is not sufficient to solve tracking even in simplistic data sets (\tabref{tab:result}), in which color is a good predictor for object identity. While \textsc{k-Means} does well at detecting objects, on VMDS even outperforming the other models, its overall tracking performance is significantly worse than the other models, mainly do to a high number of false positive and in SpMOT and VOR also due to a high number of misses, which occur in these two data sets due to multiple objects often having the same color, which a color segmentation algorithm is by definition not able to separate. In the following, we will therefore focus on analyzing the object-centric methods.

We first ask whether tracking could emerge automatically in an image-based model like \textsc{MONet}, which may produce consistent slot assignments through its learned object-slot assignment.
This is not the case: \textsc{MONet} exhibits poor tracking performance (\tabref{tab:result}). While \textsc{MONet} correctly finds and segments objects, it does not assign them to consistent slots over time (\figref{fig:vamo_vimo}). 
In the following, we will thus focus on the video models: \textsc{ViMON}, \textsc{TBA}, \textsc{OP3} and \textsc{SCALOR}. 

\textbf{SpMOT}. 
All models perform tracking well on SpMOT with the exception of one training run of \textsc{TBA} with poor results leading to high standard deviation (cp. best \textsc{TBA} model: 89.8\% MT; \tabref{tab:resultfull}). \textsc{SCALOR} outperforms the other models
on the detection and tracking metrics MD and MT, while \textsc{ViMON} exhibits the highest MOTP, highlighting its better segmentation performance on SpMOT. 

\textbf{VOR}. \textsc{TBA} is not applicable to VOR due to the dynamic background which cannot be resolved using background subtraction. \textsc{ViMON} and \textsc{OP3} show similarly good performance on detection (MD) and segmentation (MOTP), while \textsc{ViMON} outperforms \textsc{OP3} on the tracking metrics MOTA and MT.
\textsc{OP3} accumulates a high number of false positives leading to a low MOTA due to the splitting of objects into multiple masks as well as randomly segmenting small parts of the background (\figref{fig:op3_vor_fail}). In contrast, \textsc{SCALOR} has almost no false positives or ID switches, but accumulates a high number of misses leading to a poor MOTA. It often segments two objects as one that are occluding each other in the first frame, which is common in VOR due to the geometry of its scenes (\figref{scalor_vor_rand}, last row).

\textbf{VMDS}. \textsc{OP3} outperforms the other models on VMDS, on which \textsc{TBA} performs poorly, followed by \textsc{SCALOR}, which again accumulates a high number of misses. We will analyze the models on VMDS qualitatively and quantitatively in more detail in the following.

\begin{figure}[t]
\begin{subfigure}[t]{0.46\textwidth}
    \includegraphics[width=\textwidth]{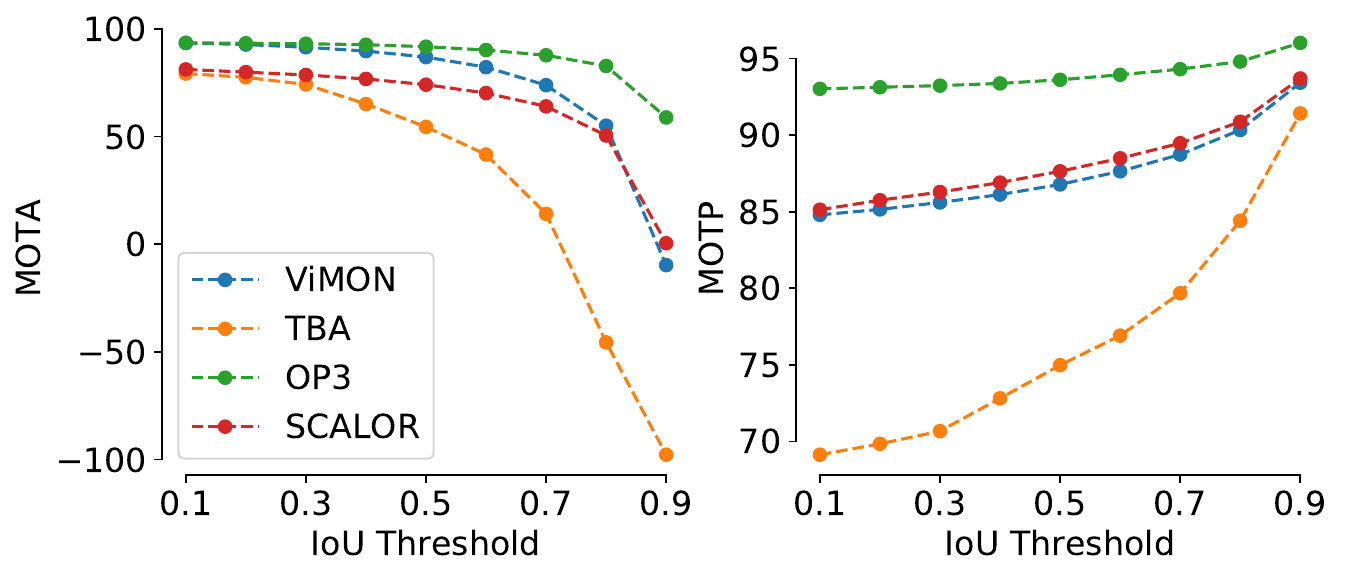}
    \caption{}
    \label{fig:iou}
\end{subfigure}
\hspace{5pt}
\begin{subfigure}[t]{0.24\textwidth}
    \includegraphics[width=\textwidth]{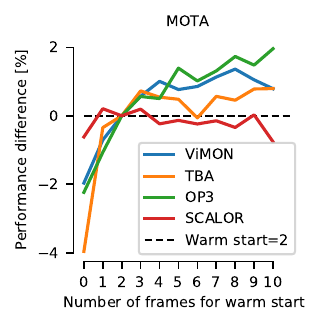}
    \caption{}
    \label{fig:temp}
\end{subfigure}
\hspace{5pt}
\begin{subfigure}[t]{0.24\textwidth}
    \includegraphics[width=\textwidth]{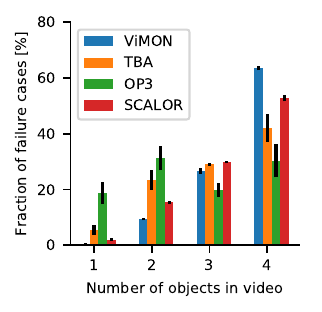}
    \caption{}
    \label{fig:fail_num_obj}
\end{subfigure}
\caption{a) MOTA and MOTP performance of models shown for different IoU threshold used for matching model predictions to ground truth objects. Note that MOTA can become negative, since the number of FPs is unbounded. b) MOTA on frames 11--20 of the VMDS test set with warm starts of 1--10 frames (0 = no warm start). Difference to performance of warm start = 2 shown. c) Distribution of failure cases dependent on number of objects in VMDS videos. Mean of three training runs. Error bars: standard deviation.}
\end{figure}

\subsection{Impact of IoU Threshold for Matching}
To examine whether the IoU threshold used for matching object predictions to ground truth impacts the ordering of the models, we compute all metrics with regard to the IoU thresholds [0.1 .. 0.9] in steps of 0.1 on the VMDS test set. The ordering of the models with regard to their MOTA is consistent over all matching thresholds except for a threshold of 0.9, but the performance difference between the models increases with higher thresholds (\figref{fig:iou}). Thus, we will perform all following experiments with an IoU threshold of 0.5. For all metrics, see \figref{fig:iou_thresh} in Appendix.

\subsection{Accumulation of Evidence over Time}
Recognition and tracking of objects should improve if models can exploit prior knowledge about the objects in the scene from previous video frames. To test whether the models exploit such knowledge, we evaluate their MOTA performance on VMDS after warm starting with up to 10 frames which are not included in evaluation (\figref{fig:temp}). Note that the models were trained on sequences of length 10, but are run for 20 frames in the case of a warm start of 10 frames.
The performance of \textsc{ViMON} improves with longer warm starts, showing that the GRU accumulates evidence over time. \textsc{TBA}, in contrast, does not use temporal information beyond 2--3 frames, while \textsc{SCALOR}'s performance slightly drops after 3 frames.  \textsc{OP3} appears to most strongly rely on past information and is able to integrate information over longer time scales: its performance does not even saturate with a warm start of 10 frames. However, the effect for all models is rather small.

\subsection{Challenging Scenarios for Different Models}
\emph{The number of objects in the sequence} matters for \textsc{ViMON}, \textsc{TBA} and \textsc{SCALOR}: more objects increase the number of failure cases (\figref{fig:fail_num_obj}). In contrast, \textsc{OP3} does not exhibit this pattern: it accumulates a higher number of false positives (FPs) in videos with fewer (only one or two) objects (\figref{fig:fail_num_obj_cl}), as it tends to split objects into multiple slots if fewer objects than slots are present.

\begin{figure}[t!]
    \centering
    \includegraphics[width=0.9\textwidth]{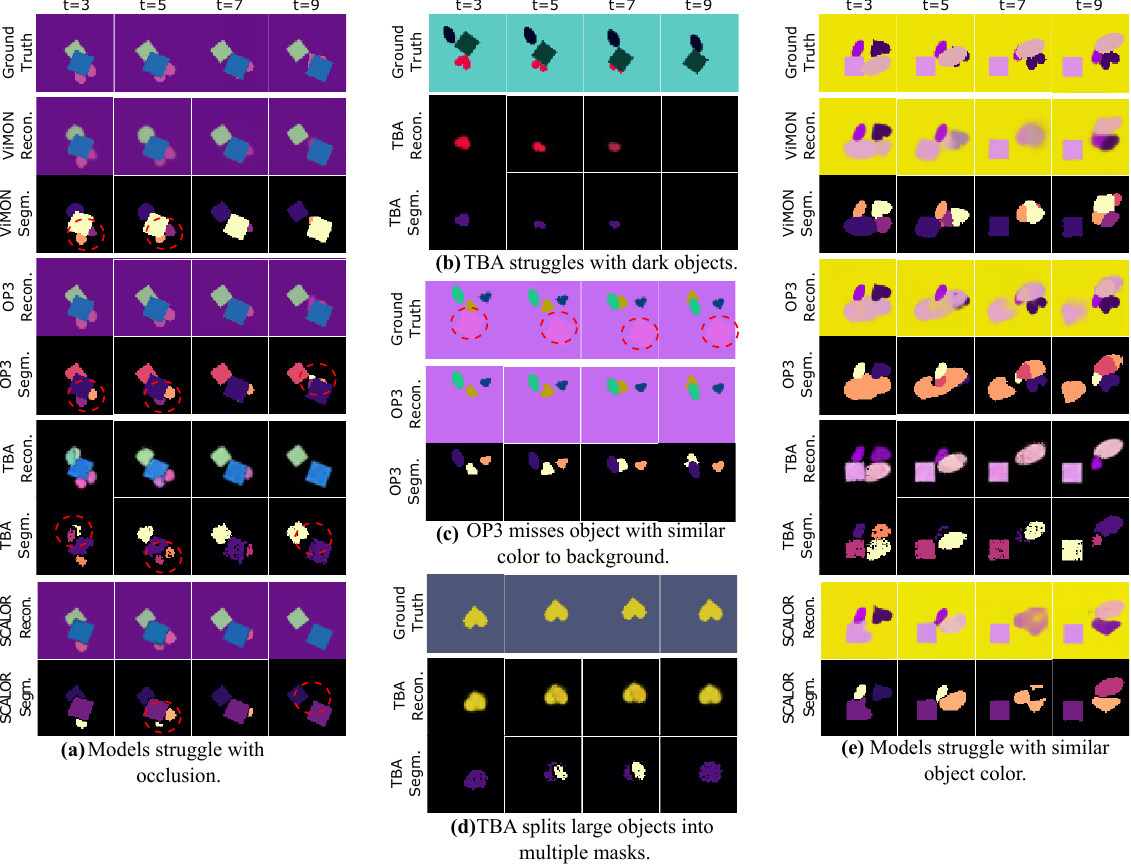}
    \caption{Example failure cases for all models on VMDS. Segmentation masks are binarized and color-coded to signify slot assignment.}
    \label{fig:fail_vis}
    \vspace{-10pt}
\end{figure}

\emph{Occlusion} leads to failure cases for all models (\figref{fig:fail_vis}a--b). 
Partial occlusion can lead to splitting of objects into multiple slots (\figref{fig:fail_vis}a). Objects that reappear after full occlusion are often missed when only a small part of them is visible (\figref{fig:fail_vis}a). In particular, \textsc{SCALOR} tends to segment two objects as one when they overlap while entering the scene, leading to a high number of misses.

\emph{Color} of the object is important. \textsc{TBA} often misses dark objects (\figref{fig:fail_vis}b).
In contrast, \textsc{ViMON}, \textsc{OP3} and \textsc{SCALOR} struggle with scenes that feature objects of similar colors as well as objects that have similar colors to the background (\figref{fig:fail_vis}c,e).

\emph{False positives} are more prevalent for \textsc{OP3} and \textsc{TBA} than for \textsc{ViMON} and \textsc{SCALOR} (\tabref{tab:result}). FPs of \textsc{OP3} are due to objects split in multiple masks (\figref{fig:fail_vis}a) and random small parts of the background being individually segmented (\figref{fig:fail_vis}e), while \textsc{TBA} tends to compose larger objects using multiple smaller, simpler components (\figref{fig:fail_vis}d).

\begin{figure}[t]
\begin{subfigure}[t]{0.5\textwidth}
    \includegraphics[width=\textwidth]{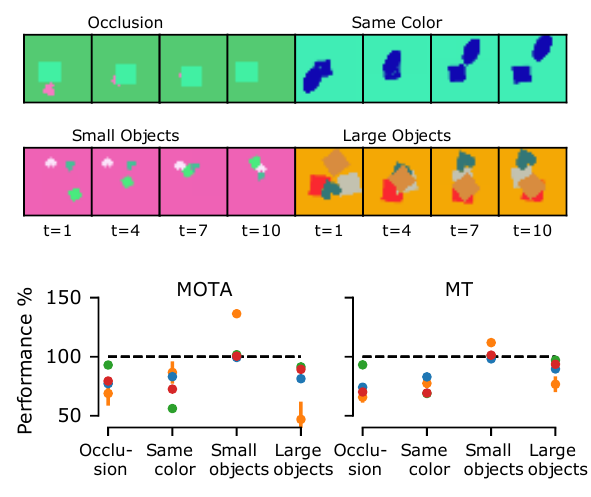}
    \caption{}
    \label{fig:chall}
\end{subfigure}
\hspace{5pt}
\begin{subfigure}[t]{0.49\textwidth}
    \includegraphics[width=\textwidth]{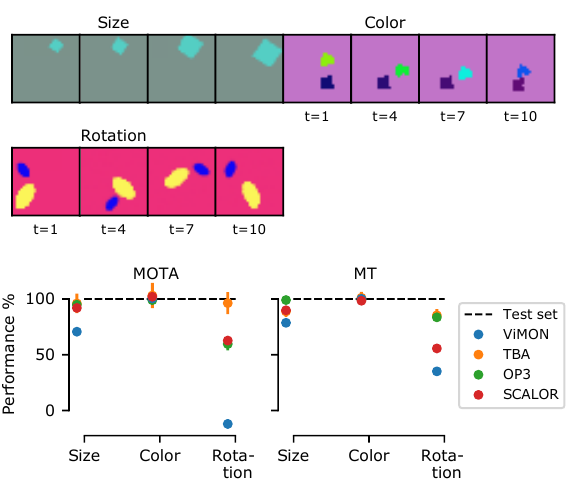}
    \caption{}
    \label{fig:ood}
\end{subfigure}
\caption{a) Performance on challenge sets relative to performance on VMDS test set (100\%) and example video sequences. b) Performance on out-of-distribution sets relative to VMDS test set (100\%) and example video sequences. Note that MOTA can become negative, since the number of FPs is unbounded.}
\end{figure}

\subsection{Performance on Challenge Sets}
Based on the challenging scenarios identified above, we design multiple `challenge sets': videos featuring \emph{(1)} heavy occlusion, \emph{(2)} objects with same colors, \emph{(3)} only small objects and \emph{(4)} only large objects (\figref{fig:chall}, top). 
For details, see Appendix \ref{vmds_chall_suppl}. 

Occlusion reduces performance of all models compared with the i.i.d. sampled VMDS test set, albeit to different degrees (\figref{fig:chall}; for absolute performance see \tabref{tab:challenge}). \textsc{OP3} is more robust to occlusion than the other models.

Tracking objects with the same color is challenging for all models (\figref{fig:chall}). In particular, \textsc{OP3} appears to rely on object color as a way to separate objects.

\textsc{OP3}, \textsc{ViMON} and \textsc{SCALOR} are not sensitive to object size (\figref{fig:chall}). They exhibit only slightly decreased performance on the large objects test set, presumably because large objects cause more occlusion (\figref{fig:chall}). 
\textsc{TBA} shows increased performance on small objects but performs poorly on the large objects set.

\subsection{Performance on Out-of-distribution Test Sets}
Next, we assess generalization to out-of-distribution (o.o.d.) changes in object appearance that are not encountered during training. In the training set of VMDS, object color, size and orientation are constant throughout a video. To test o.o.d. generalization, we evaluate models trained on VMDS on three data sets that feature unseen object transformations (\figref{fig:ood} and \tabref{tab:ood}): continuous changes in object \textbf{color} or \textbf{size} as well as continuous \textbf{rotation} around the object's centroid while moving. For details, see Appendix \ref{vmds_ood_suppl}. 

Continuous changes in object size do not pose a serious problem to \textsc{TBA}, \textsc{OP3} and \textsc{SCALOR}, while \textsc{ViMON}'s performance drops (\figref{fig:ood}). Surprisingly, continuous color changes of objects do not impact the performance of any model. Tracking performance of \textsc{ViMON} drops significantly for rotated objects, while \textsc{OP3} and \textsc{SCALOR} are affected less. \textsc{TBA}'s tracking performance is not as strongly influenced by object rotation (for absolute values, see \tabref{tab:ood}).

\subsection{Stability of Training and Runtime} \label{runtime}
\textsc{TBA} and \textsc{SCALOR} train faster and require less memory than \textsc{OP3} and \textsc{ViMON} (see \tabref{tab:runtime} for details). However, two training runs converged to suboptimal minima for \textsc{TBA} (note the high variance in \tabref{tab:result} and see \tabref{tab:resultfull} and \tabref{tab:resultfulltwo} for individual runs). 
Training \textsc{OP3} is sensitive to the learning rate and unstable, eventually diverging in all experiments. Interestingly, it often reached its best performance directly prior to divergence.
\textsc{ViMON} and \textsc{TBA} are less sensitive to hyper-parameter settings in our experiments.
For a more detailed analysis of the runtime, see Appendix \ref{runtime_appendix}.

\subsection{Towards natural data}
Object-centric methods have mostly been shown to work on synthetic data sets and seem to rely to a certain extent on color segmentation. To quantify how these methods fare with more realistic images, we created a fourth data set that keeps the simple scene composition of VMDS (1--4 object and few shapes), but has a higher level of visual complexity. To this end, we generated moving object masks as in VMDS, but filled each object and the background with the content of a different image taken from ImageNet (\tabref{tab:resultstexvmds}, bottom). We refer to this data set as \emph{texVMDS}. Humans can easily detect objects in these video through motion cues,\footnote{See \url{https://eckerlab.org/code/weis2020/} for example videos.} while the evaluated models fail at detecting and tracking objects in this data set (\tabref{tab:resultstexvmds}).

Most of the statistics of the objects in texVMDS are identical to those of VMDS (shape, size, velocity, number of objects); the main difference is the visual complexity of the object and background texture. We therefore tune the dimensionality of the latent space for \textsc{MONet} and \textsc{SCALOR}, to account for the higher visual complexity, while keeping all other hyperparameters identical to VMDS. \textsc{OP3} already has a significantly larger latent dimensionality and uses that for synthethic as well as natural data in their publication. Therefore we keep these hyperparameters. For full details see Appendix \ref{implem}.

\textsc{TBA} is not applicable to texVMDS due to the dynamic background which cannot be resolved using background subtraction. All other models perform poorly. They fail to learn the object boundaries despite the fact that there are only four different object shapes and objects and background move independently, which makes them easy to tell apart for humans. Moreover, they generate large amounts of false positives, often using multiple slots for the background or an object, with the split between slots mostly following color changes (see \figref{vimon_tex_vmds_rand}, \figref{op3_tex_vmds_rand}, \figref{scalor_tex_vmds_rand}).

\resultstexvmds

\section{Discussion}

Our experimental results provide insights into the inductive biases and failure cases of object-centric models that were not apparent from their original publications. Despite the positive results shown in each of the papers for the evaluated methods, a controlled, systematic analysis demonstrates that they do not convincingly succeed at tracking, which is fundamentally what object-centric video methods should enable.

\textsc{TBA} has a significantly lower MOTP than the other models on all datasets, suggesting that the simple rendering-based decoder using a fixed \textbf{template} might be less suitable to generate accurate segmentation masks (see also \figref{tba_spmot_rand} and \figref{tba_vmds_rand}) compared to the VAE-based approaches of \textsc{ViMON}, \textsc{OP3} and \textsc{SCALOR}.

Handling \textbf{occlusion} of objects during the video is a key property object-centric representations should be capable of. Qualitatively and quantitatively, \textsc{OP3} is more robust to occlusion than the other models, suggesting that its dynamics network which models interactions between objects is currently most successful at modeling occluded objects. Surprisingly, \textsc{TBA}  and \textsc{SCALOR}, which explicitly encode depth, do not handle occlusion more gracefully than \textsc{ViMON}, whose simpler architecture has no explicit way of dealing with depth. Moving forward, occlusion handling is a key component that object-centric video models need to master, which can be addressed by either equipping the model with a potent interaction module, that takes pairwise interaction between objects (including occlusion) into account, similar to \textsc{OP3}'s dynamics model, or ensuring that the depth reasoning of the models works as intended, which may be preferable, as explained below.

All models struggle with detecting objects that have similar \textbf{color} as the background (for \textsc{TBA}: dark objects, since background is removed and set to black in a pre-processing step). Color is a reliable cue to identify objects in these datasets. However, the auto-encoding objective incurs little extra loss for missing objects with similar color as the background and, thus, the models appear to not to learn to properly reconstruct them. In order to scale to data with more visual complexity, one might want to replace the pixel-wise reconstruction with for instance a loss based in feature space in order to focus more on reconstructing semantic content rather than high-frequency texture, as is done when using perceptual loss functions \citep{Gatys15,Hou2017} or by using contrastive learning \citep{Kipf2020}.
Furthermore, the models---particularly so \textsc{OP3}---struggle with separating objects of similar colors from each other. This result hints at a mismatch between the intuitions motivating these models and what the models actually learn: 
it should be more efficient in terms of the complexity of the latent representation to decompose two objects---even of similar colors---into two masks with simple shapes, rather than encoding the more complicated shape of two objects simultaneously in one slot. However, since none of the models handle occlusion with amodal segmentation masks (i.\,e. including the occluded portion of the object) successfully, they learn to encode overly complex (modal) mask shapes. As a consequence, they tend to merge similarly colored objects into one slot. This result suggests that resolving the issues surrounding the depth reasoning in combination with amodal segmentation masks would enable much more compact latents and could also resolve the merging of similarly colored objects.

A major difference between models is the \textbf{spatial transformer} based model formulation of \textsc{TBA} and \textsc{SCALOR}, compared to \textsc{ViMON} and \textsc{OP3}, which operate on image-sized masks. The parallel processing of objects and the processing of smaller bounding boxes renders training \textsc{TBA} and \textsc{SCALOR} to be significantly faster and more memory efficient, enabling them to scale to a larger number of objects. 
On the downside, the spatial transformer introduces its own complications. \textsc{TBA} depends strongly on its prior on object size and performs well only when this prior fits the data well as well as when the data contains little variation in object sizes, as in SpMOT (\tabref{tab:result}). However, it is not able to handle VMDS and its larger variation in object sizes and shapes.
\textsc{SCALOR} performs tracking well in scenes where objects are clearly separated, but struggles to separate objects that partially occlude each other when entering the scene. This difficulty is caused by its discovery mechanism which can propose at most one bounding box per grid cell, leading to a high number of misses on data sets which feature significant occlusion (VOR and VMDS). Unfortunately, simply increasing the number of proposals does not provide a simple solution, as \textsc{SCALOR}'s performance is sensitive to properly tweaking the number of proposals.

Choosing a class of models is therefore dependent on the data set one wants to apply it to as well as the computational resources at one's disposal. Data sets that feature a high number of objects~(>10) that are well separated from each other make a method like SCALOR, which can process objects in parallel, advisable. On data sets with a lower number of objects per scene which feature heavy occlusion, methods like OP3 and ViMON will likely achieve better results, but require a high computational budget for training.

In conclusion, our analysis shows that none of the models solve the basic challenges of tracking even for relatively simple synthetic data sets.
Future work should focus on developing robust mechanisms for reliably handling depth and occlusion, additionally combining the transformer-based efficiency of \textsc{TBA} and \textsc{SCALOR} with the stable training of \textsc{ViMON} and the interaction model of \textsc{OP3}. The key open challenges for scaling these models to natural videos include their computational inefficiency, complex training dynamics, as well as over-dependence on simple appearance cues like color.

\acks{}
This work was supported by the German Federal Ministry of Education and Research through the T\"ubingen AI Center (FKZ: 01IS18039A, 01IS18039B); Deutsche Forschungsgemeinschaft (DFG, German Research Foundation) via grant EC 479/1-1 to A.S.E; and  Collaborative Research Center (Projektnummer 276693517 -- SFB 1233: Robust Vision).
We thank the International Max Planck Research School for Intelligent Systems (IMPRS-IS) for supporting Marissa A. Weis, Kashyap Chitta, and Yash Sharma. The authors would also like to thank Eshed Ohn-Bar for helpful discussions, as well as Rishi Veerapaneni, John D. Co-Reyes, and Michael Chang particularly with regards to applying OP3 to our experimental setting.

\newpage
\appendix

\renewcommand\thefigure{\thesection.\arabic{figure}}
\renewcommand\thetable{\thesection.\arabic{table}}

\textbf{\Large Appendices}
\newline

\setcounter{figure}{0}
\setcounter{table}{0}

In the appendix, we first discuss the metrics used (\secref{metrics_suppl}) and describe the data generation process (\secref{datasets_suppl}). We then describe the methods \textsc{MONet}, \textsc{ViMON}, \textsc{TBA}, \textsc{IODINE}, \textsc{OP3} and \textsc{SCALOR} (\secref{method_suppl}).  \secref{implem} contains information regarding the implementation details and training protocols. Finally, we provide additional qualitative and quantitative experimental results in \secref{add_results}.

\section{Evaluation Protocol Details} \label{metrics_suppl}
We quantitatively evaluate all models on three data sets using the standard CLEAR MOT metrics \citep{Bernardin08}. Our evaluation protocol is adapted from the multi-object tracking (MOT) challenge \citep{Milan16}, a standard computer vision benchmark for supervised object tracking. In particular, we focus on the metrics provided by the py-motmetrics package\footnote{\label{note1}\url{https://pypi.org/project/motmetrics/}}. 

\subsection{Mapping}
In each frame, object predictions of each model in the form of binary segmentation masks are mapped to the ground truth object segmentation masks. We require that each pixel is uniquely assigned to at most one object in the ground truth and the predictions, respectively. Matching is based on the intersection over union (IoU) between the predictions and the ground truth masks \citep{Voigtlaender2019}. A valid correspondence between prediction and object has to exceed a threshold in IoU of 0.5. Predictions that are not mapped to any ground truth mask are classified as false positives (FPs). Ground truth objects that are not matched to any prediction are classified as misses. Following \citep{Bernardin08}, ground truth objects that are mapped to two different hypothesis IDs in subsequent frames are classified as ID switches for that frame.

\subsection{MOT Metrics}

\textbf{MOT Accuracy (MOTA)} measures the fraction of all failure cases, i.e. false positives (FPs), misses and ID switches compared to total number of objects present in all frames. \textbf{MOT Precision (MOTP)} measures the total accuracy in position for matched object hypothesis pairs, relative to total number of matches made. We use percentage Intersection over Union (IoU) of segmentation masks as the accuracy in position for each match. \textbf{Mostly Tracked (MT)} is the ratio of ground truth objects that have been tracked for at least 80\% of their lifespan.(i.e. 80\% of the frames in which they are visible). MT as implemented by py-motmetrics counts trajectories of objects as correctly tracked even if ID switches occur. We use a strictly more difficult definition of MT that counts trajectories with ID switches as correctly detected but not correctly tracked. Consequently, we add the \textbf{Mostly Detected (MD)} measure which does not penalize ID switches. \textbf{Match}, \textbf{Miss}, \textbf{ID Switches (ID S.)} and \textbf{FPs} are reported as the fraction of the number of occurrences divided by the total number of object occurrences.

\begin{equation}\label{mota}
    \mathrm{MOTA} = 1 - \frac{\sum_{t=1}^{T}{\mathrm{M}_t + \mathrm{FP}_t + \mathrm{IDS}_t}}{\sum_{t=1}^{T}{\mathrm{O}_t}}
\end{equation}
where $\mathrm{M}_t$, $\mathrm{FP}_t$,and $\mathrm{IDS}_t$ are the number of misses, false positives and ID switches, respectively, for time $t$, and $\mathrm{O}_t$ is the number of objects present in frame $t$. Note that MOTA can become negative, since the number of FPs is unbounded. 

\begin{equation}\label{motp}
    \mathrm{MOTP} = \frac{\sum_{t=1}^{T}\sum_{i=1}^{I}{d_t^i}}{\sum_{t=1}^{T}{c_t}}
\end{equation}
where $d_t^i$ is the total accuracy in position for the $i^{\mathrm{th}}$ matched object-hypothesis pair measured in IoU between the respective segmentation masks and $c_t$ is the number of matches made in frame $t$.

Note that we exclude the background masks for \textsc{ViMON} and \textsc{OP3} before evaluating tracking based on IoU. The Video Object Room (VOR) data set can contain up to three background segments, namely the floor and up to two wall segments. In order to exclude all background slots regardless of whether the model segments the background as one or as multiple masks, we remove all masks before the tracking evaluation that have an IoU of more than 0.2 with one of the ground truth background masks; we empirically tested that this heuristic is successful in removing background masks regardless of whether the models segments it as one or as three separate ones.

\setcounter{figure}{0}
\setcounter{table}{0}

\section{Data Set Generation Details} \label{datasets_suppl}
\subsection{Video Multi-dSprites (VMDS)}
The Multi-DSprites Video data set consists of 10-frame video sequences of 64$\times$64 RGB images with multiple moving sprites per video. In order to test temporal aggregation properties of the models, the test set contains 20 frame-long sequences. Each video contains one to four sprites following the data set proposed in \citep{Burgess19} that move independently of each other and might partially or fully occlude one another. The sprites are sampled uniformly from the dSprites data set \citep{Matthey17} and colored with a random RGB color. The background is uniformly colored with a random RGB color. Random trajectories are sampled per object by drawing $x$ and $y$ coordinates from a Gaussian process with squared exponential covariance kernel $\mathrm{cov}[x_s,x_t]=\exp[-(x_s-x_t)^2/(2\tau^2)]$ and time constant $\tau=10$ frames, and then shifted by an initial $(x,y)$-position of the sprite centroid, which is uniformly sampled from $[10,54]$ to ensure that the object is within the image boundaries. Trajectories that leave these boundaries are rejected. In occlusion scenarios, larger objects are always in front of smaller objects to disambiguate prediction of occlusion. The training set consists of 10,000 examples whereas the validation set as well as the test set contain 1,000 examples each. Additionally, we generated four challenge sets and three out-of-distribution test sets for VMDS that contain specifically challenging scenarios. Each test set consists of 1,000 videos of length 10 frames, which we describe in the following.

\subsubsection{VMDS Challenge Sets} \label{vmds_chall_suppl}

\textbf{Occlusion test set.} In each video, one or more objects are heavily occluded and thus often are not visible at all for a few frames. This is ensured by sampling object trajectories that cross path, i.e. at least in one video frame, two objects are centered on the same pixel. The time step and spatial position of occlusion is sampled randomly. Object trajectories are sampled independently as described above and then shifted such that they are at the sampled position of occlusion at time $t$.
Videos contain two to four sprites (\figref{tab:challenge}), since at least two objects are necessary for occlusion.

\textbf{Small Objects.} Videos contain one to four sprites with all sprites being of the smallest size present in the original dSprites \citep{Matthey17} data set (\figref{tab:challenge}). Other than that, it follows the generation process of the regular training and test set.

\textbf{Large Objects.} Videos contain one to four sprites with all sprites being of the largest size present in the original dSprites \citep{Matthey17} data set (\figref{tab:challenge}). Other than that, it follows the generation process of the regular training and test set. 

\textbf{Same Color.} Videos contain two to four sprites which are identically colored with a randomly chosen color. Other than that, it follows the generation process of the regular training and test set (\figref{tab:challenge}).

\subsubsection{VMDS Out-of-Distribution Test Sets} \label{vmds_ood_suppl}

\textbf{Rotation test set}. Sprites rotate around their centroid while moving. The amount of rotation between two video frames is uniformly sampled between 5 and 40 degrees, and is constant for each object over the course of the video. Direction of rotation is chosen randomly. Rotation is not included as a transformation in the training set (\figref{tab:ood}).

\textbf{Color change test set.} Sprites change their color gradually during the course of the video. The initial hue of the color is chosen randomly as well as the direction and amount of change between two frames, which stays the same for each object over the course of the video. Saturation and value of the color are kept constant. Color changes are not part of the training set (\figref{tab:ood}).

\textbf{Size change test set.} Sprites change their size gradually during the course of the video. The original dSprites data set \citep{Matthey17} contains six different sizes per object. For each object, its size is sampled as either the smallest or largest in the first frame as well as a random point in time, at which it starts changing its size. At this point in time, it will either become larger or smaller, respectively, increasing or decreasing each frame to the next larger or smaller size present in the original dSprites data set, until the largest or smallest size is reached. Size changes are not part of the training set (\figref{tab:ood}).

\subsubsection{texVMDS} \label{tex_vmds_suppl}

The Textured Multi-DSprites Video data set consists of 10-frame video sequences of 64$\times$64 RGB images with 1--4 moving sprites per video. The background is a linearly moving crop from a uniformly sampled image \citep{Deng09} from the ImageNet Large Scale Visual Recognition Challenge dataset (ILSVRC2012) and the texture of each sprite is similarly cropped from a uniformly sampled ILSVRC2012 image \citep{Deng09}. The direction of the moving background is uniformly sampled  from one of 8 directions (up, down, left, right, upper-right, upper-left, lower-right, lower-left). Other than that, texVMDS follows the generation process of the regular VDMS.

\subsection{Sprites-MOT (SpMOT)}
Sprites-MOT, originally introduced by \citep{He19}, consists of video sequences of length 20 frames. Each frame is a 128$\times$128 RGB image. It features multiple sprites moving linearly on a black background. The sprite can have one of four shapes and one of six colors. For more information, refer to the original paper~\citep{He19}. We generate a training set consisting of 9600 examples, validation set of 384 samples and test set of 1,000 examples using the author-provided public codebase\footnote{\url{https://github.com/zhen-he/tracking-by-animation}}. However, instead of using the default setting of 20 frames per sequence, we instead generate sequences of length 10, in order to facilitate comparison to the other data sets in our study which have only 10 frames per sequence.

Frames are downsampled to a resolution of 64$\times$64 for training \textsc{ViMON}, \textsc{OP3} and \textsc{SCALOR}. 

\subsection{Video Objects Room (VOR)}
We generate a video data set based on the static 3D Objects Room data set~\citep{Burgess19}, with sequences of length 10 frames each at a resolution of 128$\times$128. This data set is rendered with OpenGL using the gym-miniworld\footnote{\url{https://github.com/maximecb/gym-miniworld}} reinforcement learning environment. It features a 3D room with up to four static objects placed in one quadrant of the room, and a camera initialized at the diagonally opposite quadrant. The objects are either static cubes or spheres, assigned one of 6 colors and a random orientation on the ground plane of the room. The camera then follows one of five trajectories moving towards the objects, consisting of a small fixed distance translation and optional small fixed angle of rotation each time step. The wall colors and room lighting are randomized, but held constant throughout a sequence. The training set consists of 10,000 sequences whereas the validation set and the test set contain 1,000 sequences each.

Frames are downsampled to a resolution of 64$\times$64 for training \textsc{ViMON}, \textsc{OP3} and \textsc{SCALOR}. 

\setcounter{figure}{0}
\setcounter{table}{0}

\section{Methods} \label{method_suppl}

In this section we describe the various methods in a common mathematical framework. For details about implementation and training, please refer to \secref{implem}.

\subsection{\textsc{MONet}} \label{method_monet}
Multi-Object-Network (\textsc{MONet}) \citep{Burgess19} is an object-centric representation model designed for static images. It consists of a recurrent attention network that sequentially extracts attention masks of individual objects and a variational autoencoder (VAE) \citep{Kingma14} that reconstructs the image region given by the attention mask in each processing step.

\boldparagraph{Attention Network} The attention network is a U-Net \citep{Ronneberger15} parameterized by~$\psi$. At each processing step $k$, the attention network receives the full image $\bx \in [0,1]^{H \times W \times 3}$ as input together with the scope variable $\bs_k \in [0,1]^{H \times W}$. The scope $\bs_k$ keeps track of the regions of the image that haven't been attended to in the previous processing steps and thus remain to be explained. The attention network outputs a soft attention mask $\bm_k \in [0,1]^{H \times W}$ and the updated scope with the current mask subtracted:

\begin{align}
    \bm_k &= \bs_{k-1}\alpha_\psi(\bx, s_{k-1}) \\
    \bs_{k+1} &= \bs_{k}(1 - \alpha_\psi(\bx, \bs_{k}))
\end{align}

where $\alpha_\psi(\bx, \bs_{k}) \in [0,1]^{H \times W}$ is the output of the U-net  and $s_0 = \mathbf{1}$. The attention mask for the last slot is given by $\bm_K = \bs_{K-1}$ to ensure that the image is fully explained, i.e. $\sum_{k=1}^{K}{\bm_k}=\mathbf{1}$.

\boldparagraph{VAE} The VAE consists of an encoder $g: [0,1]^{H \times W \times 3} \times  [0,1]^{H \times W} \to \nR^{L \times 2}$ and a decoder $h: \nR^{L} \to [0,1]^{H \times W \times 3} \times  [0,1]^{H \times W}$ which are two neural networks parameterized by $\phi$ and $\theta$, respectively. The VAE encoder receives as input the full image $\bx$ and the attention mask $\bm_k$ and computes $(\bmu_k, \log \bsigma_k)$, which parameterize the Gaussian latent posterior distribution $q_{\phi}(\bz_k|\bx,\bm_k) = \mathcal{N}(\bmu_k, \bsigma_k I)$. Using the reparametrization trick \citep{Kingma14}, $\bz_k \in \nR^L$ is sampled from the latent posterior distribution. $\bz_k$ is decoded by the VAE decoder into a reconstruction of the image component $\widehat{\bx}_k \in [0,1]^{H \times W \times 3}$ and mask logits, which are used to compute the reconstruction of the mask $\widehat{\bm}_k \in [0,1]^{H \times W}$ via a pixelwise softmax across slots. 
The reconstruction of the whole image is composed by summing over the K masked reconstructions of the VAE: $\widehat{\bx} = \sum_{k=1}^{K}{\widehat{\bm}_k \odot \widehat{\bx}_k}$.

\boldparagraph{Loss} \textsc{MONet} is trained end-to-end with the following loss function:
\begin{equation}\label{monet_loss}
\begin{split}
    L(\phi; \theta; \psi; \bx) =  - \log \sum_{k=1}^{K}{ \bm_k p_\theta (\bx|\bz_k)} + \beta D_{\mathrm{KL}}(\prod_{k=1}^{K} q_\phi(\bz_k|\bx,\bm_k) \Vert p(\bz)) \\ 
    + \gamma \sum_{k=1}^{K}{ D_{\mathrm{KL}}(q_\psi(\bm_k |\bx) \Vert p_\theta(\bm_k|\bz_k))}
\end{split} 
\end{equation}

where $p_\theta(\bx|\bz_k)$ is the Gaussian likelihood of the VAE decoder and $\bz_k \in \nR^L$ is the latent representation of slot $k$.

The first two loss terms are derived from the standard VAE objective, the Evidence Lower BOund (ELBO) \citep{Kingma14}, i.e. the negative log-likelihood of the decoder and the Kullback–Leibler divergence between the unit Gaussian prior $p(\bz)=\mathcal{N}(\mathbf{0}, I)$ and the latent posterior distribution $q_{\phi}(\bz_k|\bx,\bm_k)$ factorized across slots. Notably, the decoder log-likelihood term $p_\theta(\bx|\bz_k)$ constrains only the reconstruction within the mask, since it is weighted by the mask $\bm_k$. Additionally, as a third term, the Kullback–Leibler divergence of the attention mask distribution $q_\psi(\bm_k|\bx)$ with the VAE mask distribution $p_\theta(\widehat{\bm}_k|\bz_k)$ is minimized, to encourage the VAE to learn a good reconstruction of the masks.

\subsection{Video MONet}
\label{method_vimon}
\begin{wrapfigure}[22]{r}{0.25\textwidth}
    \includegraphics[width=0.25\textwidth]{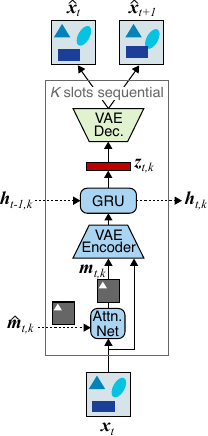}
    \centering
    \caption{\textbf{\textsc{ViMON}}. Attention network followed by VAE encoder and GRU computes latent $\bz_{t,k}$.}
    \label{fig:vimon_sppl}
    \vspace{-30pt}
\end{wrapfigure}

We propose an extension of \textsc{MONet} \citep{Burgess19}, called Video MONet (\textsc{ViMON}), which accumulates evidence over time about the objects in the scene (\figref{fig:vimon_sppl}).

\textsc{ViMON} processes a video recurrently by reconstructing one frame at a time and predicting the next frame of the video. The processing of each frame follows a logic similar to \textsc{MONet} with some notable differences. In the following, we use $t$ to indicate the time step in the video and $k$ to indicate the processing step within one video frame.

\boldparagraph{Attention Network} The attention network of \textsc{ViMON} outputs an attention mask $\bm_{t,k} \in [0,1]^{H \times W}$ in each step~$k$ conditioned on the full frame $\bx_t \in [0,1]^{H \times W \times 3}$, the scope $\bs_{t,k} \in [0,1]^{H \times W}$ and additionally the mask $\widehat{\bm}_{t,k} \in [0,1]^{H \times W}$ that was predicted by the VAE in the previous time step, in order to provide it with information about which object it should attend to in this specific slot $k$. 

\begin{equation}
    \bm_{t,k} = \bs_{t,k-1}\alpha_\psi(\bx_t, \bs_{t,k-1}, \widehat{\bm}_{t,k})
\end{equation}

\boldparagraph{VAE} The VAE of \textsc{ViMON} consists of an encoder $g(\bx_t,\bm_{t,k};\phi)$ and a decoder $h(\bz_{t,k};\theta)$. In contrast to \textsc{MONet}, the encoder in \textsc{ViMON} is followed by a gated recurrent unit (GRU) \citep{Cho14} with a separate hidden state~$h_{t,k}$ per slot~$k$. Thus, the GRU aggregates information over time for each object separately. The GRU outputs $(\bmu_{t,k}, \log \bsigma_{t,k})$ which parameterize the Gaussian latent posterior distribution $q_{\phi}(\bz_{t,k}|\bx_{t},\bm_{t,k})$ where $\bz_{t,k} \in \nR^L$ is the latent representation for slot~$k$ at time~$t$:

\begin{align}
    \bz'_{t, k} &= g(\bx_{t},\bm_{t,k}; \phi) \\
    (\bmu_{t, k}, \log \bsigma_{t,k}), \bh_{t,k} &= f(GRU(\bz'_{t, k}, \bh_{t-1,k}))) \\
    q_{\phi}(\bz_{t,k} | \bx_{t},\bm_{t,k}) &= \mathcal{N}(\bmu_{t,k}, \bsigma_{t,k} I) \qquad \forall t,k
\end{align}

where $g$ is the VAE encoder and $f$ is a linear layer.
The latent representation $\bz_{t,k}$ is sampled from the latent posterior distribution using the reparametrization trick \citep{Kingma14}. Subsequently, $\bz_{t,k}$ is linearly transformed into $\widehat{\bz}_{t+1,k}$ via a learned transformation $\bA \in \nR^{L \times L}$:  $\widehat{\bz}_{t+1,k} = \bA \bz_{t,k}$ with $\widehat{\bz}_{t+1,k}$ being the predicted latent code for the next time step $t+1$. Both $\bz_{t,k}$ and $\widehat{\bz}_{t+1,k}$ are decoded by the shared VAE decoder $h_\theta$ into a reconstruction of the image $\widehat{\bx}_{t,k} \in [0,1]^{H \times W \times 3}$ and a reconstruction of the mask $\widehat{\bm}_{t,k} \in [0,1]^{H \times W}$ as well as $\widehat{\bx}_{t+1,k}$ and $\widehat{\bm}_{t+1,k}$, respectively.

\boldparagraph{Loss} \textsc{ViMON} is trained in an unsupervised fashion with the following objective adapted from the \textsc{MONet} loss (\eqnref{monet_loss}) for videos. To encourage the model to learn about object motion, we include a prediction objective in the form of a second decoder likelihood on the next-step prediction $p_\theta(\bx_{t+1}|\widehat{\bz}_{t+1,k})$ and an additional mask loss term, which encourages the predicted VAE mask distribution $p_\theta (\widehat{\bm}_{t+1,k}|\widehat{\bz}_{t+1,k})$ to be close to the attention mask distribution $q_\psi(\bm_{t+1,k}|\bx_{t+1})$ of the next time step for each slot~$k$:

\begin{align*} \label{video_monet_loss}
    L(\phi; \theta; \psi; \bx) &= \sum_{t=1}^{T}{L_{\mathrm{negLL}} + \beta L_{\mathrm{prior}} + \gamma L_{\mathrm{mask}}} \\
    L_{\mathrm{negLL}} &=  - (\log \sum_{k=1}^{K}{ \bm_{t,k} p_\theta (\bx_t|\bz_{t,k})} + \log \sum_{k=1}^{K}{ \bm_{t+1,k} p_\theta (\bx_{t+1}|\widehat{\bz}_{t+1,k})}) \\
    L_{\mathrm{prior}} &= D_{\mathrm{KL}}(\prod_{k=1}^{K} q_\phi(\bz_{t,k}|\bx_t, \bm_{t,k}) \Vert p(\bz)) \\
    L_{\mathrm{mask}} &= \sum_{k=1}^{K}{ D_{\mathrm{KL}}(q_\psi(\bm_{t,k}|\bx_t) \Vert p_\theta (\bm_{t,k}| \bz_{t,k})) + D_{\mathrm{KL}}(q_\psi(\bm_{t+1,k}|\bx_{t+1}) \Vert p_\theta (\bm_{t+1,k}| \widehat{\bz}_{t+1,k}))}
\end{align*}

\subsection{Tracking by Animation} 
\label{method_tba}

\begin{wrapfigure}{r}{0.3\textwidth}
    \includegraphics[width=0.3\textwidth]{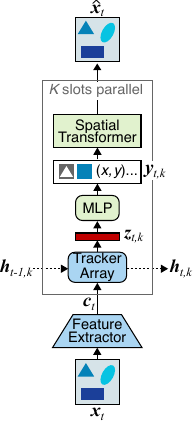}
    \centering
    \caption{\textbf{\textsc{TBA}}. Feature extractor CNN $f$ and tracker array $g$ to get latent $\bz_{t,k}$. MLP $h$ outputs mid-level representation $\by_{t,k}$, and Spatial Transformer renders reconstruction.}
    \label{fig:tba_sppl}
    \vspace{-25pt}
\end{wrapfigure}

Tracking by Animation (\textsc{TBA}) \citep{He19} is a spatial transformer-based attention model which uses a simple 2D rendering pipeline as the decoder. Objects are assigned tracking templates and pose parameters by a tracker array, such that they can be reconstructed in parallel using a renderer based on affine spatial transformation (\figref{fig:tba_sppl}). In contrast to \textsc{ViMON}, \textsc{TBA} uses explicit parameters to encode the position, size, aspect ratio and occlusion properties for each slot. Importantly, \textsc{TBA} is designed for scenes with static backgrounds, and preprocesses sequences using background subtraction \citep{Bloisi12} before they are input to the tracker array. 

\boldparagraph{Tracker Array} \textsc{TBA} uses a tracker array to output a latent representation $\bz_{t} \in \nR^{L \times K}$ at time~$t$ using a feature extractor $f(\bx_t;\psi)$ and a recurrent 'state update', where $\bc_t \in \nR^{M \times N \times C}$ is a convolutional feature representation. The convolutional feature and latent representation have far fewer elements than $\bx_t$, acting as a bottleneck:

\begin{align}
	\bc_t &= f(\bx_t;\psi), \\
	\bh_{t,k} &= RAT(\bh_{t-1,k},\bc_t; \pi), \\
	\bz_{t} &= g(\bh_t;\phi).
\end{align}

Though the state update could be implemented as any generic recurrent neural network block, such as an LSTM \citep{Hochreiter97} or GRU \citep{Cho14}, \textsc{TBA} introduces a Reprioritized Attentive Tracking (RAT) block that uses attention to achieve explicit association of slots with similar features over time. Firstly, the previous tracker state $\bh_{t-1,k}$ is used to generate key variables $\bk_{t,k}$ and  $\beta_{t,k}$:

\begin{align}
\{\bk_{t,k}, \widehat{\beta}_{t,k}\} & = \bT\bh_{t-1, k}, \label{eq_read1} \\
\beta_{t,k} & = 1 + \ln (1 +\exp({\widehat{\beta}_{t,k}})),
\end{align}

where $\bT$ is a learned linear transformation, $\bk_{t,k}\!\in\!\nR^S$ is the addressing key, and $\widehat{\beta}_{t,k}\!\in\!\nR$ is an un-normalized version of a key strength variable $\beta_{t,k}\!\in\!(1, +\infty)$. This key strength acts like a temperature parameter to modulate the feature re-weighting, which is described in the following. Each feature vector in $\bc_{t}$, denoted by $\bc_{t,m,n}\!\in\!\nR^S$, where $m\!\in\!\{1, 2, \ldots, M\}$ and $n\!\in\!\{1, 2, \ldots, N\}$ are the convolutional feature dimensions, is first used to get attention weights:

\begin{align}
W_{t,k,m,n} = \frac{\exp (\beta_{t,k} Sim(\bk_{t,k}, \bc_{t,m,n}))}{\sum_{m',n'} \exp (\beta_{t,k} Sim( \bk_{t,k}, \bc_{t,m',n'}))}. \label{eq_attention}
\end{align}

Here, $Sim$ is the cosine similarity defined as $Sim(\bp, \bq)\!=\! \bp \bq^{\mathsf{T}} / (\|\bp\| \|\bq\|)$, and $W_{t,k,m,n}$ is an element of the attention weight $\bW_{t,k}\!\in\![0,1]^{M\!\times\!N}$, satisfying $\sum_{m,n}\!W_{t,k,m,n}\!=\!1$.
Next, a read operation is defined as a weighted combination of all feature vectors of $\bc_{t}$:
\begin{align}
&\br_{t,k} = \sum_{m,n} W_{t,k,m,n}~\bc_{t,m,n} \label{eq_read2}
\end{align}
where $\br_{t,k}\!\in\!\nR^S$ is the read vector, representing the associated input feature for slot $k$. Intuitively, for slots in which objects are present in the previous frame, the model can suppress the features in $\br_{t,k}$ that are not similar to the features of that object, helping achieve better object-slot consistency. On the other hand, if there are slots which so far do not contain any object, the key strength parameter allows $\br_{t,k}$ to remain similar to $\bc_{t}$ facilitating the discovery of new objects.

The tracker state $\bh_{t, k}$ of the RAT block is updated with an RNN parameterized by $\pi$, taking $\br_{t,k}$ instead of $\bc_{t}$ as its input feature:
\begin{align}
\bh_{t, k} = RNN (\bh_{t-1, k}, \br_{t,k}; \pi) \label{eq_read3}
\end{align}

The RAT block additionally allows for sequential prioritization of trackers, which in turn allows only a subset of trackers to update their state at a given time step, improving efficiency. For full details on the reprioritization and adaptive computation time elements of the RAT block, please refer to the original paper~\citep{He19}.

\boldparagraph{Mid-Level Representation} The key feature of \textsc{TBA} is that each latent vector $\bz_{t,k}$ is further decoded into a mid-level representation $\by_{t,k} = \{y^c_{t,k}, \by^l_{t,k}, \by^p_{t,k}, \bY^s_{t,k}, \bY^a_{t,k}\}$ corresponding to interpretable, explicit object properties, via a fully-connected neural network $h(\bz_{t,k};\theta)$ as follows:

\begin{equation}
    \by_{t,k} = h(\bz_{t,k}; \theta).
\end{equation}

$h_\theta$ is shared by all slots, improving parameter efficiency. The different components of the mid-level representation are: 

\begin{itemize}
\item Confidence {$y^c_{t,k}\!\in\![0, 1]$}: Probability of existence of an object in that slot.
\item Layer {$\by^l_{t,k}\!\in\!\{0,1\}^O$}: One-hot encoding of the discretized pseudo-depth of the object relative to other objects in the frame. Each image is considered to be composed of $O$ object layers, where higher layer objects occlude lower layer objects and the background is the zeroth (lowest) layer.
E.g., when $O\!=\!4$, $\by^l_{t,k}\!=\![0, 0, 1, 0]$ denotes the third layer. For simplicity and without loss of generality, we can also denote the same layer with its integer representation $y^l_{t,k}=3$.
\item Pose {$\by^p_{t,k}\!=\![\widehat{s}^x_{t,k}, \widehat{s}^y_{t,k}, \widehat{t}^x_{t,k}, \widehat{t}^y_{t,k}]\!\in\![-1,1]^4$}: Normalized object pose for calculating the scale $[s^x_{t,k}, s^y_{t,k}] = [1 + \eta^{x} \widehat{s}^x_{t,k}, 1 + \eta^{y} \widehat{s}^y_{t,k}]$ and the translation $[t^x_{t,k}, t^y_{t,k}] = [\frac{W}{2}\widehat{t}^x_{t,k}, \frac{H}{2}\widehat{t}^y_{t,k}]$, where $\eta^{x},\eta^{y}>0$ are constants.
\item Shape {$\bY^s_{t,k}\!\in\!\{0,1\}^{U\!\times\!V}$} and Appearance {$\bY^a_{t,k}\!\in\![0,1]^{U\!\times\!V\!\times\!3}$}: Object template, with hyperparameters $U$ and $V$ typically set much smaller than the image dimensions $H$ and $W$. Note that the shape is discrete (for details, see below) whereas the appearance is continuous.
\end{itemize}
In the output layer of $h_\theta$, $y^c_{t,k}$ and $\bY^a_{t,k}$ are generated by the sigmoid function, $\by^p_{t,k}$ is generated by the tanh function, and $\by^l_{t,k}$ as well as $\bY^s_{t,k}$ are sampled from the Categorical and Bernoulli distributions, respectively. As sampling is non-differentiable, the Straight-Through Gumbel-Softmax estimator~\citep{Jang17} is used to reparameterize both distributions so that backpropagation can still be applied.
	
\boldparagraph{Renderer} To obtain a frame reconstruction, the renderer scales and shifts $\bY_{t,k}^s$ and $\bY_{t,k}^a$ according to $\by_{t,k}^{p}$ via a Spatial Transformer Network (STN) \citep{Jaderberg15}:
\begin{align}
{\bm}_{t,k} & = {STN}(\bY_{t,k}^{s}, \by_{t,k}^{p}), \\
\widehat{\bx}_{t,k} & = {STN}(\bY_{t,k}^{a}, \by_{t,k}^{p}).
\end{align}
where ${\bm}_{t,k} \! \in \! \{0, 1\}^{D}$ and $\widehat{\bx}_{t,k} \!\in \! [0, 1]^{D \!\times\! 3}$ are the spatially transformed shape and appearance respectively. To obtain the final object masks $\widehat{\bm}_{t,k}$, an occlusion check is performed by initializing $\widehat{\bm}_{t,k}=y^c_{t,k}\bm_{t,k}$, then removing the elements of $\widehat{\bm}_{t,k}$ for which there exists an object in a higher layer. That is, for $k\!=\! 1, 2, \ldots, K$ and $\forall j \ne k$ where $y^l_{t,j} > y^l_{t,k}$:
\begin{align}
\widehat{\bm}_{t,k}  = (\mathbf{1} - \bm_{t,j}) \odot \widehat{\bm}_{t,k}.
\end{align}
In practice, the occlusion check is sped up by creating intermediate `layer masks', partially parallelizing the operation. Please see the original paper for more details \citep{He19}. The final reconstruction is obtained by summing over the $K$ slots, $\widehat{\bx}_t = \sum_{k=1}^{K}{\widehat{\bm}_{t,k} \odot \widehat{\bx}_{t,k}}$.

\boldparagraph{Loss} Learning is driven by a pixel-level reconstruction objective, defined as:

\begin{align}
L(\phi; \psi; \pi; \theta; \bx) &= \sum_{t=1}^{T} \left( MSE (\widehat{\bx}_t, \bx_t) + \lambda \cdot \frac{1}{K}\sum_{k=1}^{K} s^x_{t,k}~s^y_{t,k} \right),
\end{align}

where $MSE$ refers to the mean squared error and the second term penalizes large scales $[s^x_{t,k}, s^y_{t,k}]$ in order to make object bounding boxes more compact.

\newpage

\subsection{\textsc{IODINE}}
\label{method_iodine}
\begin{wrapfigure}[23]{r}{0.5\textwidth}
    \includegraphics[width=0.5\textwidth]{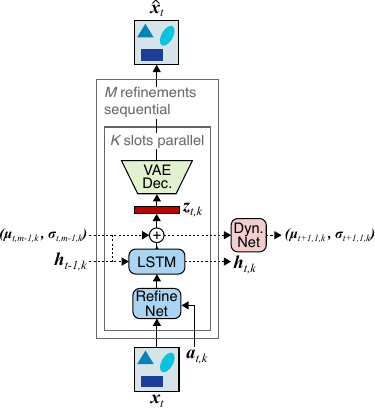}
    \centering
    \caption{\textbf{\textsc{OP3}}. Refinement network $f$ followed by LSTM and dynamics network $d$ compute latent~$\bz_{t,k}$.}
    \label{fig:op3_sppl}
    \vspace{-12pt}
\end{wrapfigure}

The Iterative Object Decomposition Inference NEtwork (\textsc{IODINE}) \citep{Greff19}, similar to \textsc{MONet} \citep{Burgess19}, learns to decompose a static scene into a multi-slot representation, in which each slot represents an object in the scene and the slots share the underlying format of the independent representations. In contrast to \textsc{MONet}, it does not recurrently segment the image using spatial attention, rather it starts from an initial guess of the segmentation of the whole image and iteratively refines it. Thus, the inference component of both models differ, while the generative component is the same.

\textbf{Iterative Inference.} As with \textsc{MONet}, \textsc{IODINE} models the latent posterior $q(\bz_k|\bx)$ per slot $k$ as a Gaussian parameterized  by $(\bmu_{m,k}, \bsigma_{m,k}) \in \nR^{L \times 2}$. To obtain latent representations for independent regions of the input image, \textsc{IODINE} starts from initial learned posterior parameters $(\bmu_{1,k}, \bsigma_{1,k})$ and iteratively refines them using the refinement network $f_\phi$ for a fixed number of refinement steps~$M$. $f_\phi$ consists of a convolutional neural network (CNN) in combination with an LSTM cell \citep{Hochreiter97} parameterized by $\phi$. In each processing step, $f_\phi$ receives as input the image $\bx \in [0,1]^{H \times W \times 3}$, a sample from the current posterior estimate $\bz_{m,k} \in \nR^L$ and various auxiliary inputs $\ba_k$, which are listed in the original paper~\citep{Greff19}. %
The posterior parameters are concatenated with the output of the convolutional part of the refinement network and together form the input to the refinement LSTM. The posterior parameters are additively updated in each step~$m$ in parallel for all $K$ slots:

\begin{equation}\label{lambda_update}
    (\bmu_{m+1,k}, \bsigma_{m+1,k}) = (\bmu_{m,k}, \bsigma_{m,k}) + f_\phi(\bz_{m,k}, \bx, \ba_k )
\end{equation}

\textbf{Decoder}. In each refinement step $m$, the image is represented by $K$ latent representations $\bz_{m,k}$. Similar to \textsc{MONet}, each $\bz_{m,k}$ is independently decoded into a reconstruction of the image $\widehat{\bx}_{m,k} \in [0,1]^{H \times W \times 3}$ and mask logits $\widetilde{\bm}_{m,k}$, which are subsequently normalized by applying the softmax across slots to obtain the masks $\bm_{m,k} \in [0,1]^{H \times W}$.
The reconstruction of the whole image at each refinement step $m$ is composed by summing over the K masked reconstructions of the decoder: $\widehat{\bx} = \sum_{k=1}^{K}{\bm_{m,k} \odot \widehat{\bx}_{m,k}}$.

\textbf{Training.}
\textsc{IODINE} is trained by minimizing the following loss function that consists of the the Evidence Lower BOund (ELBO) \citep{Kingma14} unrolled through $N$ iterations:

\begin{equation}\label{iodine_loss}
    L(\theta, \phi, (\bmu_{1,k}, \bsigma_{1,k}); \bx) = \sum_{m=1}^{M}{\frac{m}{M}\!\left[- \log \sum_{k=1}^{K}{ \bm_{m,k} p_\theta (\bx|\bz_{m,k})} + D_{\mathrm{KL}}\!\left(\prod_{k=1}^{K} q_\phi(\bz_{m,k}|\bx) \Vert p(\bz)\!\right)\!\right]}
\end{equation}

where $p_\theta (\bx|\bz_{m,k})$ is the decoder log-likelihood weighted by the mask $\bm_k$ and $D_{\mathrm{KL}}$ is the Kullback-Leibler divergence between the unit Gaussian prior $p(\bz)=\mathcal{N}(\mathbf{0}, I)$ and the latent posterior distribution $q(\bz_{m,k}|\bx)$ factorized across slots.

\subsection{Object-centric Perception, Prediction, and Planning (\textsc{OP3})} 
\label{method_op3}
Object-centric Perception, Prediction, and Planning (\textsc{OP3}) \citep{Veerapaneni19} extends \textsc{IODINE} to work on videos and in a reinforcement learning (RL) setting. It uses the above described \textsc{IODINE} as an observation model to decompose visual observations into objects and represent them independently. These representations are subsequently processed by a dynamics model that models the individual dynamics of the objects, the pairwise interaction between the objects, as well as the action's effect on the object's dynamics, predicting the next frame in latent space (\figref{fig:op3_sppl}). By modeling the action's influence on individual objects, \textsc{OP3} can be applied to RL tasks.

\textsc{OP3} performs $M$ refinement steps after each dynamics step. %

\textbf{Refinement network.} The refinement steps proceed as in the description for \textsc{IODINE} in \secref{method_iodine}. The input image $\bx_t \in [0,1]^{H \times W \times 3}$, which is the frame from a video at time $t$, is processed by the refinement network $f_\phi$ conditioned on a sample from the current posterior estimate $\bz_{t,m,k} \in \nR^L$. The refinement network outputs an update of the posterior parameters $(\bmu_{t,m,k}, \bsigma_{t,m,k})$ (see \eqnref{lambda_update}). The posterior parameters $(\mu_{1,1,k}, \sigma_{1,1,k})$ are randomly initialized. %

\textbf{Dynamics model.} After refinement, samples from the current posterior estimate $\bz_{t,M,k}$ for each slot~$k$ are used as input to the dynamics network. The dynamics model $d_\psi$ consists of a series of linear layers and nonlinearities parameterized by $\psi$. It models the individual dynamics of the objects per slot~$k$, the pairwise interaction between all combinations of objects, aggregating them into a prediction of the posterior parameters for the next time step $t+1$ for each object $k$. The full dynamics model additionally contains an action component that models the influence of a given action on each object, which we do not use in our tracking setting. The predicted posterior parameters are then used in the next time step as initial parameters for the refinement network.

\begin{equation}\label{op3_dynamics}
    (\bmu_{t,1,k}, \bsigma_{t,1,k}) = d_\psi(\bz_{t-1,M,k}, \bz_{t-1,M, [\neq k]}))
\end{equation}

\textbf{Training.}
\textsc{OP3} is trained end-to-end with the ELBO used at every refinement and dynamics step, with the loss $L(\theta, \phi; \bx)$ given by:

\begin{equation}\label{op3_loss}
    \sum_{t=1}^{T}{\frac{1}{T} \sum_{m=1}^{M+1}{\frac{\text{min}(m,M)}{M} \left(- \log \sum_{k=1}^{K}{\bm_{t,m,k} p_\theta (\bx_t|\bz_{t,m,k})} + D_{\mathrm{KL}}(\prod_{k=1}^{K} q_\phi(\bz_{t,m,k}|\bx_t) \Vert q(\bz_{t,1,k}|\bx_t))\right)}}
\end{equation}

where for time step $1$, $q(\bz_{1,1,k}|\bx_1) = \mathcal{N}(\mathbf{0}, I)$.

\subsection{SCALable Object-oriented Representation (\textsc{SCALOR})} \label{method_scalor}
SCALable Object-oriented Representation (\textsc{SCALOR})~\citep{Jiang19} is a spatial transformer-based model that extends SQAIR~\citep{Kosiorek18} to scale to cluttered scenes. Similar to \textsc{TBA} is factors the latent representations in pose, depth and appearance per object and uses spatial transformers~\citep{Jaderberg15} to render objects in parallel. In contrast to \textsc{TBA}, it can handle dynamic backgrounds by integrating a background RNN that models background transitions. 

\boldparagraph{Proposal-Rejection Module}: \textsc{SCALOR} uses a proposal-rejection module $g$ to discover new objects. All frames up to the current time step $x_{1:t}$ are first encoded using a convolutional LSTM~$f$. The resulting features are then aggregated with an encoding of propagated object masks and divided into $H \times W$ grid cells.

\begin{align}
	\bc_t^{img} &= f(\bx_{1:t};\psi) \\
	\bc_t^{mask} &= MaskEncoder(M_t^P) \\
	\bc_t^{agg} &= Concat([\bc_t^{img}, \bc_t^{mask}],
\end{align}

Per grid cell a latent variable $\bz_{t,h,w}$ is proposed. Proposal generation is done in parallel. Each $\bz_{t,h,w}$ consists of existence, pose, depth and appearance parameters ($\bz_{t,h,w}^{pres}, \bz_{t,h,w}^{pose}, \bz_{t,h,w}^{depth}, \bz_{t,h,w}^{what}$).  

\begin{align}
	\bz_{t,h,w}^{pres} &\sim Bern(\cdot| g_1(\bc_t^{agg})) \\ 
	\bz_{t,h,w}^{depth} &\sim \mathcal{N}(\cdot| g_2(\bc_t^{agg})) \\
	\bz_{t,h,w}^{pose} &\sim \mathcal{N}(\cdot| g_3(\bc_t^{agg}))
\end{align}

where $g_1, g_2$ and $g_3$ are convolutional layers.

The appearance parameters $\bz_{t,h,w}^{what}$ are obtained by first taking a glimpse from frame $x_t$ of the area specified by $\bz_{t,h,w}^{pose}$ via a Spatial Transformer Network (STN)~\citep{Jaderberg15} and subsequently extracting features from it via a convolutional neural network:

\begin{align}
	\bc_{t,h,w}^{att} &= STN(x_t, \bz_{t,h,w}^{pose}) \\ 
	\bz_{t,h,w}^{what} &\sim \mathcal{N}(\cdot| GlimpseEnc(\bc_{t,h,w}^{att})) \\
	\bo_{t,h,w}, \bm_{t,h,w} &= STN^{-1}(GlimpseDec(\bz_{t,h,w}^{what}), \bz_{t,h,w}^{pose})
\end{align}

where $\bo_{t,h,w}$ is the object RGB glimpse and $\bm_{t,h,w}$ is the object mask glimpse.

\begin{wrapfigure}[27]{r}{0.4\textwidth}
    \includegraphics[width=0.4\textwidth]{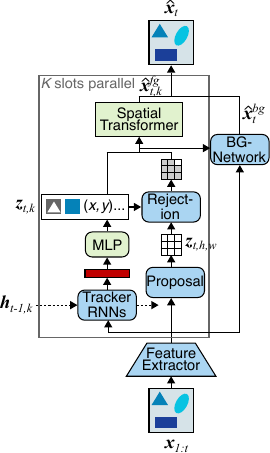}
    \centering
    \caption{\textbf{\textsc{SCALOR}} Feature extractor CNN  $f$ followed by tracker RNNs or proposal-rejection module to compute latent $\bz_{t,k}$. Spatial Transformer in addition to background module renders reconstruction.}
    \label{fig:scalor_sppl}
\end{wrapfigure}

In the rejection phase, objects that overlap more than a threshold $\tau$ in pixel space with a propagated object from the previous time step are rejected.

\boldparagraph{Propagation Module}: During propagation, for each object $k$ from the previous time step $t-1$ a feature attention map $\ba_{t,k}$ from the encoded frame features $\bc_t^{img}$ is extracted centered on the position of the object in the previous time step and used to update the hidden state $\bh_{t,k}$ of the tracker RNN for object $k$.

\begin{align}
	\ba_{t,k} &= att(STN(\bc_t^{img}, \bz_{t-1, k}^{pose}) \\
	\bh_{t,k} &= GRU([\ba_{t,k}, \bz_{t-1,k}], \bh_{t-1,k}) \\
	\bz_{t,k} &= update(\ba_{t,k}, \bh_{t,k}, \bz_{t-1,k})
\end{align}

where $STN$ is a spatial transformer module \citep{Jaderberg15}. If $\bz_{t,k}^{pres} = 1$ the latent representation $\bz_{t,k}$ of the respective object $k$ will be propagated to the next time step.

\boldparagraph{Background}: The background of each frame $x_t$ is encoded using a convolutional neural network conditioned on the masks $M_{t}$ of the objects present at time step $t$ and decoded using a convolutional neural network.

\begin{align}
	(\bmu^{bg}, \bsigma^{bg}) &= BgEncoder(x_t, (1-M_t)) \\
	\bz_t^{bg} &\sim \mathcal{N}(\bmu^{bg}, \bsigma^{bg}) \\
	\widehat{\bx}_t^{bg} &= BgDecoder(\bz_t^{bg})
\end{align}

\boldparagraph{Rendering}:
To obtain frame reconstructions $\widehat{\bx}_t$ foreground object appearances and masks are scaled and shifted using via a Spatial Transformer Network (STN):

\begin{align}
    \widehat{\bx}_{t,k}^{fg} &= STN^{-1}(\bo_{t,k}, \bz_{t,k}^{pose}) \\
    \bgamma_{t,k} &= STN^{-1}(\bm_{t,k} \cdot \bz_{t,k}^{pres} \sigma(-\bz_{t,k}^{depth}), \bz_{t,k}^{pose}) \\
    \widehat{\bx}_t^{fg} &= \sum_{K} \widehat{\bx}_{t,k}^{fg} \bgamma_{t,k}
\end{align}

Subsequently, foreground objects and background reconstruction are combined as follows to obtain the final reconstruction:

\begin{align}
    \widehat{\bx}_t = \widehat{\bx}_t^{fg} + (1 - M_t) \odot \widehat{\bx}_t^{bg}
\end{align}

\boldparagraph{Training}:
\textsc{SCALOR} is trained on frame reconstruction using the evidence lower bound (ELBO):

\begin{equation}\label{scalor_loss}
    \sum_{t=1}^{T} - \log  p_\theta (\bx_t|\bz_{t}) + D_{\mathrm{KL}}( q_\phi(\bz_{t}|\bz_{<t},\bx_{\leq t}) \Vert q(\bz_{t}|\bz_{<t}))
\end{equation}

\FloatBarrier

\setcounter{figure}{0}
\setcounter{table}{0}

\section{Model Implementation Details} \label{implem}

\subsection{Baseline: k-Means}\label{implem_kmeans}
As a baseline, we implement a simple color segmentation algorithm to evaluate if this is sufficient to solve the benchmark data sets. To this end, we cluster the videos using the k-Means algorithm as implemented by sklearn \citep{Pedregosa11}. The number of clusters are chosen equal to the number of slots in the other models, i.e. $K=5$ for SpMOt, $K=6$ for VMDS  and $K=8$ for VOR. We run the clustering three times per video with different random seeds. Each pixel is treated as one sample with 3 features (RGB values). k-Means clustering is performed for each frame in the video separately. To obtain the tracking performance, we match objects in consecutive frames using IoU of the extracted segmentation masks with the Hungarian matching algorithm \citep{Kuhn55}.

\subsection{\textsc{MONet} and Video \textsc{MONet}}\label{implem_vimon}
\boldparagraph{VAE}
Following \citep{Burgess19}, the VAE encoder is a CNN with 3x3 kernels, stride 2, and ReLU activations (\tabref{tab:enc}). It receives the input image and mask from the attention network as input and outputs $(\mu$, $\log \sigma)$ of a 16-dimensional Gaussian latent posterior. The GRU has 128 latent dimensions and one hidden state per slot followed by a linear layer with 32 output dimensions. The VAE decoder is a Broadcast decoder as published by \citep{Watters19b} with no padding, 3x3 kernels, stride 1 and ReLU activations (\tabref{tab:dec}). The output distribution is an independent pixel-wise Gaussian with a fixed scale of $\sigma=0.09$ for the background slot and $\sigma=0.11$ for the foreground slots.
\vaeencodertable
\vaedecodertable

\boldparagraph{Attention Network}
The attention network is a U-Net \citep{Ronneberger15} and follows the architecture proposed by \citep{Burgess19}. The down and up-sampling components consist each of five blocks with 3x3 kernels, 32 channels, instance normalisation, ReLU activations and down- or up-sampling by a factor of two. The convolutional layers are bias-free and use stride 1 and padding 1. A three-layer MLP with hidden layers of size 128 connect the down- and the up-sampling part of the U-Net.

\boldparagraph{Training}
\textsc{MONet} and \textsc{ViMON} are implemented in PyTorch \citep{Paszke19} and trained with the Adam optimizer \citep{Kingma14b} with a batch size of 64 for \textsc{MONet} and 32 for \textsc{ViMON}, using an initial learning rate of 0.0001. Reconstruction performance is evaluated after each epoch on the validation set and the learning rate is decreased by a factor of 3 after the validation loss hasn't improved in 25 consecutive epochs for \textsc{MONet} and 100 epochs for \textsc{ViMON}, respectively. \textsc{MONet} and \textsc{ViMON} are trained for 600 and 1000 epochs, respectively. The checkpoint with the lowest reconstruction error is selected for the final MOT evaluation.
\textsc{MONet} is trained with $\beta=0.5$ and $\gamma=1$ and \textsc{ViMON} is trained with $\beta=1$ and $\gamma=2$. $K=5$ for SpMOt, $K=6$ for VMDS  and $K=8$ for VOR. Due to the increased slot number for VOR, batch size for \textsc{ViMON} had to be decreased to 24 to fit into the GPU memory. Respectively, the initial learning rate is set to 0.000075 for \textsc{ViMON} on VOR.
We initialize the attention network and the VAE in \textsc{ViMON} with the pre-trained weights from \textsc{MONet} to facilitate learning and speed up the training. 
Note that for all evaluations, the reconstructed masks $\widehat \bm$ from the VAE were used.

\textbf{Sprites-MOT Initialization:} When training \textsc{MONet} and Video \textsc{MONet} on Sprites-MOT from scratch, \textsc{MONet} struggles to learn the extreme color values of the objects that Sprites-MOT features. Instead it completely focuses on learning the shapes. To circumvent that, we initialized the weights of the models with \textsc{MONet} weights that were trained for 100 epochs on Multi-dSprites. 

\boldparagraph{Hyperparameter tuning}
Hyperparameters for \textsc{MONet} and \textsc{ViMON} were selected using the best MSE on the validation set of VMDS. Hyperparameter tuning for \textsc{MONet} included $\beta$ from \{0.5, 1.0\} and $\gamma$ from \{0.5, 1.0, 2.0\} and initial learning rate from \{1e-05, 1e-04\}.

Hyperparameter tuning for \textsc{ViMON} included $\beta$ from \{0.5, 1.0\} and $\gamma$ from \{1.0, 2.0, 5.0, 10.0, 50.0\} and initial learning rate from \{1e-05, 1e-04, 1e-03\}.

Hyperparameters for \textbf{texVMDS} were kept fixed to the best selected hyperparameters for VMDS, except for the VAE latent dimension, which is tuned to account for the higher visual complexity. The VAE latent dimension 16 is selected from \{16, 32, 64, 128\} as the one with the best MOTA for \textsc{MONet} on texVMDS validation set --- analogous to the SCALOR hyperparameter tuning paradigm --- and subsequently also used for training of \textsc{ViMON} on texVMDS.

\subsection{Tracking by Animation}
\textbf{Preprocessing:} \textsc{TBA} expects its input frames to contain only foreground objects. In \citep{He19}, the authors use Independent Multimodal Background Subtraction (IMBS) \citep{Bloisi12} to remove the background from data sets consisting of natural videos with static backgrounds. Background subtraction algorithms maintain a spatio-temporal window around each pixel in the sequence, and remove the dominant mode based on a histogram of color values. Since the default implementation of IMBS has several hand-tuned thresholds corresponding to natural videos (e.g., for shadow suppression), it cannot be directly applied to synthetic data sets like VMDS without significant hyper-parameter tuning. We instead re-generate all of the VMDS data sets with identical objects and motion but a black background for our experiments with \textsc{TBA}, to mimic a well-tuned background subtraction algorithm.

\textbf{Architecture:} For SpMOT, we follow the same architecture as in \citep{He19}, while we increase the number of slots from $K=4$ to $K=5$ and number of layers from $O=3$ to $O=4$ for VMDS. Since \textsc{TBA} does not model the background, this makes the number of foreground slots equal to the other models in our study.

\textbf{Hyperparameter tuning:}
Further, we increase the size prior parameters $U\times V$ used for the shape and appearance templates from $21\times 21$ which is used for SpMOT, to $64\times 64$ for VMDS, which we empirically found gave the best validation loss among $48\times 48$, $56\times 56$, $64\times 64$ and $72\times 72$. All other architectural choices are kept fixed for both data sets, and follow \citet{He19}. Note that due to this, we trained the \textsc{TBA} models at its default resolution of 128$\times$128 unlike the 64$\times$64 resolution used by \textsc{MONet} and \textsc{OP3}.

\textbf{Training and Evaluation:} We train for 1000 epochs using the same training schedule as \citet{He19}. The checkpoint with the lowest validation loss is selected for the final MOT evaluation. Further, we observed that the discrete nature of the shape code used in TBA's mid-level representation leads to salt-and-pepper noise in the reconstructed masks. We therefore use a $2\times 2$ minimum pooling operation on the final output masks to remove isolated, single pixel foreground predictions and generate $64\times 64$ resolution outputs, similar to \textsc{MONet} and \textsc{OP3} before evaluation.

\textbf{Deviation of SpMOT results compared to original publication:} Our results were generated with 100k training frames, while the original TBA paper \citep{He19} uses 2M training frames for the simple SpMOT task. Further, we report the mean of three training runs, while the original paper reports one run (presumably the best). Our best run achieves MOTA of 90.5 (\tabref{tab:resultfull}). Third, we evaluate using intersection over union (IoU) of segmentation masks instead of bounding boxes.

\subsection{\textsc{OP3}}
\textbf{Training:}
The \textsc{OP3} loss is a weighted sum over all refinement and dynamics steps (\eqnref{op3_loss}). For our evaluation on multi-object tracking, we weight all time steps equally. In contrast to the original training loss, in which the weight value is linearly increased indiscriminately, thus weighting later predictions more highly, we perform the linear increase only for the refinement steps between dynamics steps, thus weighting all predictions equally. 

\textsc{OP3}, as published by \citep{Veerapaneni19}, uses curriculum learning. For the first 100 epochs, $M$ refinement steps are taken, followed by a single dynamics step, with a final refinement step afterwards. Starting after 100 epochs, the number of dynamics steps is incremented by 1 every 10 epochs, until five dynamics steps are reached.
Thus, only 5 frames of the sequence are used during training at maximum. 

We chose to use an alternating schedule for training, where after each dynamics step, $M=2$ refinement steps are taken, and this is continued for the entire sequence. Thus, the entire available sequence is used, and error is not propagated needlessly, since the model is enabled to refine previous predictions on the reconstruction before predicting again. Note that this is the schedule \textsc{OP3} uses by default at test-time, when it is used for model predictive control. Note that we still use $4$ refinement steps on the initial observation to update the randomly initialized posterior parameters, as in the released implementation. We split all 10-step sequences into 5-step sequences to avoid premature divergence. 

\textbf{Hyperparameter tuning:}
Note OP3 by default uses a batch size of $80$ with the default learning rate of $0.0003$. As we found training \textsc{\textsc{OP3}} to be very unstable, leading to eventual divergence in almost all experiments that have been performed for this study, we tested batch sizes [$64$, $32$, $16$, $8$] and found reducing the batch size significantly, to $16$, improved performance. We found the default learning rate to be too large for SpMOT in particular, as the model diverged significantly earlier than on VMDS and VOR. We tested learning rates [$0.0002$, $0.0001$, $0.00006$, $0.00003$], and found decreasing the learning rate to $0.0001$ rectified the observed premature divergence.

The checkpoint prior to divergence with the lowest KL loss is selected for the final MOT evaluation, as the KL loss enforces consistency in the latents over the sequence. Interestingly, the checkpoint almost always corresponded to the epochs right before divergence.

\subsection{\textsc{SCALOR}}
\textbf{Architecture:} 
We follow the same architecture as in \citep{Jiang19}. We use a grid of $4 \times 4$ for object discovery with a maximum number of objects of 10. The standard deviation of the image distribution is set to 0.1. Size anchor and variance are set to 0.2 and 0.1, respectively.

For SpMOT, background modeling is disabled and the dimensionality of the latent object appearance is set to 8.

For VMDS, the dimensionality of background is set to 3 and the dimensionality of the latent object appearance is set to 16. For object discovery, a grid of $3 \times 3$ cells with a maximum number of objects of 8 is used.

For VOR, the dimensionality of background is set to 8 and the dimensionality of the latent object appearance is set to 16.

For texVMDS, the dimensionality of background is set to 32 and the dimensionality of the latent object appearance is set to 16. For object discovery, a grid of $3 \times 3$ cells with a maximum number of objects of 8 is used.

Best model checkpoint according to the validation loss was chosen for MOT evaluation.

\textbf{Hyperparameter tuning:}
For VMDS, we run hyperparameter search over number of grid cells \{$3 \times 3$, $4 \times 4$\}, background dimension \{1, 3, 5\}, maximum number of objects \{5, 8, 10\} (dependent on number of grid cells),  size anchor \{0.2, 0.25, 0.3, 0.4\}, z\textsuperscript{what} dimenisonality \{8, 16, 24\} and end value of tau \{0.3, 0.5\}.

For SpMOT,  we run hyperparameter search over maximum number of objects \{4, 10\}, size anchor \{0.1, 0.2, 0.3\}, z\textsuperscript{what} dimensionality \{8, 16\} and whether to model background (with background dimensionality 1) or not.

For VOR,  we run hyperparameter search over size anchor \{0.2, 0.3\} and background dimensionality \{8, 12\}.

For texVMDS, we keep hyperparameters fixed as for VMDS and only tune background dimension \{3, 8, 16, 32, 64, 128\} and z\textsuperscript{what} dimenisonality \{16, 32, 64, 128\} to account for the higher visual complexity of the data set.

We picked best hyper parameters according to the validation loss with the exception of the data set texVMDS, for which the best validation loss was given by the model using a latent dimensionality of 128 for both background and object. This led to the model using only the background slot to reconstruct the whole frame without using the object slots (object matches \raisebox{-0.9ex}{\~{}}3\%). As a sanity check, we picked the model which had the best MOTA on the validation set (shown in \tabref{tab:resultstexvmds}), but it also did not perform well on the more complex texVMDS data set.

We additionally tested if down-weighing the log-likelihood during training by a factor of 10, while keeping the high latent dimensionality (128 for both z\textsuperscript{what} and background) would help, by decreasing the focus on reconstructing the high frequency textures, but it still relied on the background to reconstruct the image without using object slots.

\textbf{Training:}
We train \textsc{SCALOR} with a batch size of 16 for 300 epochs using a learning rate of 0.0001 for SpMOT and VOR and for 400 epochs for VMDS. For the final MOT evaluation, the checkpoint with the lowest loss on the validation set is chosen.

\setcounter{figure}{0}
\setcounter{table}{0}

\section{Additional Results}\label{add_results}

\tabref{tab:resultfull} lists the individual results for the three training runs with different random seeds per model and data set. The results of \textsc{ViMON} and \textsc{SCALOR} are coherent between the three runs with different random seed, while \textsc{TBA} has one run on SpMOT with significantly lower performance than the other two and shows variation in the three training runs on VMDS. \textsc{OP3} exhibits one training run on SpMOT with lower performance than the other two.

\resultstablefull
\resultstablefulltwo

\begin{figure}[t]
    \includegraphics[width=\textwidth]{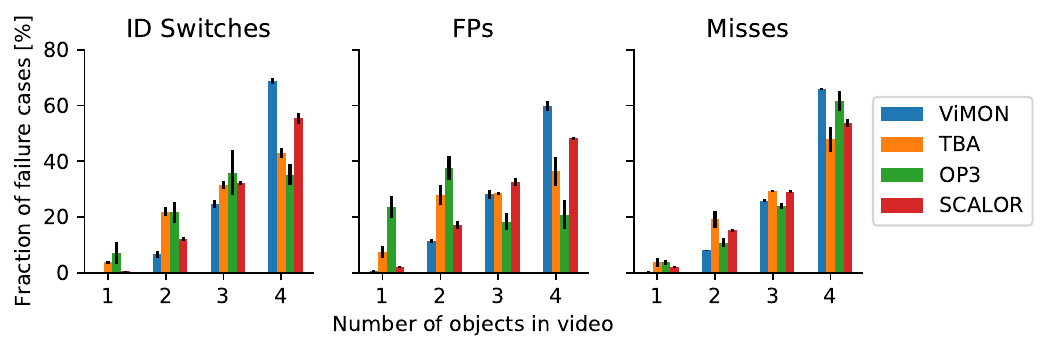}
    \centering
    \caption{Distribution of failure cases dependent on number of objects in VMDS videos split by failure class. Mean of three training runs. Error bars: SD.}
    \label{fig:fail_num_obj_cl}
\end{figure}

\begin{figure}[t]
    \includegraphics[width=0.3\textwidth]{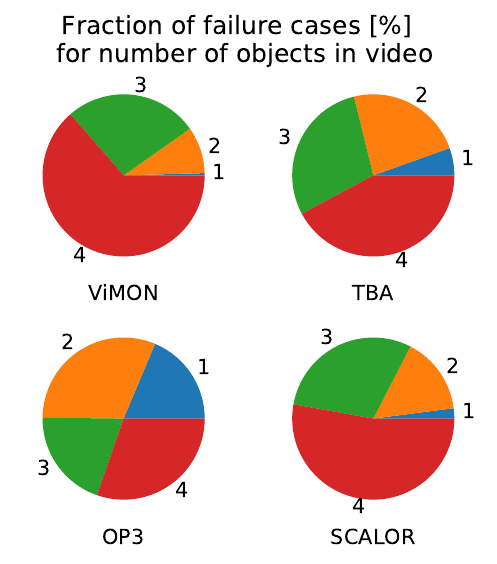}
    \centering
    \caption{Distribution of failure cases dependent on number of objects in VMDS videos for each model. Mean of three training runs.}
    \label{fig:fail_num_obj_pie}
\end{figure}

\figref{fig:fail_num_obj_pie} shows the fraction of failure cases dependent on the number of objects over all types of failures (misses, ID. switches and FPs). \figref{fig:fail_num_obj_cl} shows the fraction of failure cases dependent on the number of objects present in the video for the three different failure cases separately; ID switches, FPs and misses. For \textsc{ViMON}, \textsc{TBA} and \textsc{SCALOR}, the number of failures increase with the number of objects present regardless of the type of failure. In contrast, \textsc{OP3} shows this pattern for ID switches and misses, while it accumulates a higher number of false positives (FPs) in videos with fewer (only one or two) objects.

\begin{figure}[t]
    \centering
    \includegraphics[width=0.7\textwidth]{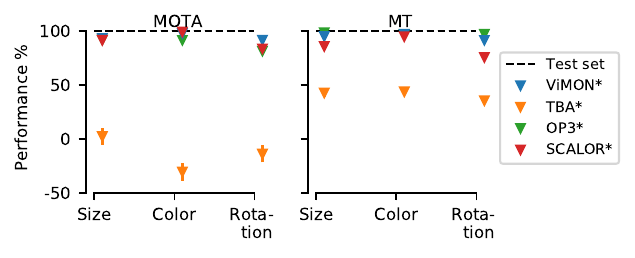}
    \caption{Performance on out-of-distribution sets relative to VMDS test set (100\%). * indicates that models were trained on a data set that included color, size and orientation changes of objects.}
    \label{fig:ood_app}
\end{figure}

\begin{figure}[t]
    \centering
    \includegraphics[width=0.95\textwidth]{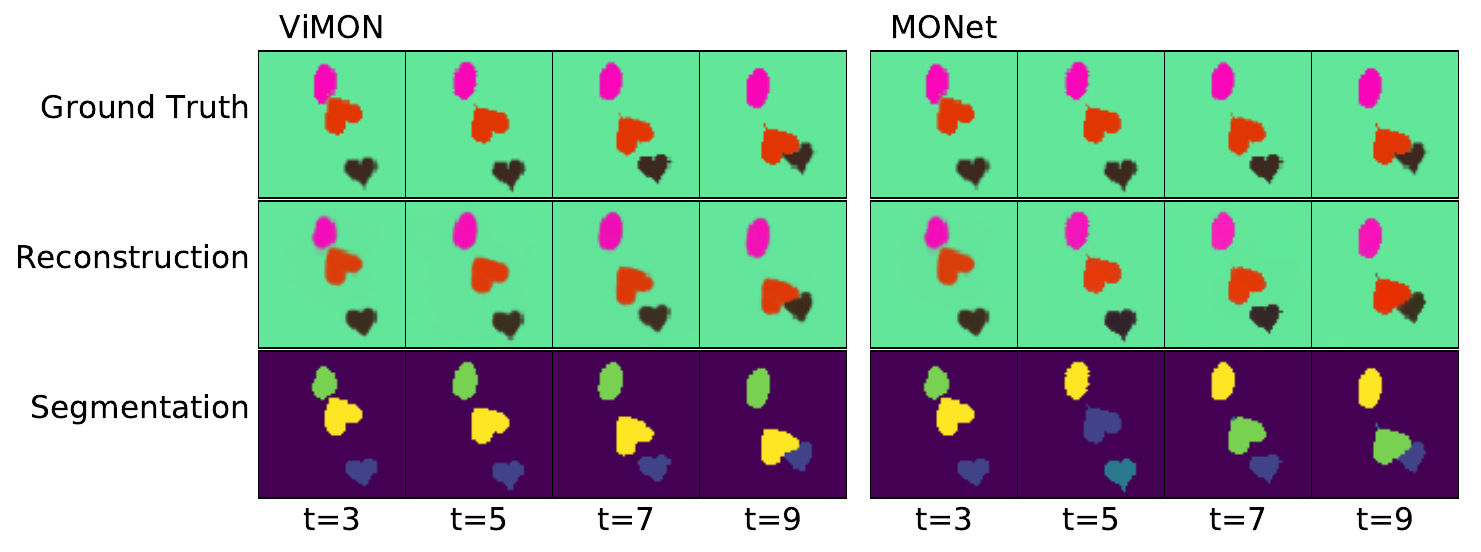}
    \caption{Comparison of \textsc{MONet} and \textsc{ViMON} on VMDS. Example sequence of data set shown with corresponding outputs of the model. Reconstruction shows sum of components from all slots, weighted by the attention masks. Color-coded segmentation maps in third row signify slot-assignment. Note how the object-slot assignment changes for consecutive frames (3rd row) for \textsc{MONet}, while \textsc{ViMON} maintains a consistent slot assignment throughout the video.}
    \label{fig:vamo_vimo}
\end{figure}

\figref{fig:vamo_vimo} shows a comparison between \textsc{MONet} and \textsc{ViMON} on VMDS. \textsc{MONet} correctly finds and segments objects, but it does not assign them to consistent slots over time, while \textsc{ViMON} maintains a consistent slot assignment throughout the video.

\begin{figure}
    \centering
    \includegraphics[width=0.45\textwidth]{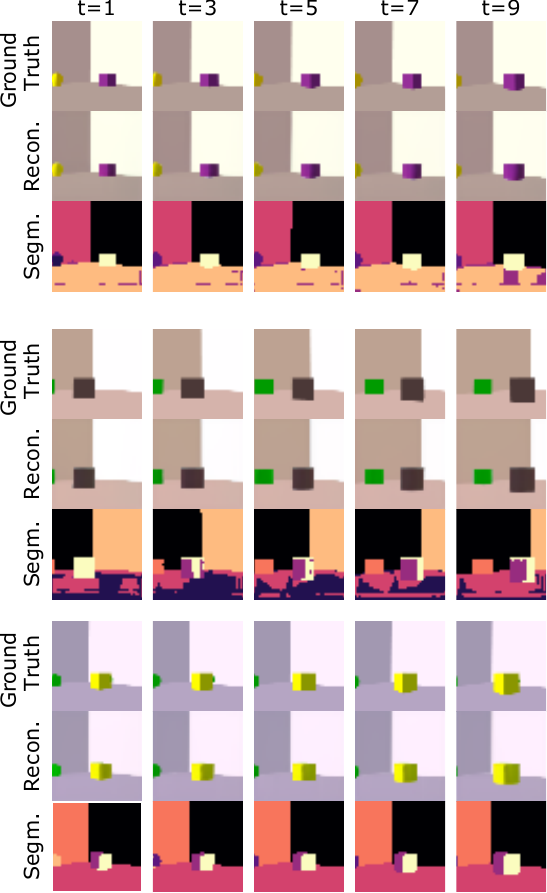}
    \caption{Failure cases of \textsc{OP3} on VOR. Example sequences of VOR test set shown with corresponding outputs of the model after final refinement step. Binarized colour-coded segmentation maps in third row signify slot-assignment.}
    \label{fig:op3_vor_fail}
\end{figure}

\figref{fig:op3_vor_fail} shows failures cases of OP3 on VOR.

\tabref{tab:challenge} and \tabref{tab:ood} list the results for the four models, \textsc{ViMON}, \textsc{TBA}, \textsc{OP3} and \textsc{SCALOR}, on the VMDS challenge sets and out-of-distribution (o.o.d.) sets respectively. Results are shown as the mean and standard deviation of three training runs with different random seed per model.

\subsection{Out-of-distribution test sets}
To test whether the models can in principle learn additional object transformations as featured in the VMDS o.o.d. sets, we additionally train the models on a new training set that includes size and color changes as well as rotation of objects.

\textsc{ViMON}, \textsc{OP3} and \textsc{SCALOR} are able to learn additional property changes of the objects when they are part of the training data while \textsc{TBA} fails to learn tracking on this more challenging data set (\figref{fig:ood_app}; for absolute values \tabref{tab:ood}).

\challengetable
\testtableood

\subsection{Stability of training and runtime} \label{runtime_appendix}
To assess runtime in a fair way despite the models being trained on different hardware, we report the training progress of all models after one hour of training on a single GPU (\tabref{tab:runtime}). In addition, we quantify inference time on the full VMDS test set using a batch size of one.

\runtimetable

\subsection{\textsc{ViMON} Ablations} \label{vimon_abl}

\ablationtable

Removing the GRU or the mask conditioning of the attention network reduces tracking performance (MOTA on VMDS from 86.8\% to 70.6\% and 81.4\%, respectively; \tabref{tab:abl})

\subsection{Impact of IoU Threshold for Matching}
\begin{figure}
    \includegraphics[width=0.83\textwidth]{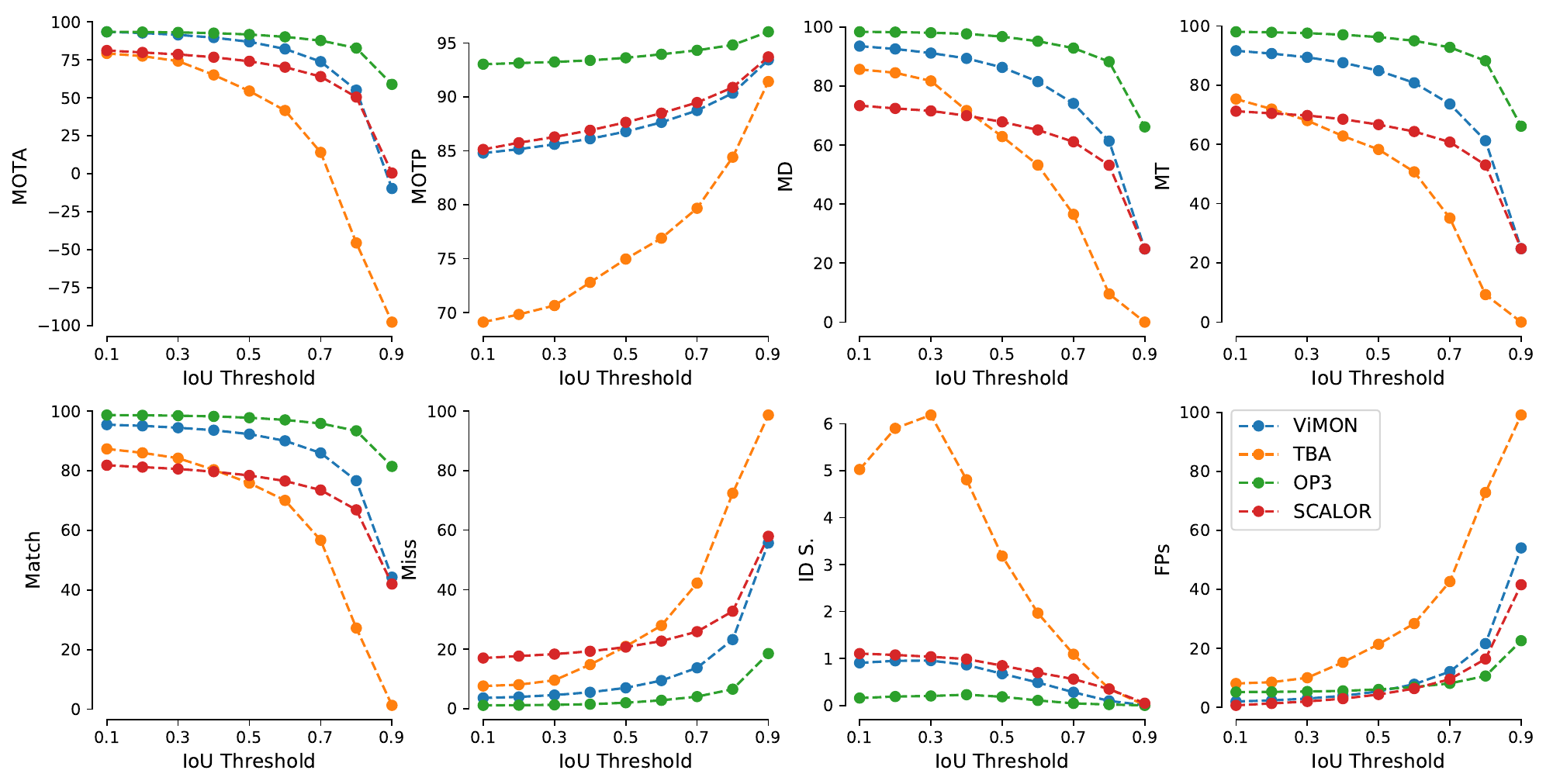}
    \centering
    \caption{Results of \textsc{ViMON}, \textsc{TBA}, \textsc{OP3} and \textsc{SCALOR} on the VMDS test set using different IoU thresholds for matching ground truth objects to model predictions.}
    \label{fig:iou_thresh}
\end{figure}

To examine whether the IoU threshold used for matching object predictions to ground truth impacts the ordering of the models, we compute all metrics with regard to the thresholds [0.1 .. 0.9] in steps of 0.1 on the VMDS test set. The ordering of the models with regard to their MOTA is consistent over all matching thresholds, but the performance difference between the models increases with higher thresholds (\figref{fig:iou_thresh}). For low thresholds (0.1 -- 0.3) \textsc{TBA} outperforms \textsc{SCALOR} in MD, MT, Matches and Misses but has a significantly lower segmentation performance as measured by MOTP.

\FloatBarrier

\setcounter{figure}{0}
\setcounter{table}{0}

\section{Supplementary Figures}\label{add_visual}
\begin{figure}
    \includegraphics[width=0.83\textwidth]{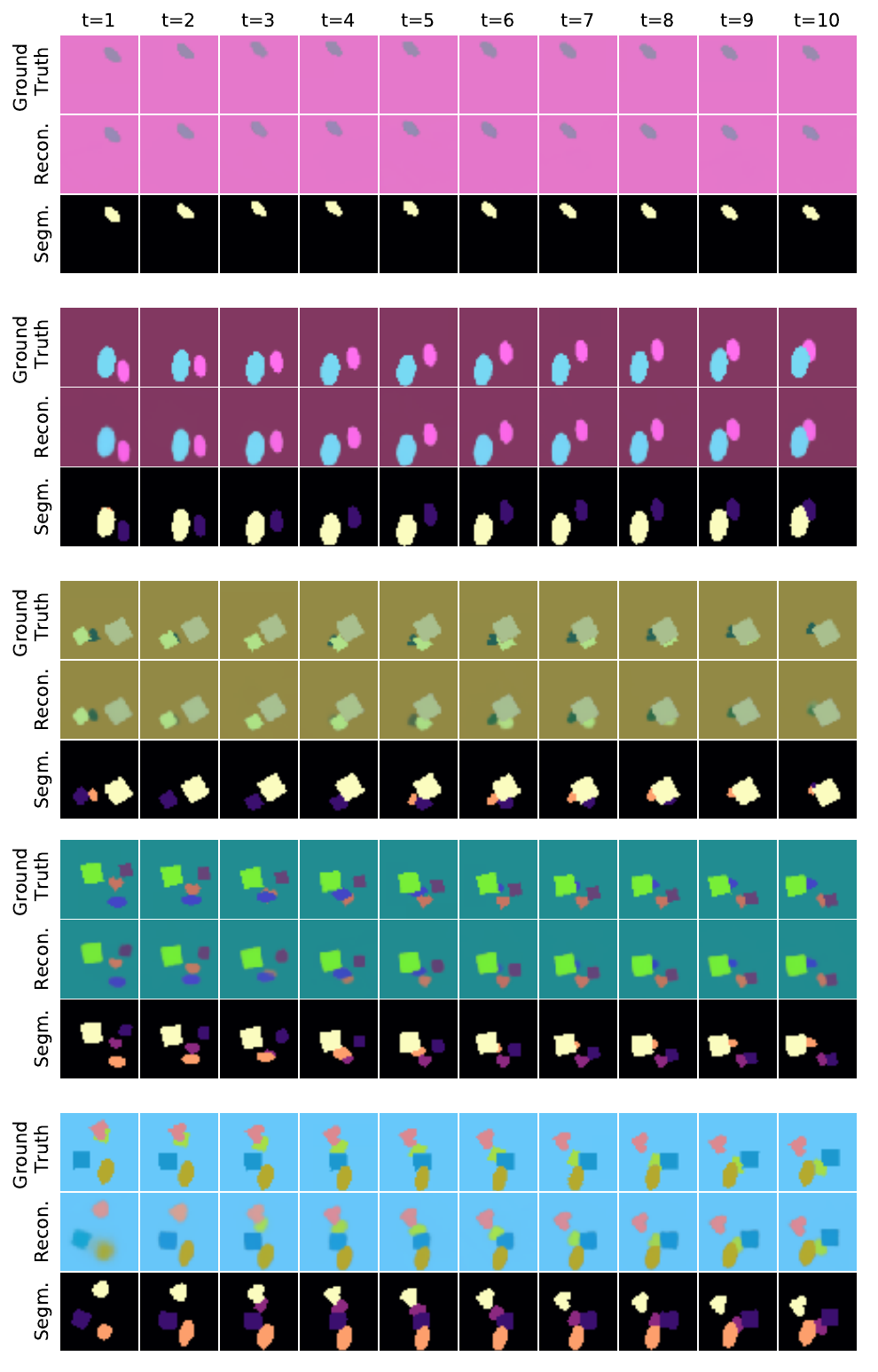}
    \centering
    \caption{Results of \textsc{ViMON} on VMDS. Random example sequences of VMDS test set shown with corresponding outputs of the model. Reconstruction shows sum of components from all slots, weighted by the reconstructed masks from the VAE. Binarized colour-coded segmentation maps in third row signify slot-assignment.}
    \label{vimon_vmds_rand}
\end{figure}

\begin{figure}
    \includegraphics[width=0.83\textwidth]{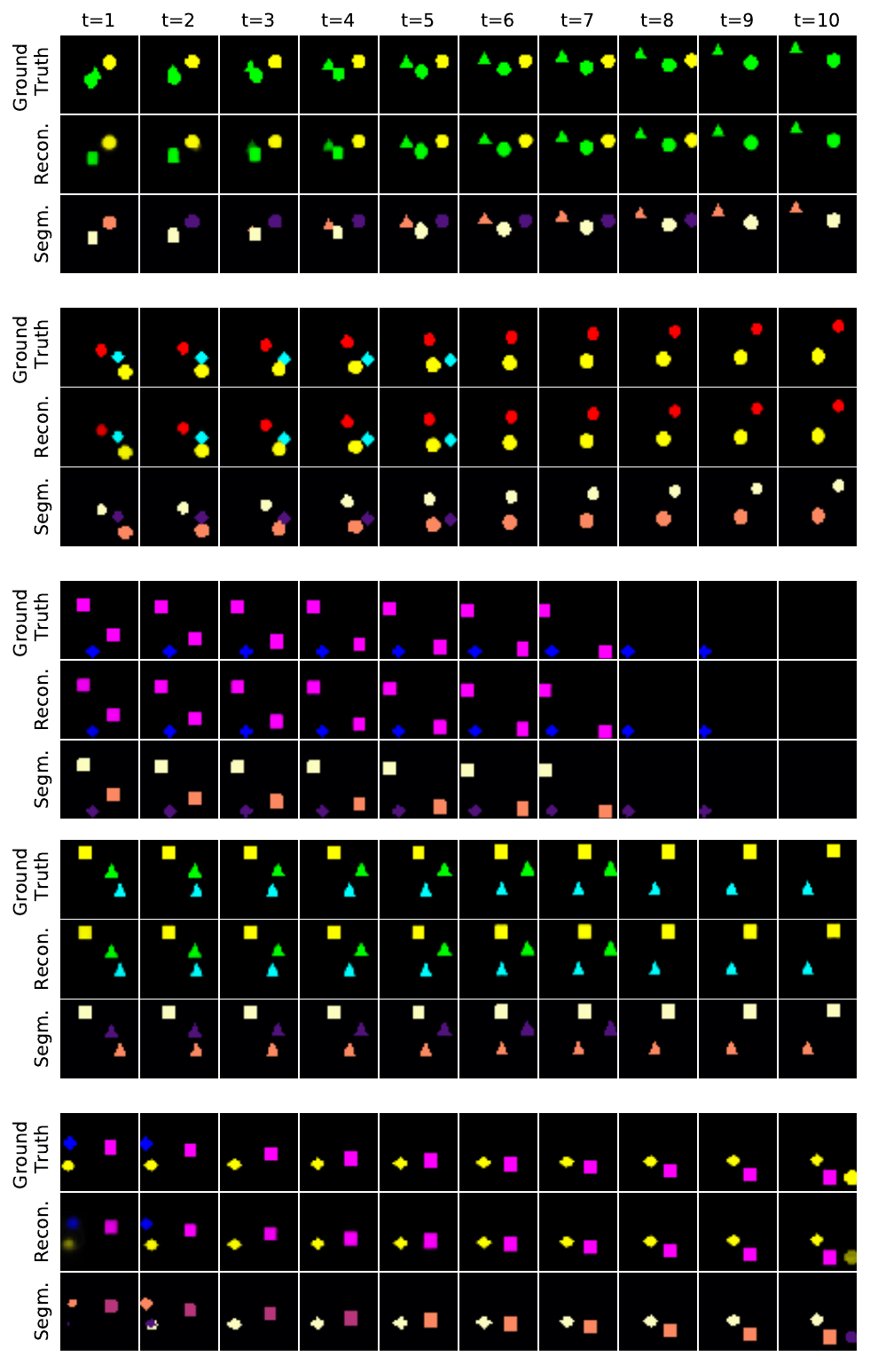}
    \centering
    \caption{Results of \textsc{ViMON} on SpMOT. Random example sequences of SpMOT test set shown with corresponding outputs of the model. Reconstruction shows sum of components from all slots, weighted by the reconstructed masks from the VAE. Binarized colour-coded segmentation maps in third row signify slot-assignment.}
    \label{vimon_spmot_rand}
\end{figure}

\begin{figure}
    \includegraphics[width=0.83\textwidth]{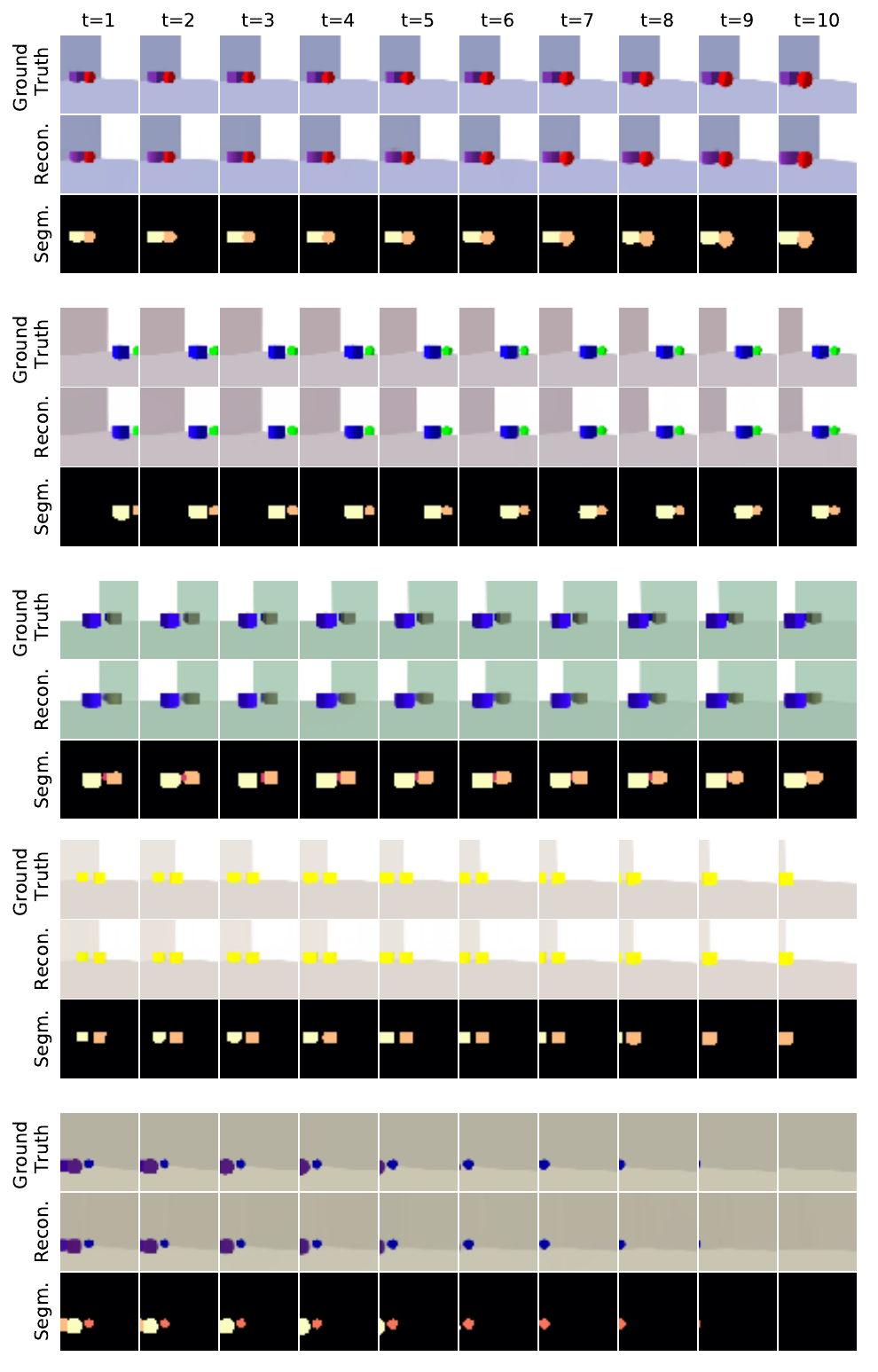}
    \centering
    \caption{Results of \textsc{ViMON} on VOR. Random example sequences of VOR test set shown with corresponding outputs of the model. Reconstruction shows sum of components from all slots, weighted by the reconstructed masks from the VAE. Binarized colour-coded segmentation maps in third row signify slot-assignment.}
    \label{vimon_vor_rand}
\end{figure}

\begin{figure}
    \includegraphics[width=0.83\textwidth]{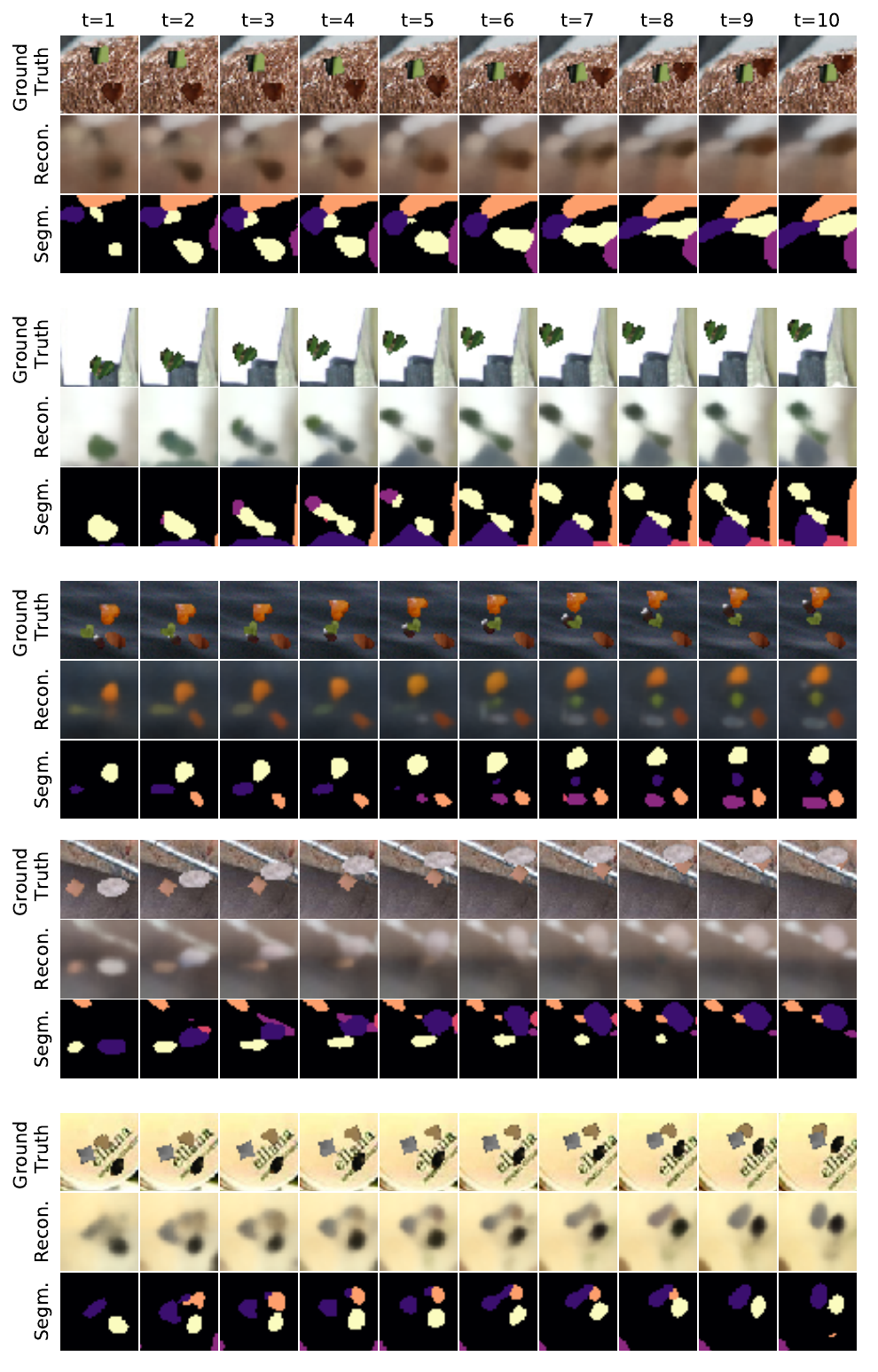}
    \centering
    \caption{Results of \textsc{ViMON} on texVMDS. Random example sequences of texVMDS test set shown with corresponding outputs of the model. Reconstruction shows sum of components from all slots, weighted by the reconstructed masks from the VAE. Binarized colour-coded segmentation maps in third row signify slot-assignment.}
    \label{vimon_tex_vmds_rand}
\end{figure}

\begin{figure}
    \includegraphics[width=0.83\textwidth]{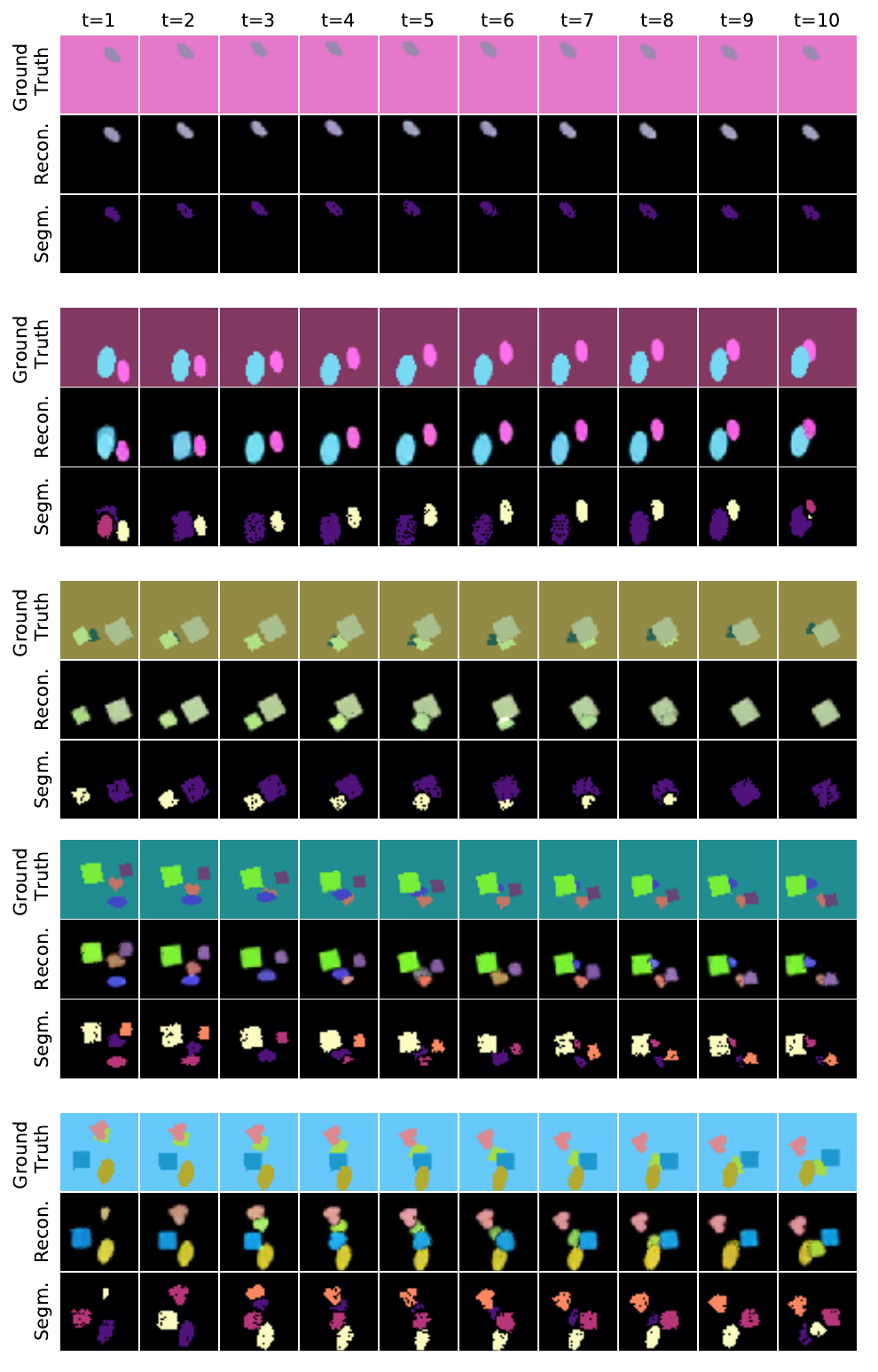}
    \centering
    \caption{Results of \textsc{TBA} on VMDS. Random example sequences of VMDS test set shown with corresponding outputs of the model. Binarized colour-coded segmentation maps in third row signify slot-assignment. Note that background subtraction is performed in the preprocessing of TBA, hence the black background in the reconstructions.}
    \label{tba_vmds_rand}
\end{figure}

\begin{figure}
    \includegraphics[width=0.83\textwidth]{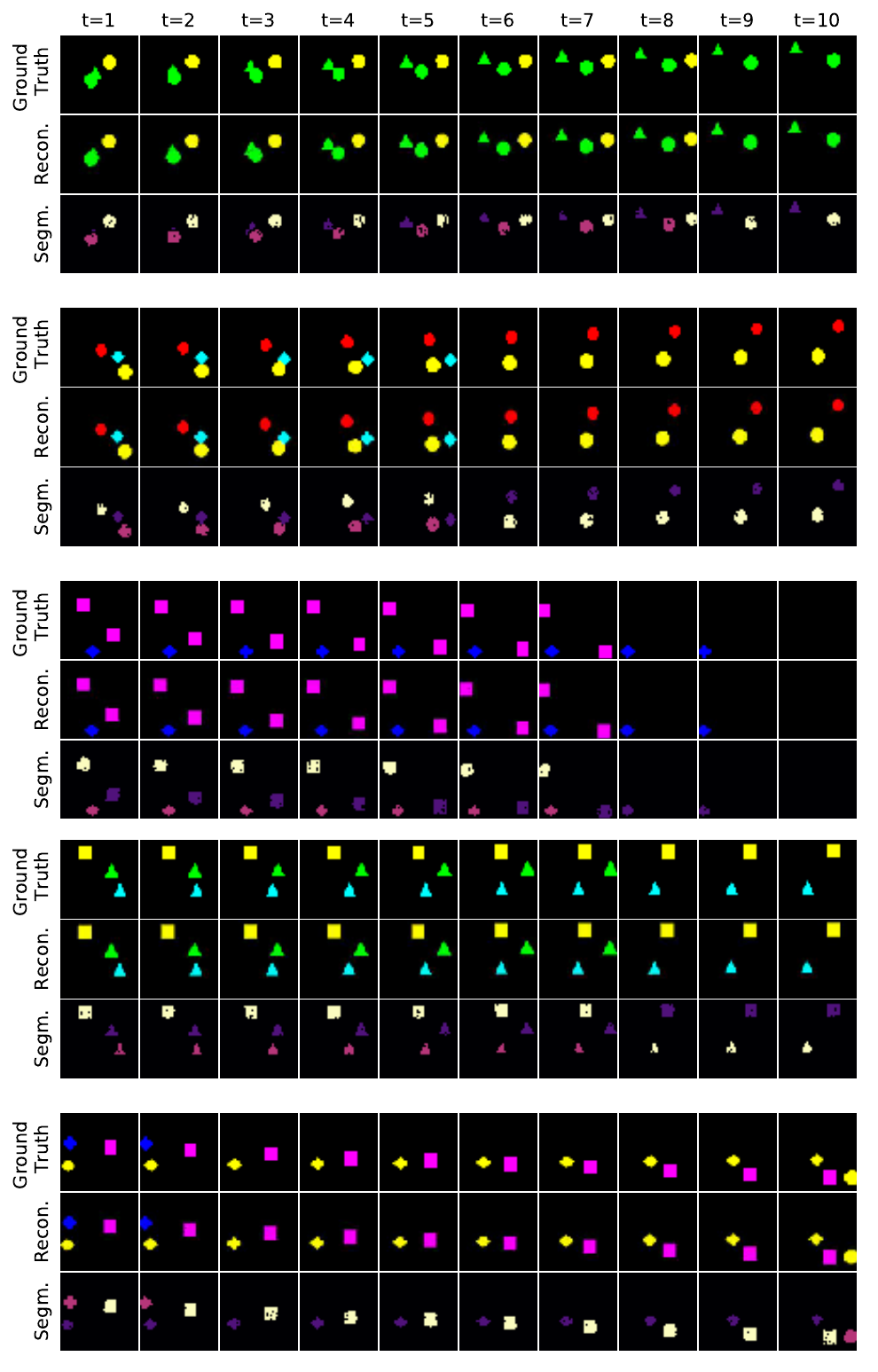}
    \centering
    \caption{Results of \textsc{TBA} on SpMOT. Random example sequences of SpMOT test set shown with corresponding outputs of the model. Binarized colour-coded segmentation maps in third row signify slot-assignment.}
    \label{tba_spmot_rand}
\end{figure}

\begin{figure}
    \includegraphics[width=0.83\textwidth]{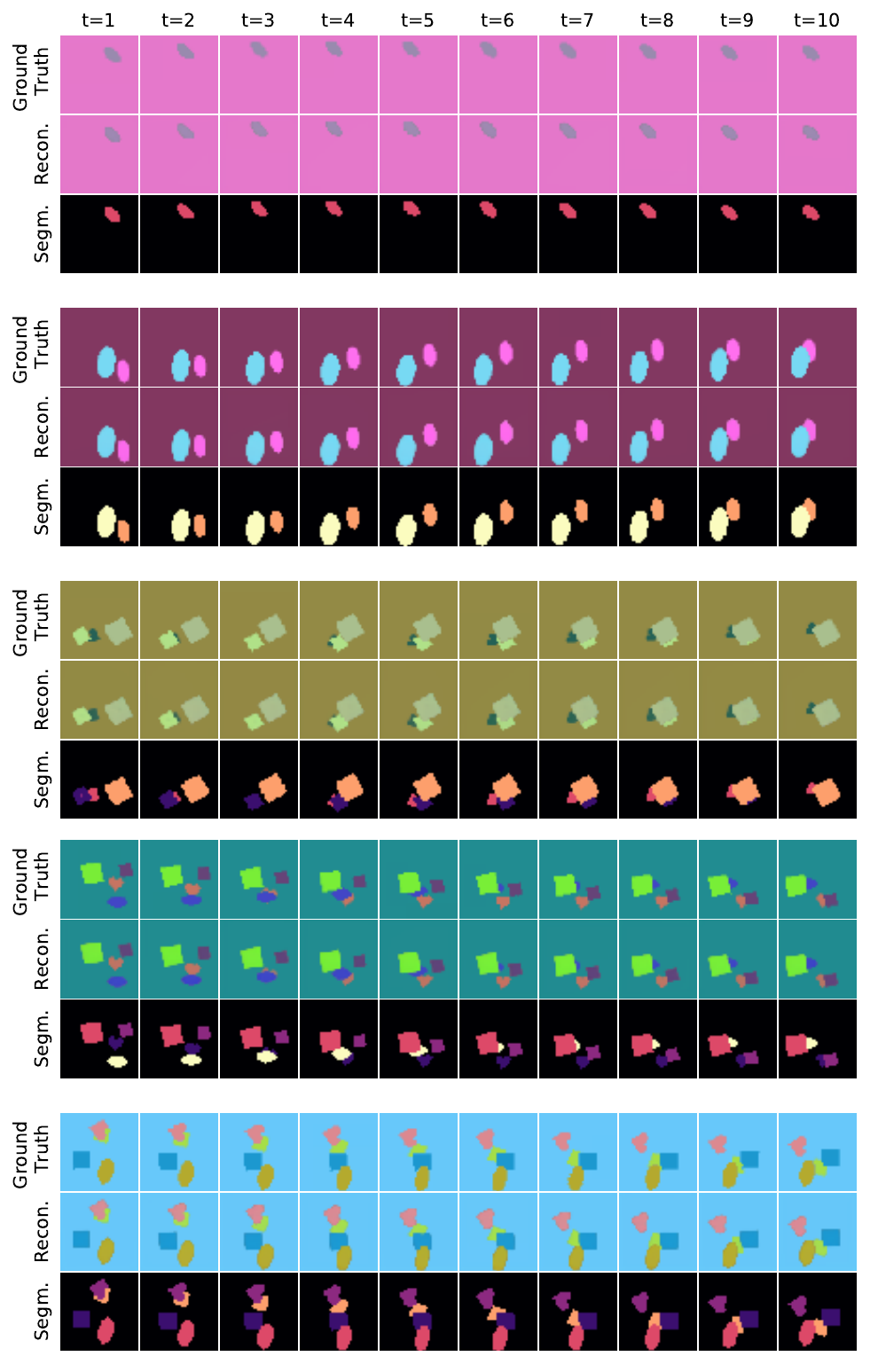}
    \centering
    \caption{Results of \textsc{OP3} on VMDS. Random example sequences of VMDS test set shown with corresponding outputs of the model after final refinement step. Binarized colour-coded segmentation maps in third row signify slot-assignment.}
    \label{op3_vmds_rand}
\end{figure}

\begin{figure}
    \includegraphics[width=0.83\textwidth]{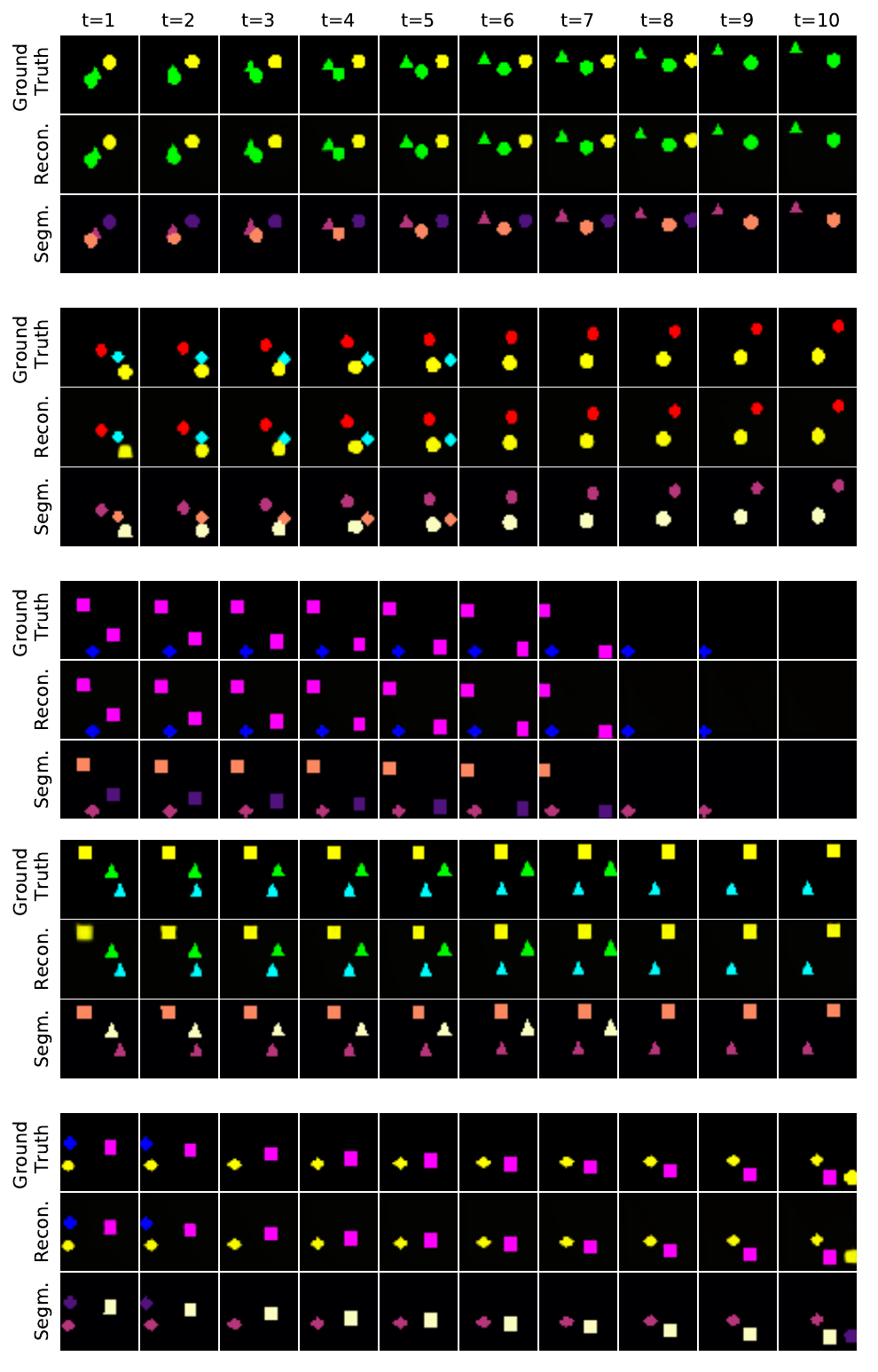}
    \centering
    \caption{Results of \textsc{OP3} on SpMOT. Random example sequences of SpMOT test set shown with corresponding outputs of the model after final refinement step. Binarized colour-coded segmentation maps in third row signify slot-assignment.}
    \label{op3_spmot_rand}
\end{figure}

\begin{figure}
    \includegraphics[width=0.83\textwidth]{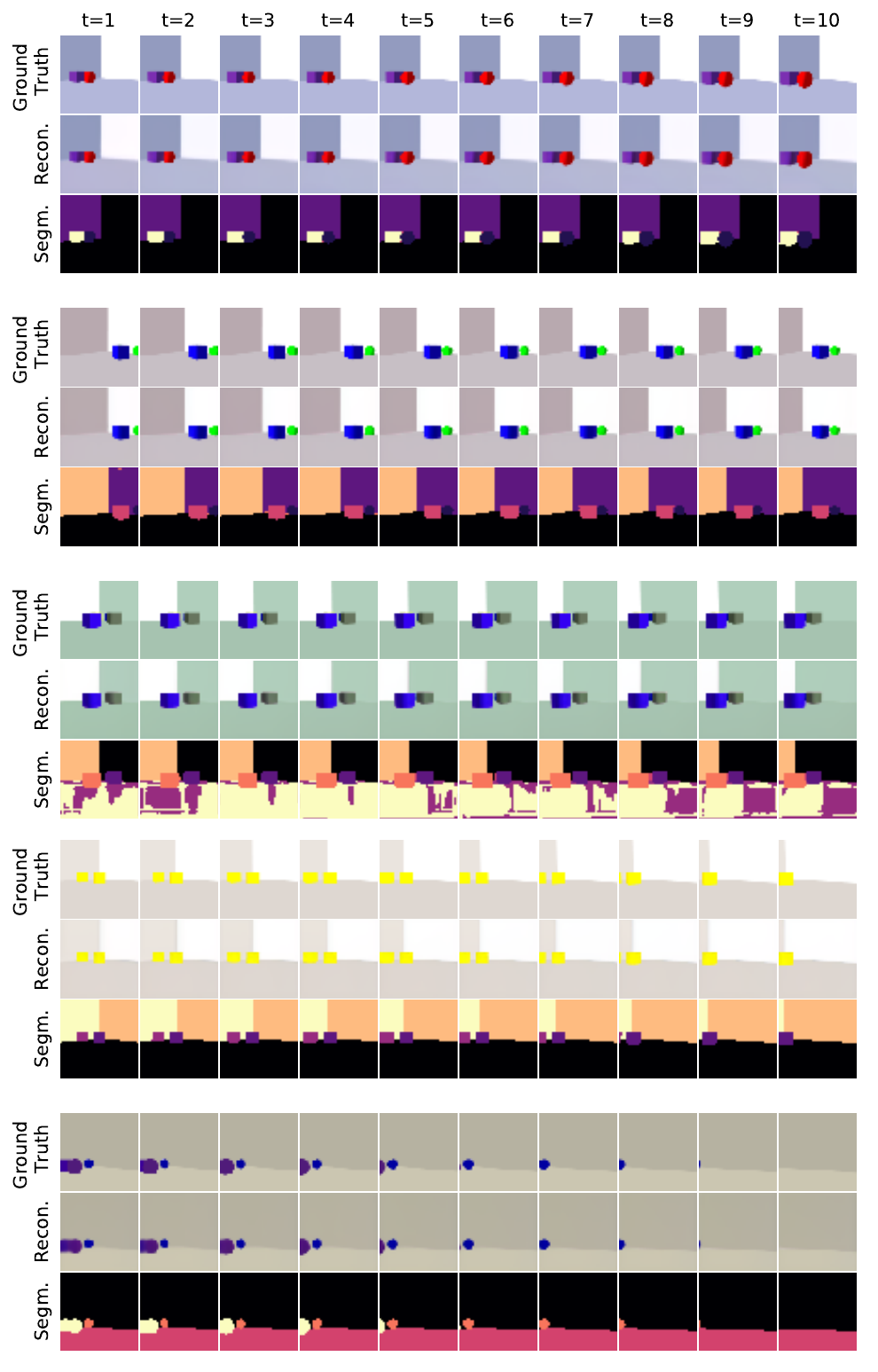}
    \centering
    \caption{Results of \textsc{OP3} on VOR. Random example sequences of VOR test set shown with corresponding outputs of the model after final refinement step. Binarized colour-coded segmentation maps in third row signify slot-assignment.}
    \label{op3_vor_rand}
\end{figure}

See figures \ref{vimon_vmds_rand} -- \ref{op3_vor_rand} for additional, randomly picked examples of reconstruction and segmentation for \textsc{ViMON}, \textsc{TBA}, \textsc{OP3} and \textsc{SCALOR} on the four data sets (VMDS, SpMOT, VOR and texVMDS).

\begin{figure}
    \includegraphics[width=0.83\textwidth]{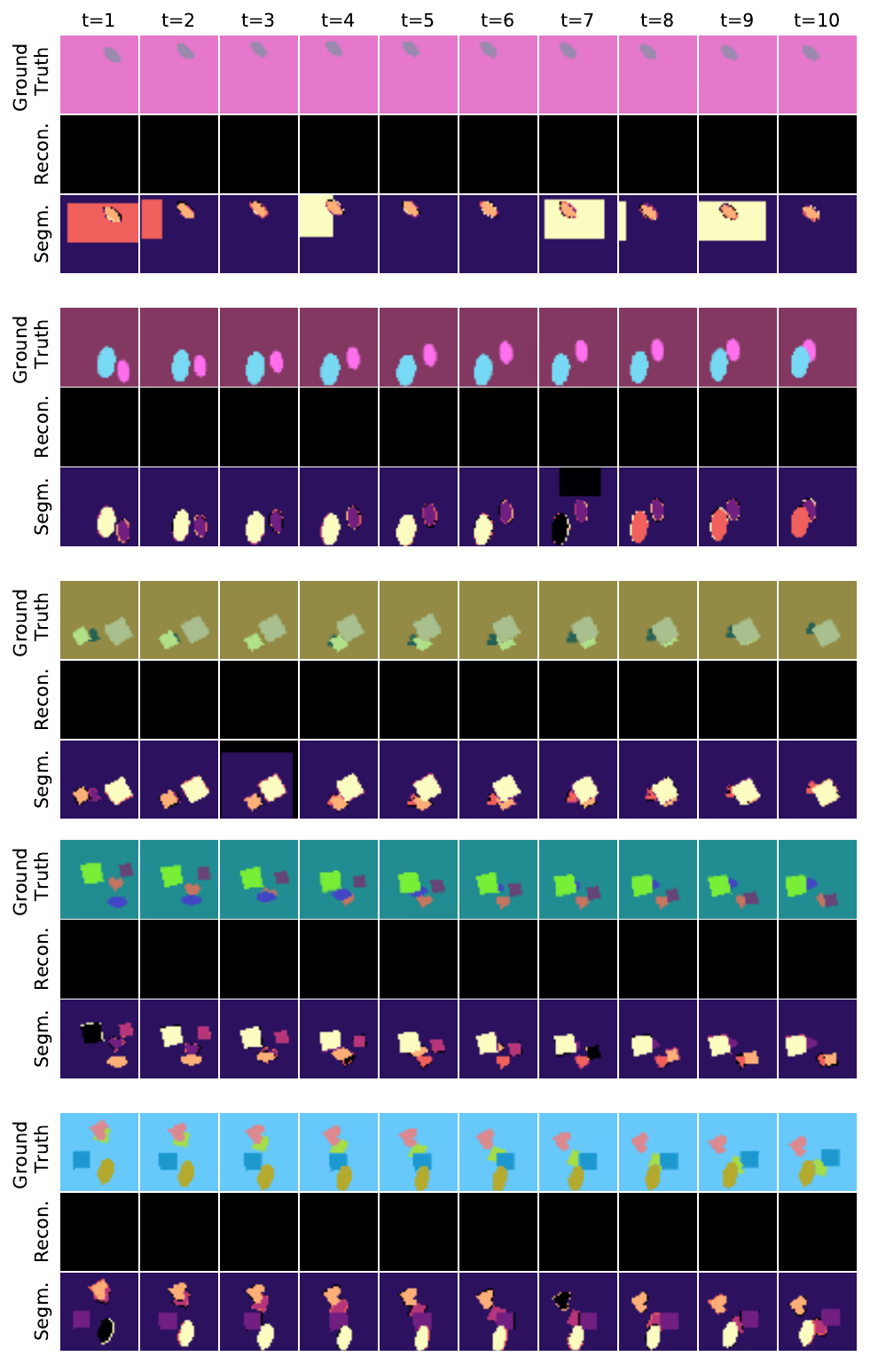}
    \centering
    \caption{Results of \textsc{k-Means} on VMDS. Random example sequences of VMDS test set shown with corresponding outputs of the model. Note that k-Means does not give a reconstruction of the input.}
    \label{kmeans_vmds_rand}
\end{figure}

\begin{figure}
    \includegraphics[width=0.83\textwidth]{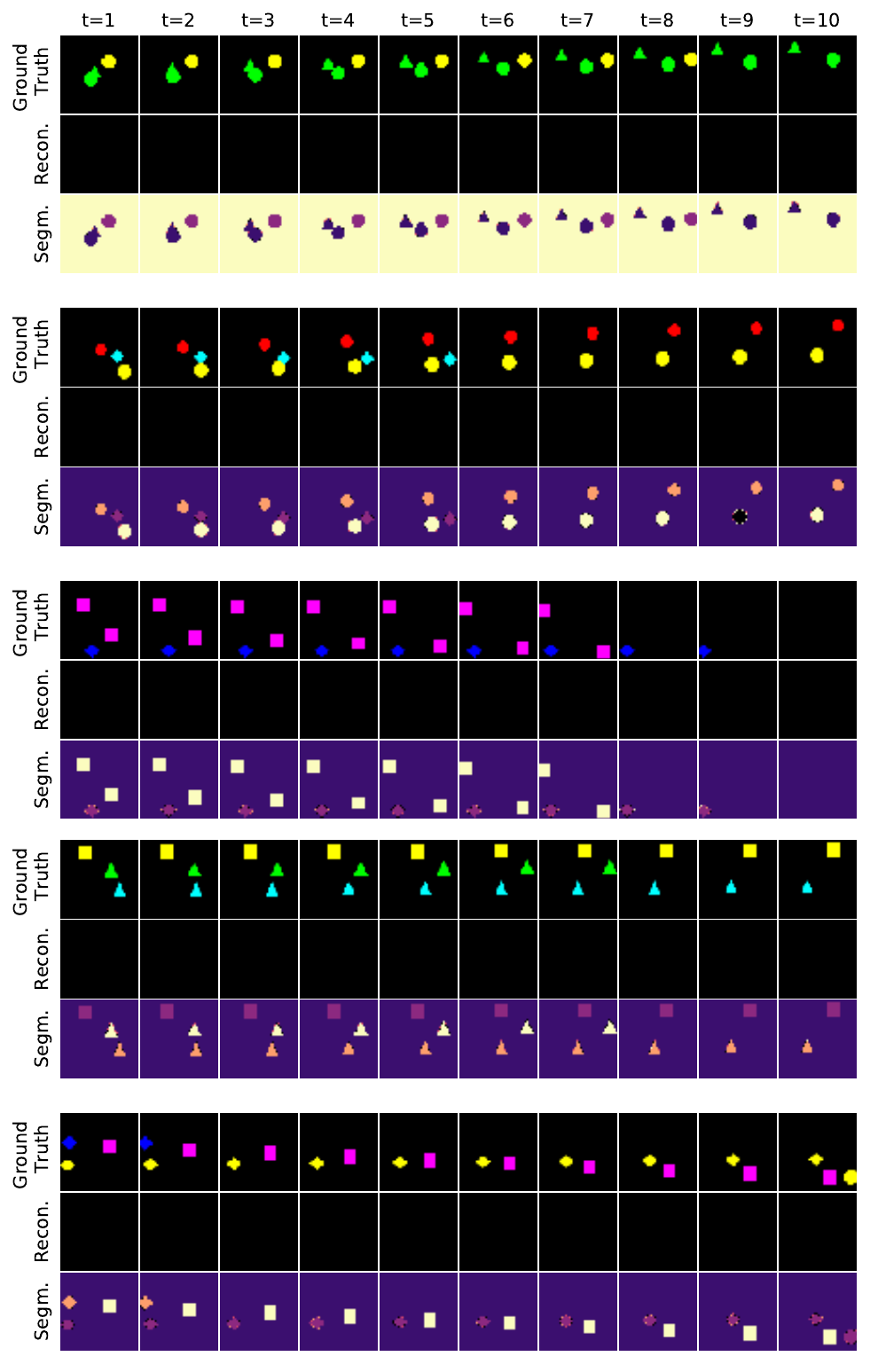}
    \centering
    \caption{Results of \textsc{k-Means} on SpMOT. Random example sequences of SpMOT test set shown with corresponding outputs of the model. Note that k-Means does not give a reconstruction of the input.}
    \label{kmeans_spmot_rand}
\end{figure}

\begin{figure}
    \includegraphics[width=0.83\textwidth]{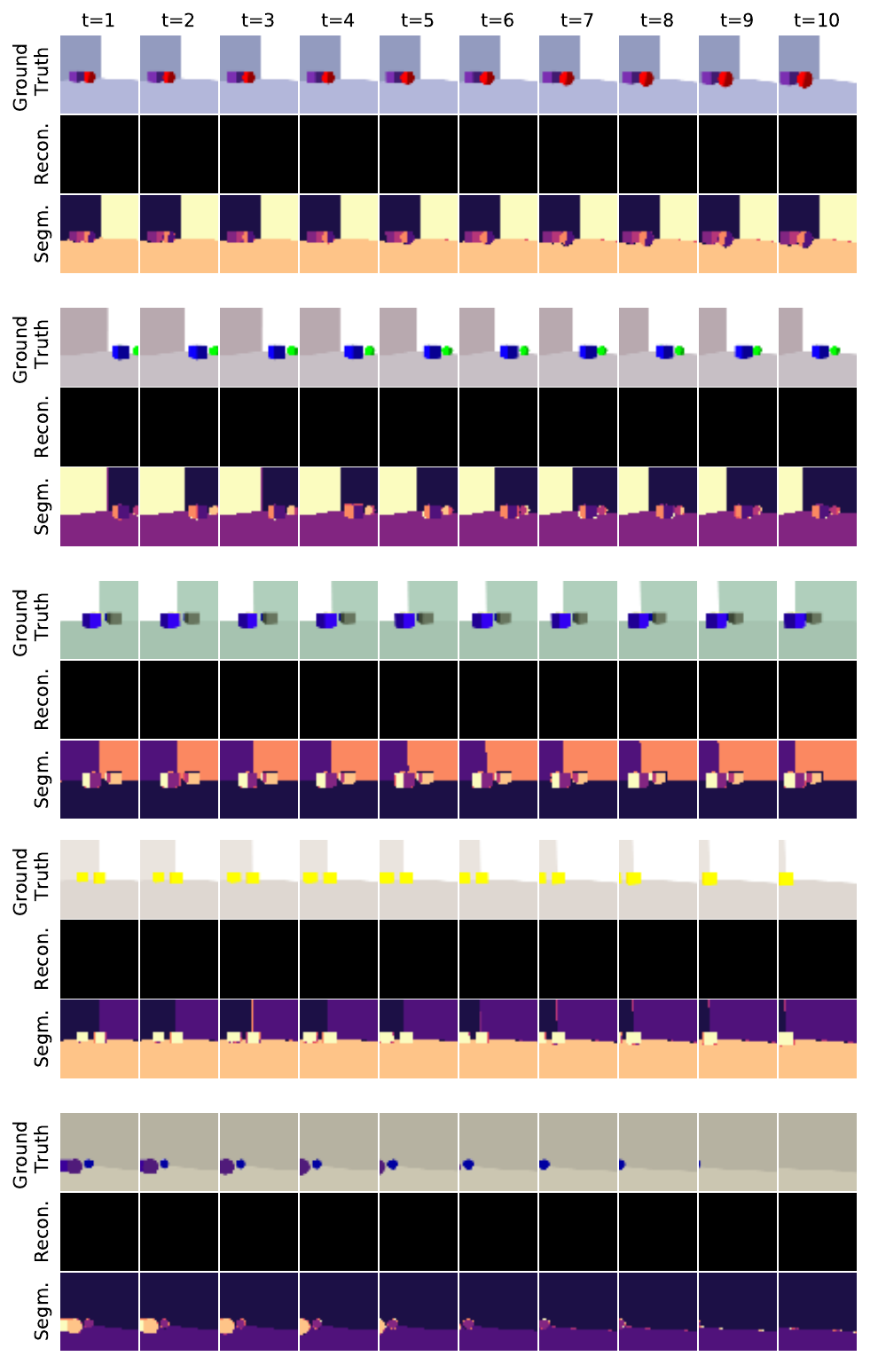}
    \centering
    \caption{Results of \textsc{k-Means} on VOR. Random example sequences of VOR test set shown with corresponding outputs of the model. Note that k-Means does not give a reconstruction of the input.}
    \label{kmeans_vor_rand}
\end{figure}

\begin{figure}
    \includegraphics[width=0.83\textwidth]{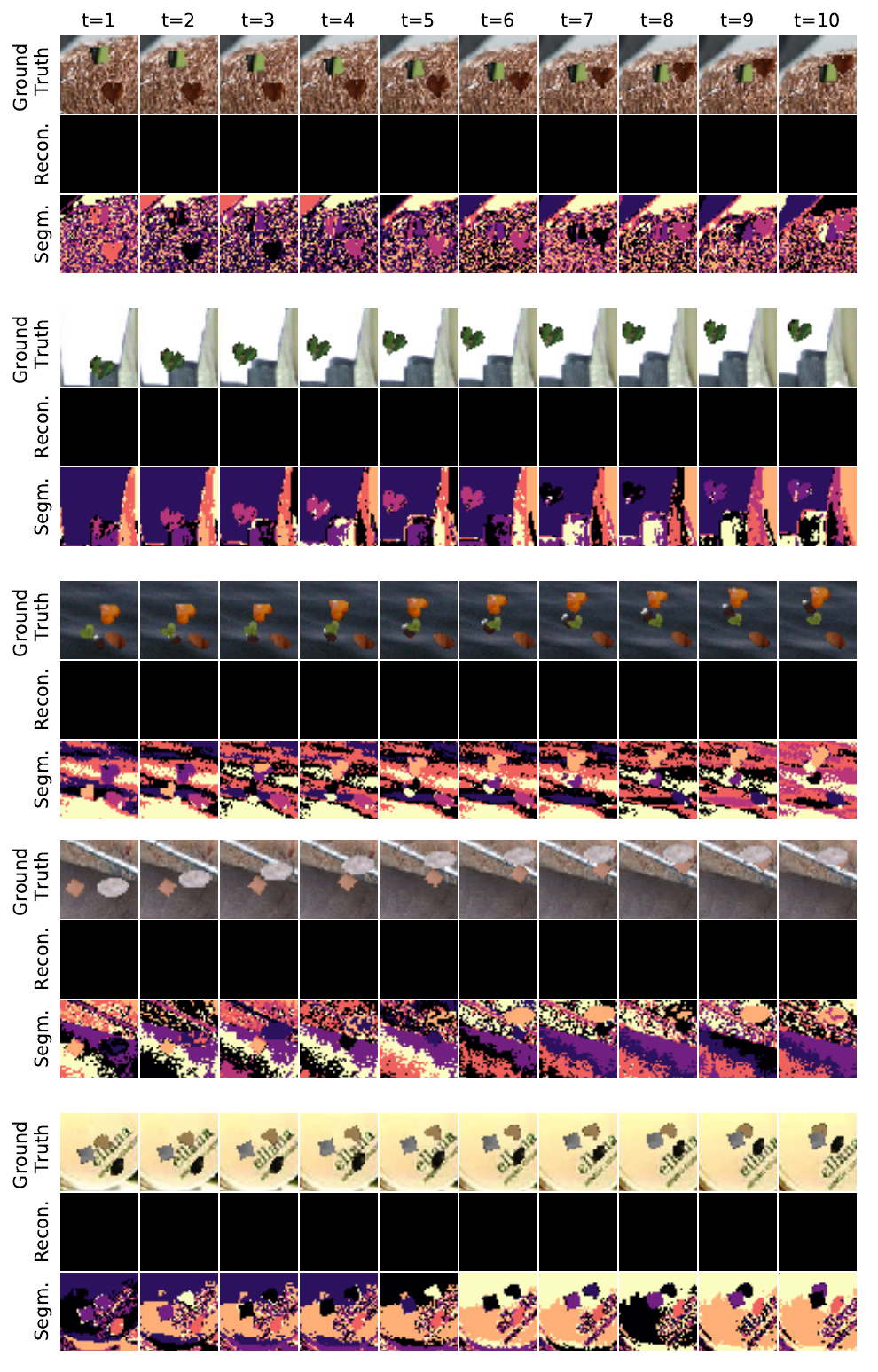}
    \centering
    \caption{Results of \textsc{k-Means} on texVMDS. Random example sequences of texVMDS test set shown with corresponding outputs of the model. Note that k-Means does not give a reconstruction of the input.}
    \label{kmeans_texvmds_rand}
\end{figure}

\begin{figure}
    \includegraphics[width=0.83\textwidth]{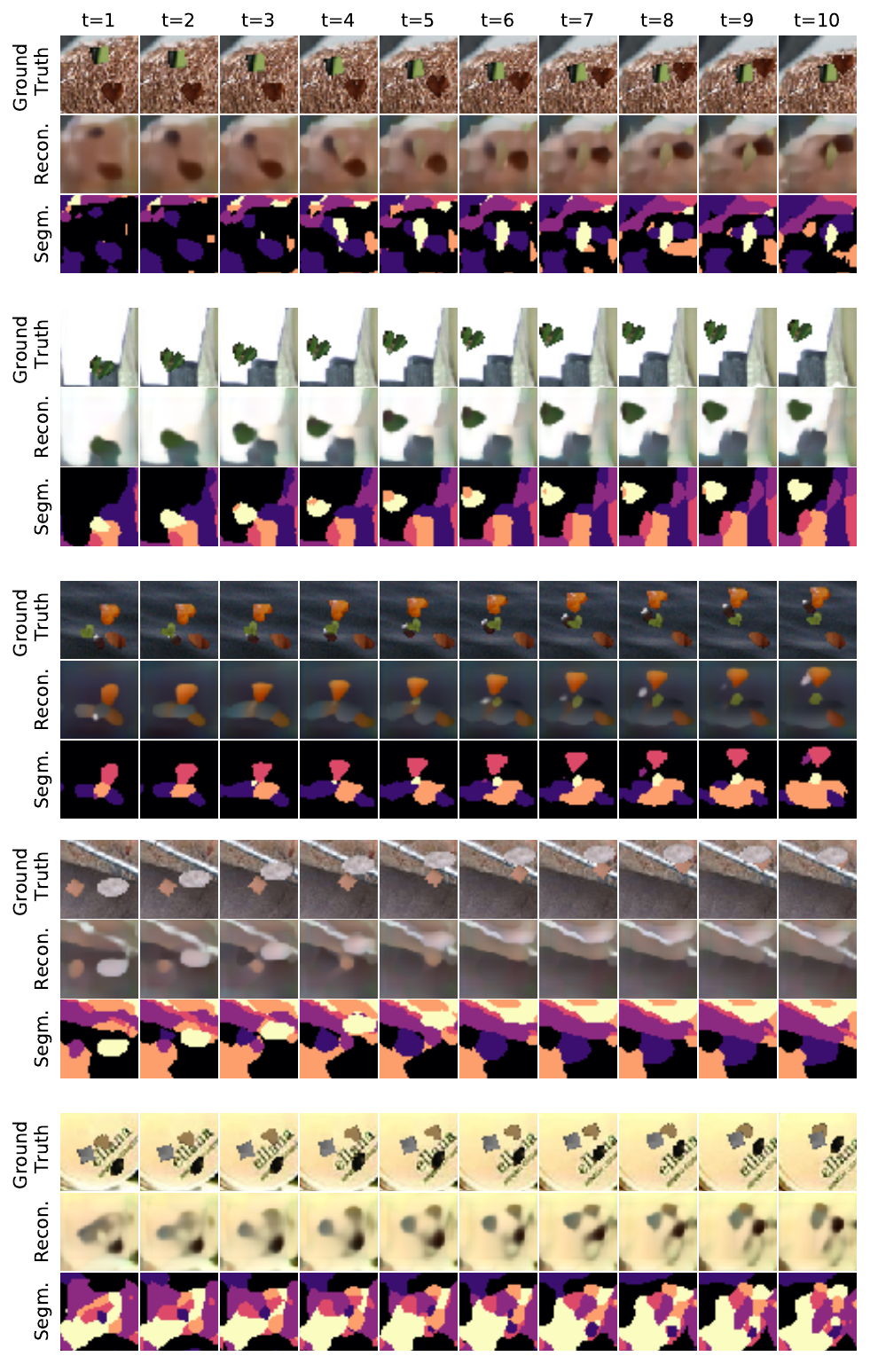}
    \centering
    \caption{Results of \textsc{OP3} on texVMDS. Random example sequences of texVMDS test set shown with corresponding outputs of the model after final refinement step. Binarized colour-coded segmentation maps in third row signify slot-assignment.}
    \label{op3_tex_vmds_rand}
\end{figure}

\begin{figure}
    \includegraphics[width=0.83\textwidth]{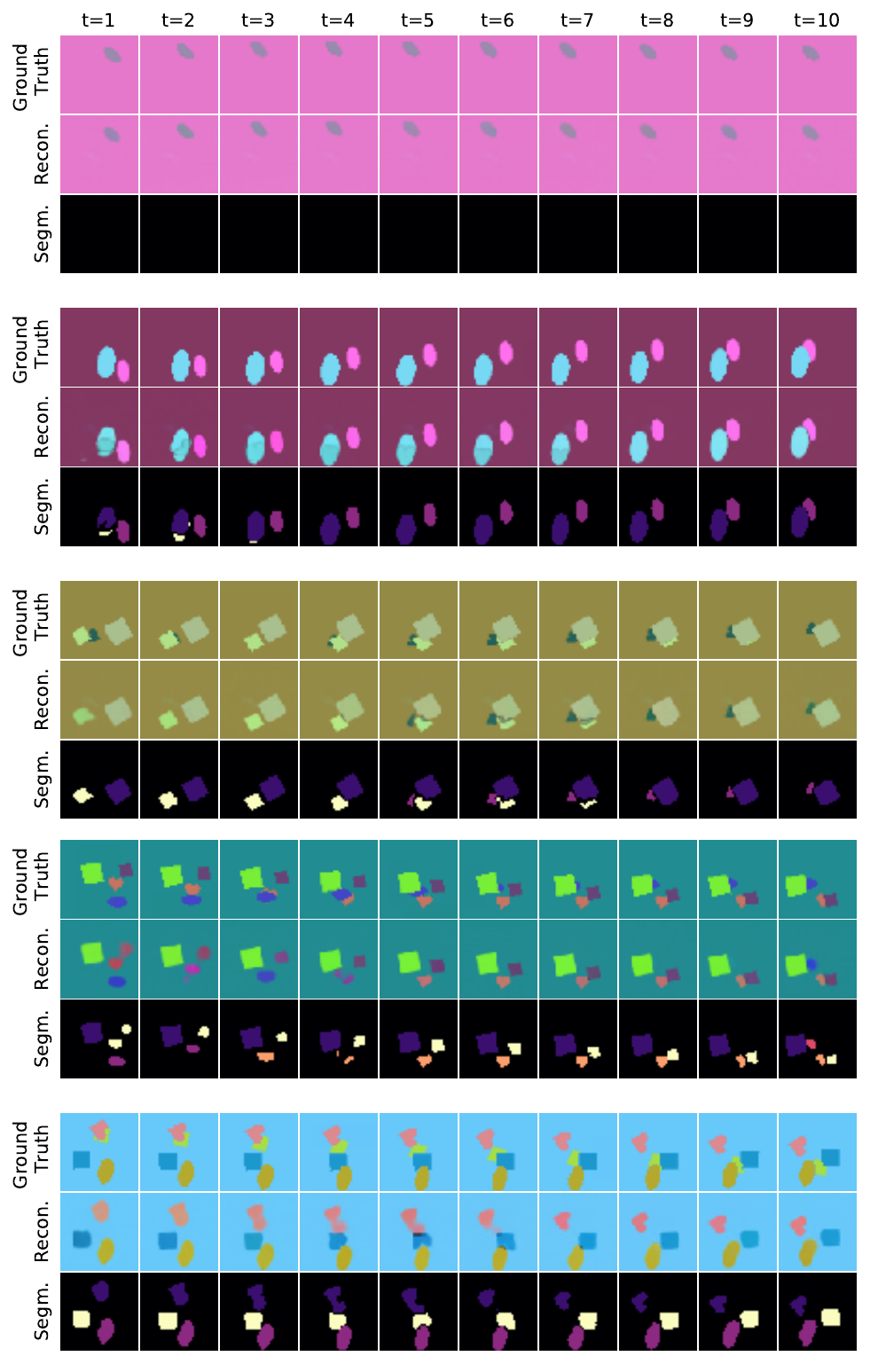}
    \centering
    \caption{Results of \textsc{SCALOR} on VMDS. Random example sequences of VMDS test set shown with corresponding outputs of the model. Binarized colour-coded segmentation maps in third row signify slot-assignment.}
    \label{scalor_vmds_rand}
\end{figure}

\begin{figure}
    \includegraphics[width=0.83\textwidth]{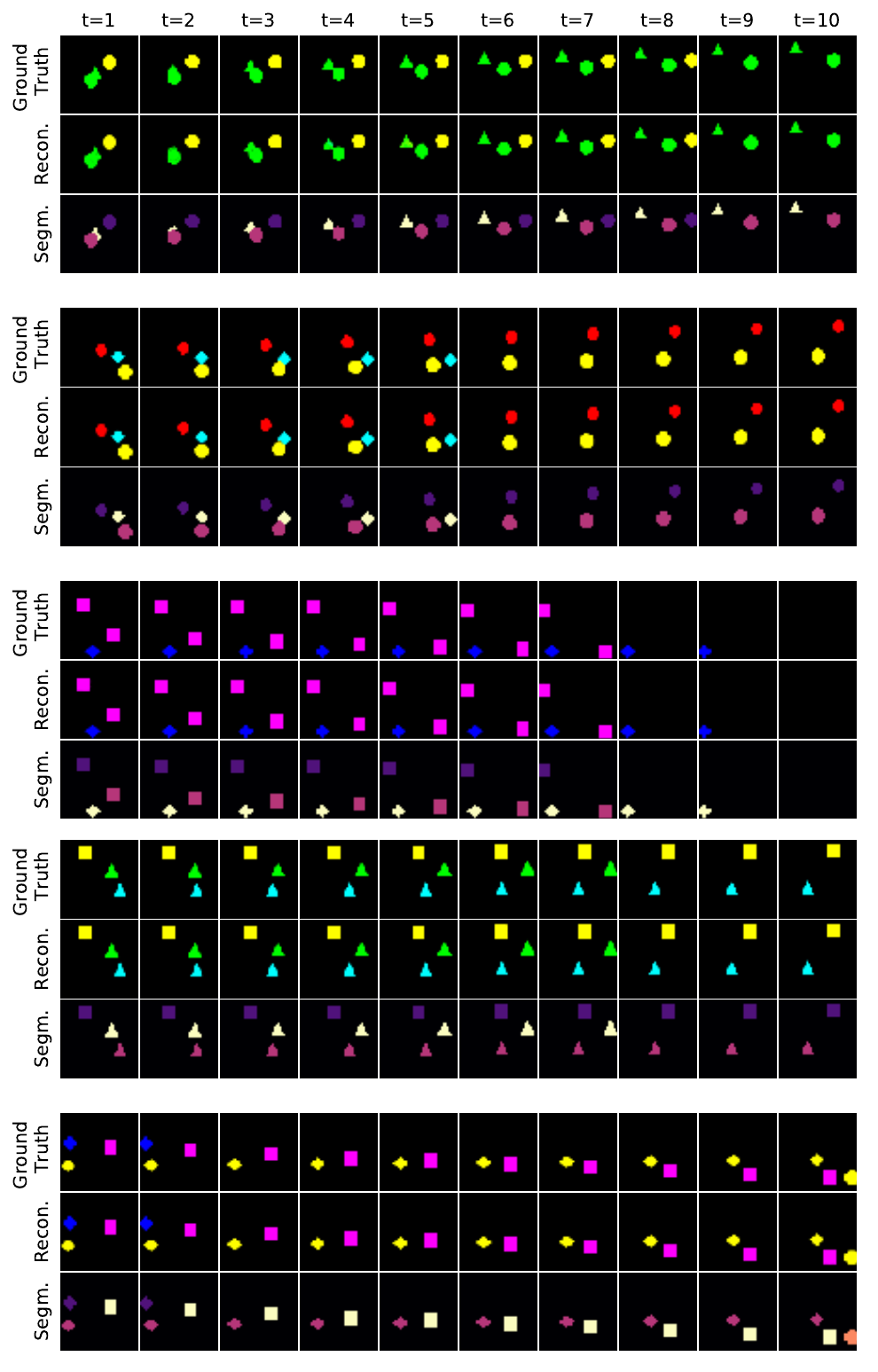}
    \centering
    \caption{Results of \textsc{SCALOR} on SpMOT. Random example sequences of SpMOT test set shown with corresponding outputs of the model. Binarized colour-coded segmentation maps in third row signify slot-assignment.}
    \label{scalor_spmot_rand}
\end{figure}

\begin{figure}
    \includegraphics[width=0.83\textwidth]{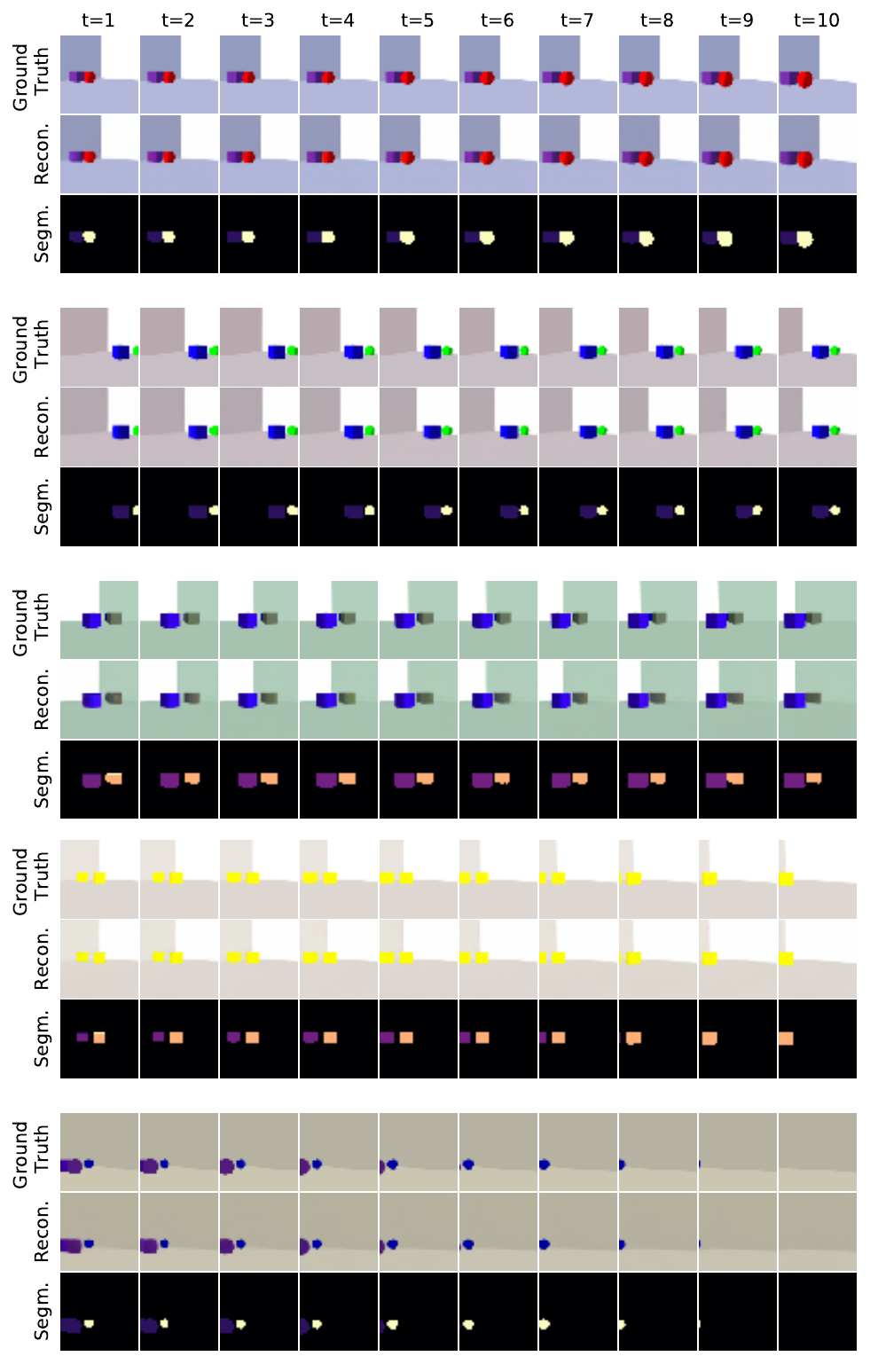}
    \centering
    \caption{Results of \textsc{SCALOR} on VOR. Random example sequences of VOR test set shown with corresponding outputs of the model. Binarized colour-coded segmentation maps in third row signify slot-assignment.}
    \label{scalor_vor_rand}
\end{figure}

\begin{figure}
    \includegraphics[width=0.83\textwidth]{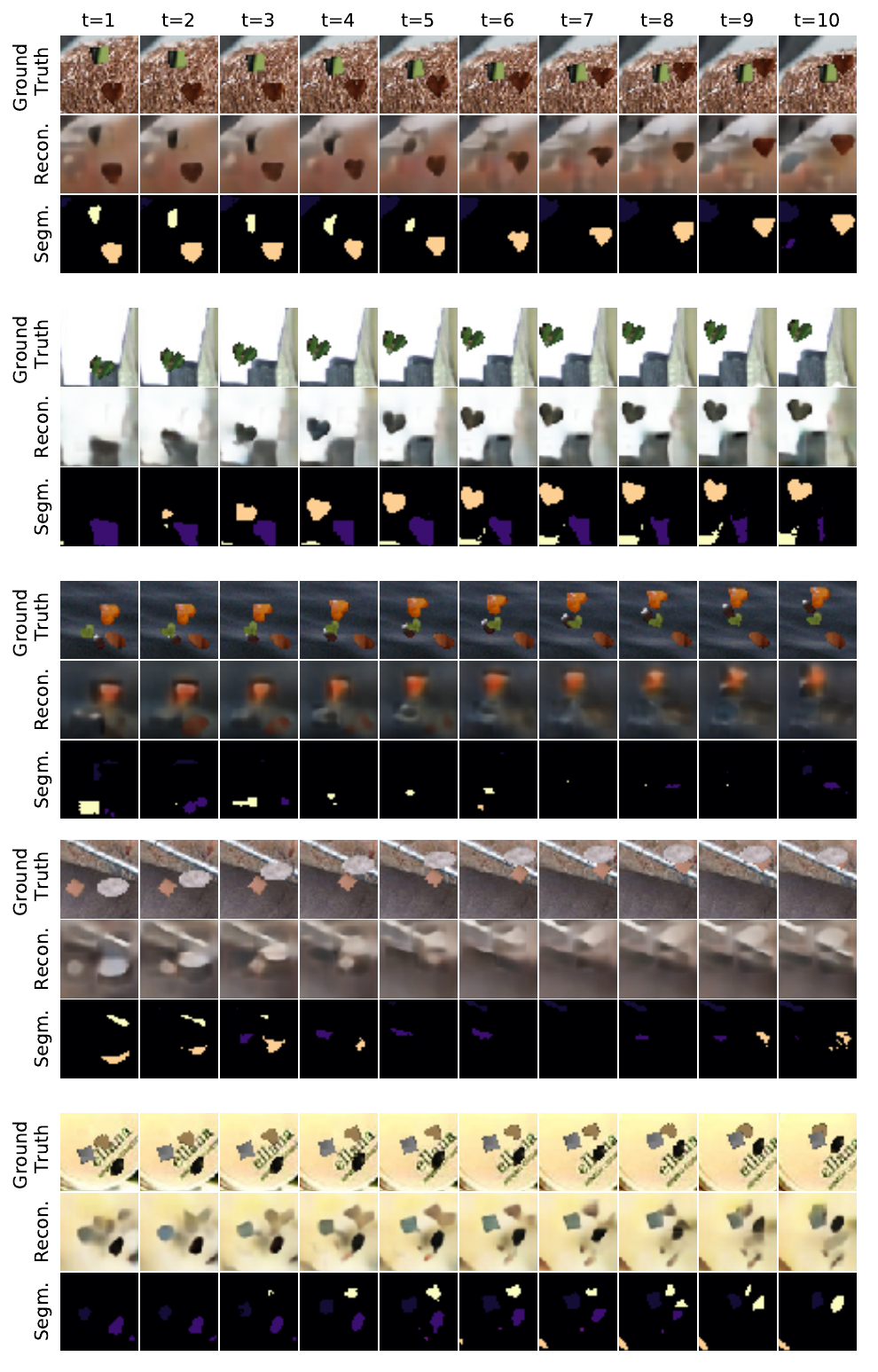}
    \centering
    \caption{Results of \textsc{SCALOR} on texVMDS. Random example sequences of texVMDS test set shown with corresponding outputs of the model. Binarized colour-coded segmentation maps in third row signify slot-assignment.}
    \label{scalor_tex_vmds_rand}
\end{figure}

\FloatBarrier


\end{document}